\documentclass[journal, dvipsnames]{IEEEtran}
\usepackage{inputenc}
\usepackage{amsmath, amssymb , graphicx}
\usepackage{mathtools}
\usepackage{wasysym}
\usepackage{bm}
\usepackage{bbm}
\usepackage{cite}

\usepackage[export]{adjustbox}

\usepackage[font=footnotesize]{subcaption}
\usepackage{booktabs}
\usepackage{array}
\usepackage{dblfloatfix}
\newcolumntype{L}[1]{>{\raggedright\let\newline\\\arraybackslash\hspace{0pt}}m{#1}}
\newcolumntype{C}[1]{>{\centering\let\newline\\\arraybackslash\hspace{0pt}}m{#1}}
\newcolumntype{R}[1]{>{\raggedleft\let\newline\\\arraybackslash\hspace{0pt}}m{#1}}

\usepackage{balance}
\usepackage{graphicx}

\usepackage{multirow}
\usepackage{tabularx}

\usepackage{algpseudocode}
\usepackage[linesnumbered,ruled,vlined,noend]{algorithm2e}

\usepackage{hyperref}
\usepackage{xcolor}

\DeclareMathOperator*{\argmin}{arg\,min}

\title{\LARGE \bf CRANE: A Redundant, Multi-Degree-of-Freedom Computed Tomography Robot for Heightened Needle Dexterity within a Medical Imaging Bore}

\author{Dimitrious Schreiber$^{\dag,\S}$ \IEEEmembership{Student Member, IEEE}, Zhaowei Yu$^{\dag}$, Taylor Henderson$^{\dag}$, Derek Chen$^{\dag}$,
Alexander Norbash$^\ddag$, and Michael C. Yip$^\dag$ \IEEEmembership{Senior Member, IEEE}

\thanks{$^\dag$ denotes affiliation with the Department of Electrical and Computer Engineering, University of California San Diego, La Jolla, CA 92093 USA. {\tt\small \{dschreib, zhy125, tjwest, dec001, yip\}@ucsd.edu}}%
\thanks{$^\ddag$ denotes affiliation with the Department of Radiology, University of California San Diego, La Jolla, CA 92093 USA. {\tt\small anorbash@ucsd.edu }}

\thanks{$^\S$ denotes affiliation with Air Surgical, Inc., San Diego, CA 92121 USA. {\tt\small dimitri@air-surgical.com}}}

\begin{document}

\maketitle
%\thispagestyle{empty}
%\pagestyle{empty}

%Be more general about the transmissions it applies.

\begin{abstract}
    Computed Tomography (CT) image guidance enables accurate and safe minimally invasive treatment of diseases, including cancer and chronic pain, with needle-like tools via a percutaneous approach. The physician incrementally inserts and adjusts the needle with intermediate images due to the accuracy limitation of free-hand adjustment and patient physiological motion. Scanning frequency is limited to minimize ionizing radiation exposure for the patient and physician. Robots can provide high positional accuracy and compensate for physiological motion with fewer scans. To accomplish this, the robots must operate within the confined imaging bore while retaining sufficient dexterity to insert and manipulate the needle. 
This paper presents CRANE: CT Robotic Arm and Needle Emplacer, a CT-compatible robot with a design focused on system dexterity that enables physicians to manipulate and insert needles within the scanner bore as naturally as they would be able to by hand. We define abstract and measurable clinically motivated metrics for in-bore dexterity applicable to general-purpose intra-bore image-guided needle placement robots, develop an automatic robot planning and control method for intra-bore needle manipulation and device setup, and demonstrate the redundant linkage design provides dexterity across various human morphology and meets the clinical requirements for target accuracy during an in-situ evaluation. 

\end{abstract}


\vspace{-3mm}

\section{Introduction}
Interventional radiologists (IRs) perform minimally invasive procedures under real-time medical imaging guidance, such as ultrasound, fluoroscopy, computed tomography (CT), and magnetic resonance imaging (MRI). In recent years, percutaneous CT-guided procedures have increased dramatically in both type and frequency due to their decreased complications and recovery time compared to more invasive open-surgery and laparoscopic surgery, enabled by technique and technological advances  \cite{jones_best_2018, leng_radiation_2018}. They allow the minimally invasive diagnosis and therapy of numerous diseases. These include lung, liver, and kidney cancer and chronic lumbar-sacral spine pain, which in combination affect approximately one-in-six people during their lifetime \cite{noauthor_lifetime_nodate, noauthor_global_nodate}. Lung cancer alone has approximately 230,000 new diagnoses and 150,000 deaths yearly in the United States \cite{cancerorg}. Physicians use needle-like tools to diagnose and treat these diseases, including fine needles for injection and aspiration, core-biopsy needles, and ablation probes. Accurately needle insertion is challenging, especially for small lesions. This results in the average diagnosis necessitating 1.7 insertion attempts \cite{zhang_biopsy_2020}, in addition to frequent complications and a high repeat procedure rate of 43\% \cite{chiu_costs_2021, zhang_biopsy_2020}, with an increasing incidence of severe side-effects (including collapsed lung, up to 62.2\%) for the patient and a 40-80\% increase in cost attributable to repeat procedures \cite{wiener_population-based_2011}. Liver and renal procedures also have similarly high repeat procedure rates, costs, and accompanying side-effects for repeat procedures \cite{di_tommaso_role_2019, izzo_radiofrequency_2019, massarweh_trends_2010, beermann_1000_2018}.

\begin{figure}[!t]
    \centering
      \includegraphics[width=\linewidth]{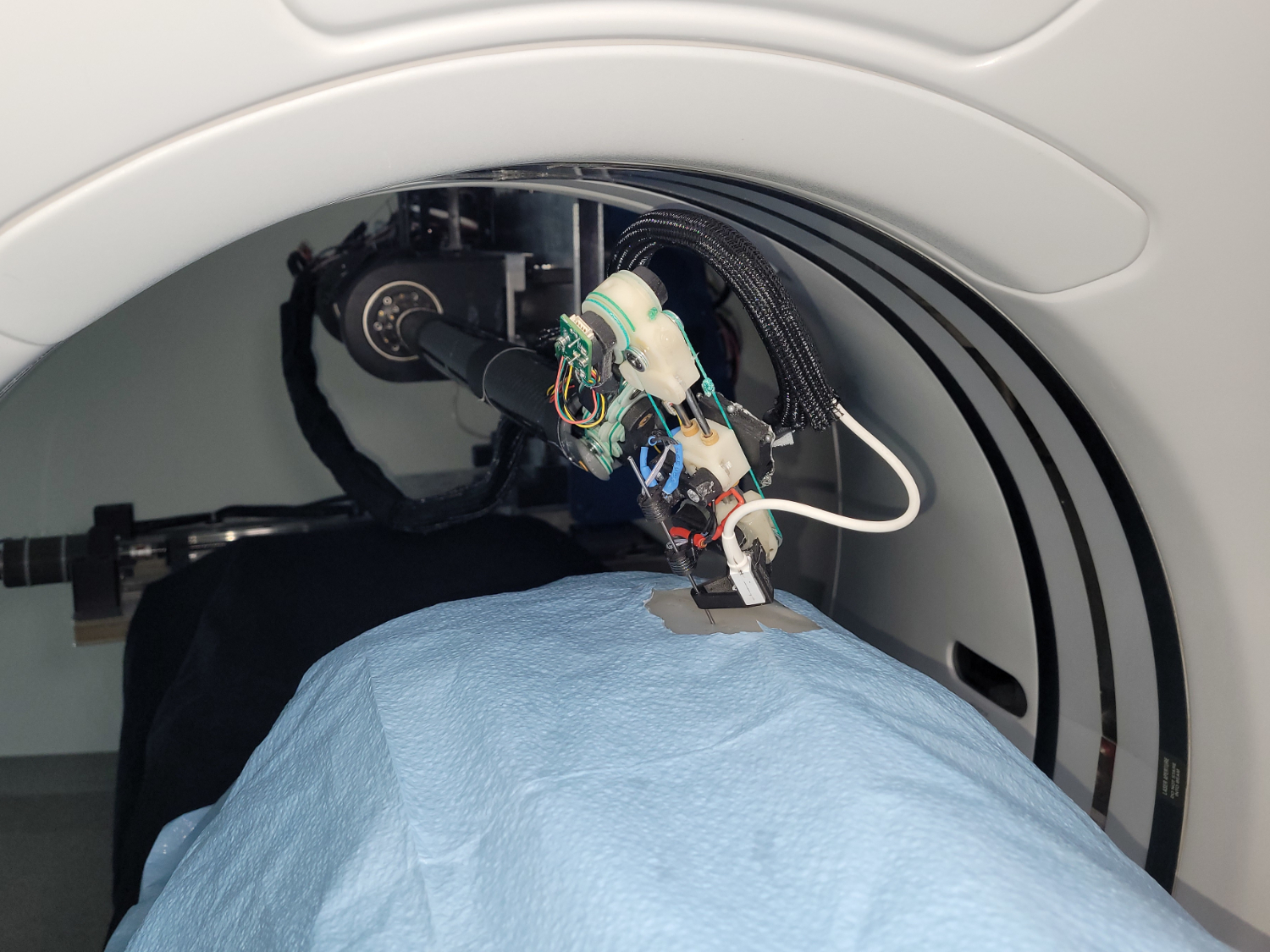}
    \caption{Needle insertion within imaging bores provides direct volumetric visualization of the anatomy and tool, improving the accuracy of needle insertion procedures such as retroperitoneal biopsy and lumbosacral spine nerve block. However, this enclosed environment limits the space for manipulation and line-of-site visibility for devices. CRANE overcomes these challenges with its cable-driven serial link design and integrated planning-control method to enable fully in-bore dexterous needle manipulation without requiring manual setup.}
    \label{fig:cover_photo}

    \vspace{-3mm}
\end{figure}

Typically, CT-guided procedures involve multiple steps where the IRs alternate between manually and incrementally advancing the needle-like tool and stepping away from the gantry to scan the patient. 
%The multiple scans assist the physician in planning and controlling the insertion of the needle while compensating for physiological motion and avoiding anatomical obstacles on their needle insertion trajectory. 
The physician must balance frequency of scanning to understand the in-body environment with the radiation exposure for both the patient and physician. This trade-off directly affects tip-to-target accuracy  \cite{leng_radiation_2018},  \cite{yu_radiation_2009}. Frequently, the IR withdraws the patient from the imaging bore for needle insertion between scans for improved ergonomics. The combination of delayed feedback and freehand control presents a challenge when high accuracy is required and results in physicians performing a move-and-wait strategy \cite{ferrell_supervisory_1967} where they scan the patient, view the scan, make a conservative needle adjustment, and scan again.
  
Robotics can improve tip-to-target accuracy and ergonomics while decreasing radiation exposure by providing real-time visualization of the system state to avoid many of the current move-and-wait challenges and allowing for the precise adjustment of tools while the patient is within the bore  \cite{cornelis_comparison_2015, levy_clinical_2021}.

Many robotic platforms have been developed for transabdominal and trans-thoracic access, where the robot must ultimately operate well within the CT bore. Existing robotic platforms have demonstrated good needle placement accuracy, including small-size systems with low weight and large platforms with a fully active workspace and numerous methods of inserting needles  \cite{unger_robot-assisted_2021, siepel_needle_2021}, and a variety of levels of human-involvement required during the procedure. 
However, existing systems fail to offer automatic hands-free device setup, control, and needle insertion within the imaging bore, resulting in performance compromises, the need for clinical workflow changes with increased overhead, the loss of existing abilities, or decreased safety in some situations \cite{arnolli_overview_2015, safaee_techniques_2016, schulz_accuracy_2013, ben-david_evaluation_2018, schulz_accuracy_2013, beermann_1000_2018,  Maurin2006}. %patient straps, jet ventilation w/ anesthesia, manual insertion
This is partly due to limited exploration of the clinical considerations at the intersection of planning, design, and control for a large active workspace and high dexterity robot within the imaging bore, and \textcolor{black}{the lack of comprehensive in-bore design dexterity evaluation methods}.

In this paper, we present CRANE, shown in Fig \ref{fig:cover_photo}, which provides a large workspace with high dexterity and accuracy with a simple automated method for in-bore robot control. {Specifically, we tackle the challenge of designing a low-profile robot which has minimal backlash and high accuracy with sufficient dexterity and low artifact-generating metallic components, how to automate dexterous robot setup within an imaging bore for a needle insertion problem, how to evaluate a design early on in the development phase, and how to interface a robot with a needle-like tool such that both active insertion and passive guidance are supported}. We previously presented the mechatronic design of CRANE with teleoperated control \cite{schreiber_crane_2022}; this works extends from that \textcolor{black}{teleoperated} approach and investigates the development and integration of an automated workflow, including analysis of the design's dexterity, planning, and control for hands-free in-bore needle insertion. The \textcolor{black}{novel} contributions of this paper are as follows:

\begin{enumerate}
    \item \textbf{Robot design for in-bore dexterity} – a low profile and redundant serial link robot design \textcolor{black}{utilizing outside-of-bore actuation with serial cable-driven in-bore joints and redundant sensing; achieving minimal in-bore footprint, dexterity to insert needles across a patient body, a large active workspace, passive backdriveability, and clinically required accuracy.} %evaluated} across clinical and synthetic cases using novel metrics and achieving high accuracy via multi-level closed-loop control.

    \item \textbf{Framework for in-bore planning and control of redundant robot} – unified framework incorporating physician-specified task prioritizing for control of a redundant robot enables automated device setup operation. This framework is used for simulated design evaluation and control synthesis in-situ experiments.

    \item \textcolor{black}{\textbf{Comprehensive robot dexterity evaluation} - a clinically representative in-bore robot manipulation dexterity evaluation is developed, including clinical and synthetic cases. This method can help improve robot design to maximize the possible needle insertion trajectories in patients, resulting in improved safety, ease of use, and a greater breadth of possible procedures with a single platform.}

    \item \textbf{Mechanism for grasping a needle} – a novel clutching needle gripper using Shape-Memory-Alloy (SMA) actuator enables deep needle insertion with a short overall needle insertion mechanism length via electromechanical clutching. \textcolor{black}{This mechanism allows for selecting active remote driving or physician-in-the-loop control where the system only provides insertion pose control without active insertion.}
\end{enumerate}

%Robots can eliminate the multiple punctures and procedures currently required while enabling physicians to treat small early-stage cancer via a minimally invasive approach. However, existing robotic platforms have a difficult and long setup, limited applicability and large size, imaging artifacts, insufficient accuracy, and limited needle/probe compatibility have limited their clinical application. 

Our robotic platform, CRANE, tackles these main limiting issues via its novel mechanical design coupled with planning method while retaining the accuracy of previous systems.
This provides the ability to work with patients of all sizes while automatically setting up the system, giving compatibility with off-the-shelf biopsy and ablation probes, and inserting the needle toward the target with only high-level control and minimal interaction.

\vspace{-2mm}
\section{Clinical Procedure: Overview, Considerations, and Related Works} \label{section:clinical_overview}
Surgical robots must carefully balance clinical value with workflow changes. Surgical robotics using image guidance (such as CT or MR) has a long history, including many of the first applications of robots within healthcare  \cite{george_origins_2018,khanna_path_2021}. However, image-guided robotics have only achieved strong clinical presences within orthopedic and neuro-surgery \cite{li_clinical_2021}, where procedures are several hours long, and the systems operate outside imaging bores with less significant space constraints. Within Interventional Radiology, navigation systems that track the needle's base position providing real-time needle visualization have received higher adoption due to their simple setup \cite{noauthor_intuitive_nodate, noauthor_quality_nodate} and have demonstrated improvements in clinical outcomes \cite{khanna_path_2021, wallach_comparison_2014, beermann_1000_2018, widmann_targeting_2011, bale_stereotactic_2019, schullian_frequency_2021, widmann_frameless_2012}. However, these systems' performance is limited within the imaging bore \cite{arnolli_overview_2015} and have decreased clinical accuracy in comparison with robotic approaches \cite{van_baarsen_itvt-07_2021}. Recent work presented many challenges relating to decreased adoption of image-guided robotics outside of orthopedic and neurosurgery \cite{fichtinger_image-guided_2022} with a strong focus on the procedural workflow changes required to integrate the device overcoming their added value.

\subsection{Overview and Considerations}
Imaging scanners, such as a CT scanner, comprise an imaging gantry and a couch. The imaging gantry has a bore that the patient resides within during imaging. 
%This bore can be a physical bore, such as in a CT or MR scanner, or a virtual constraint, such as in a C-arm fluoroscopy scanner or a medical linear accelerator. The patient rests on a couch consisting of a base and a table. The imaging gantry is typically stationary for CT scanners, and the tabletop of the couch moves in and out of the bore. 
The area between the patient, patient table, and the scanner bore is limited, placing in-bore space at a premium. 
%\textcolor{blue}{Withdrawing the patient from the scanner provides more working room. However, this lengthens the procedure time and raises the chance of a mismatch between the captured image and the patient's current state due to physiological motion or shifting. Working within the bore enables shorter, more precise procedures but reduces physician ergonomics due to the limited space. Even then, they must remove their hands and body from the bore between intermittent scanning to avoid radiation. Surgical robot platforms can enable more ergonomic manipulation of surgical instruments within the imaging bore and, with the appropriate design, even increase the space available versus a physician by providing a lower profile.}% The device design affects all stages of the procedure.
An image-guided percutaneous surgical procedure has three main phases: Preoperative Setup, Procedure Planning, and Procedure Execution.

\setlength{\tabcolsep}{5.5pt}
\begin{table*}[!b]
    \centering
    \caption{\textit{Fully Actuated} Robots for In-bore Abdominal Percutaneous Interventions}
    \label{table:comparison_robotic_platforms_inbore_needle}
    \begin{center}
    \begin{tabular}{llm{10mm}m{12.5mm}m{12.5mm}m{15mm}m{12.5mm}lm{15mm}}
    \toprule
    Project & Mounting & Active DoF & Active Pos. Space & Active Ori. Space & Needle Insertion Method & In-bore Volume & Control Method & Manually Positionable\\\midrule
    %Korean group floor mounted, netherlands group table mounted, maxio and japanese one I'm not sure if they go in the bore?
    %DDD & Floor & 6 & Large & Large & Passive & Large & IK w/ col. detect. & \cite{}  \\\hline
     %MDPI & Floor & 5 & Large & Large & Prismatic & Medium & IK & \cite{gherman_risk_2022}  \\\hline %weird robot...
    %  \cite{won_validation_2017} & Floor & 5 & Large & Large & Passive & Medium & IK & No  \\
    ZeroBot \cite{Hiraki2017, komaki_robotic_2020} & Floor & 6 & Large & Large & Prismatic & Large & IK & No  \\
    \cite{tsekos_prototype_2005, tsekos_general-purpose_2008, NVTsekos2007, christoforou_performance_2007, ozcan_interconnection_2008} & Floor & 7 & Large & Large & Prismatic & Low & IK w/ col. detect. & No  \\
    Acubot/PAKY \cite{stoianovici_acubot_2003, solomon2002robotically, challacombe_randomized_2005, d_modular_1998, masamune_system_2001, shah_robotically_2008, cleary_precision_2005} & Couch & 6 & Medium & Medium & Prismatic & Low & IK & Yes   \\
    I-Sys/B-Rob II \cite{schulz_accuracy_2013, martinez_ct-guided_2014} & Couch & 5 & Small & Large & Passive & Low & IK & Yes \\ % , song_robotic_2011, kettenbach_robot-assisted_2005
    Innomotion \cite{melzer_innomotion_2008, hempel_mri-compatible_2003, zangos_mr-compatible_2011} & Couch & 6 & Medium & Medium & Prismatic & High & IK & Yes  \\
    Open 7-DoF  \cite{Schreiber2019} & Couch & 7 & Medium & Medium & Prismatic & Low & IK & No  \\
    LPR  \cite{ozcan_interconnection_2008} & Patient & 5 & Medium & Low & Clutching & High & IK & No  \\
    CTBot \cite{Maurin2006, maurin_ctbot_2005, maurin_patient-mounted_2008, piccin_force_2009} & Patient & 5 & Low & Low & Clutching & Medium & IK & No  \\
    XACT \cite{ben2018evaluation} & Patient & 5 & Low & Medium & Gear & Low & IK & Yes  \\
    CRANE (ours) & Floor/Couch & 8 & Large & Large & Clutch & Low & Automated Planning & Yes  \\
    %LPR & Patient & Medium & Low & Clutching & High & \cite{ozcan_interconnection_2008}  \\\hline
\bottomrule
\end{tabular}
\end{center}
\end{table*}

\begin{enumerate}
\item \textbf{Preoperative Setup Phase} – Before the surgery, the physician reviews the preoperative images to determine a rough plan of the procedure approach, including patient positioning (e.g., prone, supine, left or right lateral recumbent). %These images are typically from initial diagnostic imaging scans (e.g., CT, MR) used to diagnose the existence of an ailment that requires an intervention. Frequently, these scans are from days, weeks, or months prior.
If using a robot, the device is registered to the CT scanner coordinate system using an image-based alignment method so that the physician may physically operate in the scanner's coordinate system.

\item \textbf{Procedure Planning Phase} – With the patient in position, the physician takes an initial volumetric scan to provide an accurate view of the patient's anatomy with which they can plan their precise needle-insertion approach. 
%Procedures require planning a trajectory for the needle through the body, from which an initial needle insertion point is selected. 
They must consider the in-body obstacles (e.g., bone, blood vessels, other organs) and the confined space they are working in (e.g., imaging bore, patient) to avoid collisions between the environment and the needle. %Typically, this step is performed within the image captured via the medical scanner. 
%The physician can select their preferred visualization (2D slices or 3D rendering) within the scanner software during this step. 
With a robot, an additional challenge involves setting up the robot to the planned trajectory, assuming it is reachable and stable concerning the kinematics of the robot.  

\item \textbf{Procedure Execution Phase} – The physician inserts the needle into the patient towards the target point within the patient's body. 
The physician alternates between making minor adjustments and inserting the needle with intermediate control scans. Upon complications or after multiple failures of attempting to navigate the needle to the target, the physician may loop back to \textbf{Step 2 - Procedure Planning Phase}, determine a new needle insertion trajectory and begin again. 
\end{enumerate}
\vspace{-2mm}

\subsection{Related Works for Percutaneous Robotic Needle Insertion}

\setlength{\tabcolsep}{8pt}
\begin{table}[t!]
    \caption{Technologies for Needle Insertion}
    \label{table:technology_system_comparison}
    \begin{center}
    \color{black}\begin{tabular}{lll}
    \toprule
    Category & Technology & References \\ \midrule
    \multirow{3}{*}{Kinematic Design} & Floor & \cite{cornelis_comparison_2015, won_validation_2017, hiraki_robotic_2018, komaki_robotic_2020, Yang2010a}\\ %tovar-arriaga_development_2011, ebert_virtobotmulti-functional_2010, ebert_virtobot_2014
    %& &\cite{noauthor_performance_nodate, gherman_risk_2022, tsekos_prototype_2005, tsekos_general-purpose_2008, NVTsekos2007, christoforou_performance_2007, ozcan_interconnection_2008, tovar-arriaga_development_2011, won_validation_2017}   \\
    & Couch & \cite{d_modular_1998, schulz_accuracy_2013, masamune_system_2001, Schreiber2019} \\ %  groenhuis_22_2020, goos_new_2000, hempel_mri-compatible_2003  melzer_innomotion_2008}  %stoianovici_acubot_2003, Kettenbach2014, cleary_precision_2005, shahriari_computed_2017, zangos_mr-compatible_2011}    \\ 
    & Patient & \cite{Maurin2006, hungr_design_2016, walsh2007evaluation, wu_remotely_2019, Yang2017, n_body-mounted_2021, walsh_evaluation_2009, number6, Lim2019} \\ \midrule %maurin_ctbot_2005, maurin_patient-mounted_2008, 
    \multirow{2}{*}{Dexterity Analysis} & In-body & \cite{belbachir_automatic_2018}   \\
    & Out-of-body & \cite{Schreiber2019}   \\ \midrule
    \multirow{3}{*}{Active Needle Interface} &  Fixed Prismatic & \cite{Yang2010a, Schreiber2019}  \\
    &  Clutching &  \cite{frishman2021, ghelfi_evaluation_2018, bricault_light_2008, Frishman2020}  \\
    &  Roller & \cite{seitel_development_2009, shahriari_design_2015}  \\ \midrule
    \multirow{2}{*}{Needle Trajectory} & Manual & \cite{won_validation_2017, hiraki_robotic_2018, komaki_robotic_2020, Yang2010a, Maurin2006, hungr_design_2016, walsh2007evaluation, wu_remotely_2019, Yang2017, n_body-mounted_2021, walsh_evaluation_2009, number6, Lim2019}\\
    & Automated & \cite{seitel_computer-assisted_2011, pourarab_dynamic_2020, hamze_evolutionary_2017, garg_exact_2014, Lee2012, chen_optimizing_2009, Tsumura2018, Shamir2010, ren_treatment_2014, alterovitz_constant-curvature_2008, niyaz_following_2020, shahriari_computed_2017, duindam_3d_2010, van_den_berg_lqg-based_2011, xu_motion_2008, patil_interactive_2010} \\ \midrule
    \multirow{2}{*}{Robot Trajectory} &IK & \cite{won_validation_2017, hiraki_robotic_2018, komaki_robotic_2020, Yang2010a, Maurin2006, hungr_design_2016, walsh2007evaluation, wu_remotely_2019, Yang2017, n_body-mounted_2021, walsh_evaluation_2009, number6, Lim2019}\\
    &Path Planning & \cite{ebert_automatic_2016, liu_automatic_2016, Liu2019}\\
    %\multirow{1}{*}{System-off Gravity Force} & End-effector & $<1.0N$\\ \hline
    % \multirow{3}{*}{Joint Acceleration} & Needle & $<1.0$ \space N\\
    % &In-bore joints & $<0.35$ \space N/m\\
    % &Linear joints & $<60N$\\ \hline
    % \multirow{3}{*}{Joint Velocity} & Needle & $<1.0$ \space N\\
    % &In-bore joints & $<0.35$ \space N/m\\
    % &Linear joints & $<60N$\\ \hline
\bottomrule
\end{tabular}
\end{center}
\vspace{-3mm}
\end{table}

This related work focuses on several device design attributes significantly affecting clinical workflow (e.g., device setup, control), value-add (e.g., accuracy, procedure time, general-purpose applicability), and safety (e.g., failure mode, device energy). The primary focuses are the device's kinematic structure and workspace, needle interface, and automation level for planning and control. 
%User interface, device control scheme, device kinematic structure, and needle interface method significantly affect clinical application and efficacy. 

We discuss and compare works focused on these attributes with multi-organ applicability in the thoracic and abdominal regions. Table \ref{table:comparison_robotic_platforms_inbore_needle} provides comparison of several fully-active systems which can adjust the needles angle and position. 

\subsubsection{Kinematic Design: Device Setup, Workspace, and Safety}
The surgical robot platform's mounting location (e.g., floor, scanner couch, or patient) sets many constraints on the device's kinematic design, size, and mass. This affects the device's setup, workspace, and safety. Setup is frequently time-consuming and challenging and can require restarting from the beginning due to device self-collisions and environment collissions\cite{george_origins_2018, khanna_path_2021, lieberman_bone-mounted_2006}. Large workspace fully-active systems provide many advantages, including lower procedure setup complexity, remote teleoperation, and the ability to regulate cartesian space tip stiffness to limit tissue damage in the case of gross or respiratory patient motion \cite{Kim2017, Moreira2015}.

%floor mounted systems
Floor-mounted platforms (e.g., Innomotion, Maxio, Virtobot) are frequently fully active and simple to set up with large active workspaces. However, they can pose a safety risk in the case of relative motion between the patient on the scanner couch and the robot platform \cite{cornelis_comparison_2015, won_validation_2017, hiraki_robotic_2018, komaki_robotic_2020, Yang2010a, tovar-arriaga_development_2011, ebert_virtobotmulti-functional_2010, ebert_virtobot_2014, noauthor_performance_nodate, gherman_risk_2022}. \textcolor{black}{\cite{tsekos_prototype_2005, tsekos_general-purpose_2008, NVTsekos2007, christoforou_performance_2007, ozcan_interconnection_2008} present a design with a fairly low in-bore profile using remote actuation via telescoping shafts and bevel gears with minimal metallic components, achieving MRI compatibility. However, the design suffers from high backlash (upto $3^\circ$) and is non-backdriveable.}
Frequently, these designs use industrial robots with a custom end-effector (EE), \cite{tovar-arriaga_development_2011, won_validation_2017} which occupy significant space within the scanner bore, have large inertia, and are not passively backdrivable. This limits the ability of the device to perform in-bore needle insertion, to adjust for unexpected patient motion safely, results in high energy within the system during robot motion, and is difficult to remove in case of system failure. Therefore, the needle insertions are typically performed outside the scanner bore and with the patient under general anesthesia with decreased accuracy, longer procedure time, higher risk of complications, and higher cost.

% couch-mounted systems
Couch-mounted systems (e.g. Acubot/PAKY, I-Sys/Biorob) \cite{d_modular_1998, schulz_accuracy_2013, masamune_system_2001, Schreiber2019, groenhuis_22_2020, goos_new_2000, hempel_mri-compatible_2003, melzer_innomotion_2008, stoianovici_acubot_2003, shahriari_computed_2017} may be fully active or partially active with the use of setup joints. These devices are attached to the couch and passively move with the patient during couch motion for intraoperative scanning. However, scanner attachment risks system damage and patient injury in the case of improper fixation \cite{noauthor_class_nodate}. \textcolor{black}{\cite{melzer_innomotion_2008} provides a fully active and backdriveable system with low backlash. However, the system occupies significant space ($>70mm$ cross section) and has limited joint travels.} Manual setup joints significantly increase device adjustability while retaining a low profile and stiffness without backlash but are challenging to use and preclude the option of a fully automatic device setup. Instead, the physician must carefully manipulate the device to the correct position by hand while considering potential collisions with the environment. 
Many devices in this class use small, highly geared, non-backdriveable motors, which limits their dynamic performance and requires patients to be under general anesthesia.

%patient mounted systems
Patient-mounted systems (including Light Puncture Robot, XACT, and Robopsy) are small and intrinsically compensate for gross patient and table motion improvement \cite{maurin_ctbot_2005, maurin_patient-mounted_2008, hungr_design_2016, walsh2007evaluation, wu_remotely_2019, Yang2017, n_body-mounted_2021, walsh_evaluation_2009, Lim2019}. However, they frequently occupy more space in the scanner and possess less dexterity. 
Balancing system stability and patient comfort is also challenging when mounting to patients.
%\subsubsection{Workspace Analysis}

CRANE's fully active workspace and redundant joints remove the requirement for manual device setup to a needle insertion transform while providing dexterity for needle insertion within a scanner bore. The transmission enables a low profile in the scanner bore, low backlash, low inertia, and backdriveability. This maximizes the space within the scanner for the patient, enables high accuracy and more straightforward device control, and improves procedural safety.

Throughout these device designs, in-bore workspace and dexterity analysis focus on swept volume accessible by the robot's tip in free-space \cite{Maurin2006, hungr_design_2016, Yang2017, Jiang2017}, potentially with a single reference human body size \cite{wu_remotely_2019} or the ability to reach target organs and tumors in specific patient cases \cite{belbachir_automatic_2018}. \cite{Schreiber2019} used orientable needle insertion evaluation on a demonstration human mesh surface as a proxy for in-body reachability. These evaluations have limited general-purpose applicability and do not consider dexterity in patients with large body habitus.

Here, we contribute a method and results for robot design analysis and planning to evaluate the general-purpose ability of a robot to perform dexterous in-bore needle insertion. Our method focuses on the area outside the patient: can the robot arm manipulate and insert a rigid needle-like tool into a patient's body? We consider the ability to dexterously manipulate the needle around the nominal insertion point to enable adjustment for disturbances that may occur during insertion. {Our analysis specifically focuses on a variety of clinically important metrics missing from prior approaches, including EE manipulability for task-space forces, torques, and position adjustments as required to advance the needle or adjust its base pose and to compensate for needle insertion error}. This analysis is developed based on considerations across several clinical cases from which we construct a general and exhaustive evaluation across simulated procedures.
\subsubsection{Needle Interface}
Needle interfaces assist a physician in inserting the needle into the patient while maintaining a target pose. They must be sterile, safe, and replaceable. Many needle interfaces exist, including passive guides  \cite{d_modular_1998, stoianovici_acubot_2003, melzer_innomotion_2008,bricault_light_2008} and active mechanisms  \cite{frishman2021, Yang2010a, Schreiber2019, shahriari_design_2015, seitel_development_2009, shah_robotically_2008, levy_clinical_2021}. Passive guides orient and position the needle at the surface entry point where the physician manually inserts the needle. They decrease the unintentional application of torques and forces to the needle, deflecting it and decreasing accuracy.

Active needle drivers enable precise and quick insertion while the physician remains remote and away from radiation or contagious diseases. Existing active needle driver designs have utilized numerous mechanisms including limited travel prismatic axis \cite{Yang2010a, Schreiber2019, shahriari_design_2015}, friction rollers \cite{seitel_development_2009, shahriari_design_2015}, clutching graspers \cite{frishman2021, ghelfi_evaluation_2018, bricault_light_2008, Frishman2020}, and rack-and-pinion \cite{levy_clinical_2021} while incorporating aspects such as force sensing and safety release  \cite{shah_robotically_2008}. \textcolor{black}{These active designs have high complexity and limited replaceability resulting in potentially challenging sterilization procedures and limited compatibility with different needles and probes while providing sufficient gripping force to insert them without slipping at $<10N$ \cite{walsh_image-guided_2010} or damaging the probe itself due to high Hertzian contact stresses\cite{s_robotically_2002}. Frequently, the designs require additional hydraulic or pneumatic lines which are difficult to interface on a disposable component}.

Our presented clutching needle insertion mechanism uses a simple, fully solid-state design with a flexure-based needle interface providing low-manufacturing cost and active gripping only when power is applied. This supports single-use with pre-sterilization, even gripping force, and  \textcolor{black}{support for both passive guidance with the ability for the physician to advance or remove the needle by hand for improved safety}.
 
\subsubsection{Automation Level - User Interface, Planning, and Control}
Device setup and user control methods significantly affect the \textbf{Procedure Execution Phase} and fall into three main categories: direct joint control, EE control, and automated. %This results in systems that are manually manipulated (patient attachment or robot's setup joints) by the user \cite{Christoforou2014, seitel_development_2009} to automatic setup \cite{guiu_feasibility_2021} with the robot determining and executing a path to a specified pose.
%Direct joint control

Direct joint control \cite{Christoforou2014, seitel_development_2009, boehm_current_2021} has the physician manually control the robot joints. The system helps a physician perform a static insertion trajectory and achieve fine adjustment. Stereotactic frames are examples of this in image-guided surgery today. However, they can be complex and unintuitive to set up and plan for due to their frequently unintuitive kinematics and joint limitations \cite{safaee_techniques_2016}.

%Teleoperated end-effector control
Teleoperated EE control  \cite{Boctor2004, Maurin2006, hungr_design_2016, Yang2017, Schreiber2019, maurin_patient-mounted_2008} via Inverse Kinematics (IK), either in robot coordinates or image frame, allows a physician to abstract away the device's joint motions and make precise needle adjustments. 
%Closed-form solutions exist for non-redundant manipulators and iterative methods for a redundant manipulator. 
This is the most common operating model today, especially for small workspace and under-actuated systems, which require manual setup for rough positioning and do not have to solve large-scale motion planning problems for device setup. However, with this approach, the responsibility of considering potential collisions with obstacles in the environment and limitations of the robot's dexterity is left to the physician operators, which is cognitively challenging to keep track of given the variety and complexity of linkage designs across different platforms. More broadly within the field of robotics, IK algorithms exist that locally consider collision avoidance, and device dexterity \cite{hollerbach1987redundancy, baerlocher_task-priority_1998, aghakhani_task_2013, haviland_neo_2021} but have not yet been applied to the problem of in-bore needle insertion. 
\textcolor{black}{A variety of techniques have been developed to minimize the effects of local minima on IK algorithms\cite{Book}, including multi-start gradient descent and simulated annealing. Additionally, specialized methods for global optimization of IK solutions considering obstacles\cite{SimplicialHomologyAlgorithm}. }%IK algorithms are local solutions prone to getting stuck in local minima within a constrained environment.

%Automated end-effector control: fine adjustments 
Automated EE control with planning directly in imaging space \cite{pojskic_initial_2021, guiu_feasibility_2021} enables the physician to select their target needle insertion trajectory from which they guide the system to a course alignment. Then the robot automatically performs the fine adjustment based on image tracking, decreasing the physician's cognitive burden and improving procedural accuracy \cite{van_baarsen_itvt-07_2021}. However, the physician is still responsible for the device setup for course alignment and environmental collision considerations.
%in-bore collission checking: a precollission system
Existing works have included methods for environmental and self-collision checking \cite{christoforou_performance_2007, tsekos_general-purpose_2008, tsekos_prototype_2005, maurin_patient-mounted_2008, huang_current_2021} %outside \cite{huang_current_2021} imaging bore
prior to a commanded motion. \textcolor{black}{However, the physician still must determine the collision-free robot configuration and trajectory them-self, in effect performing one of the most challenging parts of robot setup.}

%out of bore planning for paths, noone has done in bore?
Collision-free motion planning from image-space needle trajectory allows the physician to select their target needle insertion trajectory and then allows the robotic system to solve the associated motion planning problem to avoid self and environment collisions while supporting angulation adjustment during the procedure. This is most relevant in large fully-active workspace systems, which can perform automated setup, allows the physician to focus on the higher-level surgical tasks, and decreases the cognitive overhead of introducing a robotic system. Ideally, the physician is confident in the system's ability to follow through on this task regardless of the specific patient and needle insertion location. Previous works on collision-free planning for image-guided needle insertion provide methods of planning for nominal trajectory insertions outside the bore \cite{ebert_automatic_2016, liu_automatic_2016, Liu2019} or in simulation within a scanner bore \cite{belbachir_automatic_2018} without considering dexterity required for needle pose adjustment around the nominal insertion pose for device kinematics or collision constraints with the imaging bore.

%automated needle path angulation planning
Significant research exists for assisting physicians in determining optimal trajectories for needle insertion within the patient's body during the \textbf{Procedure Planning Phase} to perform the primary surgical task (e.g., biopsy, ablation, deep brain stimulation) while optimizing for a variety of objectives (e.g., obstacle avoidance, cranial-caudal needle angulation, needle deflection, tissue deflection, brachytherapy radiation distribution, ablation zone) from medical imaging data \cite{Scorza2021}. Many approaches exist for both straight \cite{seitel_computer-assisted_2011, pourarab_dynamic_2020, hamze_evolutionary_2017, garg_exact_2014, Lee2012, chen_optimizing_2009, Tsumura2018, Shamir2010, ren_treatment_2014} and steerable needles \cite{alterovitz_constant-curvature_2008, niyaz_following_2020, shahriari_computed_2017, duindam_3d_2010, van_den_berg_lqg-based_2011, xu_motion_2008, patil_interactive_2010}. %These methods primarily focus on determining a needle insertion trajectory within the body rather than on a method of performing the insertion with a robot. %Several works focus on determining a curved path that satisfies the steerable needle's non-holonomic constraints while retaining reachability and controllability within the patient body \cite{}. 
A select few of these methods integrate this in-body needle trajectory planning with robot planning for out-of-bore insertion \cite{liu_automatic_2016, Liu2019, barkhordari_robot_2018} or within a simulated environment for in-bore \cite{belbachir_automatic_2018}. 
%our planning / control method compared to others

\textcolor{black}{Our robot configuration optimization method automates the robot setup and control given a target straight-line needle insertion trajectory specified by the physician while considering kinematic robot dexterity and environment collisions around the nominal needle insertion pose. This removes a significant cognitive consideration from the physician's procedure planning}

\vspace{-1mm}

\subsection{Design Requirements and Specifications}

\textcolor{black}{For our system, we define the following design requirements and specifications. Our system performance, actuation, and sensors are summarized in Table \ref{table:system_comparison}.}
\begin{enumerate}
    \item \textcolor{black}{Forces: inserting and adjusting the needle requires up to $10N$ of axial force and $0.05Nm$ of torque \cite{Walsch_thesis_2010, Poniatowski2016, Walsh2008}.}
    \item \textcolor{black}{Passive Failure: system is hand back-driveable and applies $<1N$ of force when the system powered off.}
    \item \textcolor{black}{Precision: insertion requires $<4mm$ position and $<4^\circ$ orientation error \cite{durand_computer_2017}. Our system backlash should be significantly less than $<4mm$ position and $<4^\circ$ to provide fine adjustability without chatter.}
    \item \textcolor{black}{Workspace: Able to reach across a patient's abdomen within a $70cm$ scanner bore, inserting up to $140mm$\cite{number15} deep with a short insertion joint length, $<70mm$ max cross-sectional dimension to preserve patient space, and adjusting orientation up to $60^\circ$ cranial-caudal and medial-lateral.}
    \item \textcolor{black}{Needle Interface: the needle interface should be easily replaceable, and able to function in both an active driving approach and passive guide modes.}
    %\item \textcolor{black}{Registration Accuracy: <1mm Target Registration Error such that the robot's needle can be accurately guided to the target within the imaging frame}
    \item \textcolor{black}{Imaging Artifacts: high-density materials (e.g. metals) cause reconstruction artifacts in CT images, reducing the physician's ability to perform to see the target anatomy and perform the procedure. Actuators must be outside the bore or non-metallic.}

\end{enumerate}

%This problem is solved hierarchically. First, a dexterous robot configuration for a provided straight-line needle insertion within an imaging bore based on weighted metrics as the optimization objective. The nominal robot configuration is determined to retain dexterity while utilizing the system's redundancy to satisfy physician preferences, including obstacle avoidance with obstacle priorities. Then, a collision-free trajectory is determined from this setup pose to the needle insertion pose. Finally, a controller modulates the joint positions to remain within this validated dexterous workspace.% with redundancy resolution to minimize joint angle distance from this nominal configuration via null-space projection.

%\textcolor{black}{ADD SUMMARY OF THE SECTION HERE???}
\setlength{\tabcolsep}{8pt}
\begin{table}[t!]
    \caption{CRANE's design specifications highlighting the dexterity, accuracy, and safety}
    \label{table:system_comparison}
    \begin{center}
    \begin{tabular}{lll}
    \toprule
    Specifications & Category & Values \\ \midrule
    \multirow{2}{*}{Degree of Freedom} & Positioning Joints & 8 Dof   \\
    & Needle Gripper & 2 Dof   \\ \midrule
    \multirow{1}{*}{Needle Insertion Mechanism} &  Length & $180 mm$  \\
    \multirow{2}{*}{Accuracy} & Positional & $<1mm$\\
    & Angular & $<1^{\circ}$ \\
    \multirow{2}{*}{Communication} & Embedded & UDP \\
    & Desktop & TCP/IP \\ \midrule
    \multirow{3}{*}{Sensing} & Motors & Opt. encoder\\
    & In-bore joints & Magn. encoder \\
    & End effector & Magn. tracker\\ \midrule
    \multirow{3}{*}{Sensor resolution} & Opt. encoder & $<0.01^{\circ}$\\
    &Magn. encoder & $<0.1^{\circ}$\\
    &Magn. tracker & $<0.1mm$\\ \midrule
    \multirow{3}{*}{Backdriving force} & Needle & $<1.0N$\\
    &Trunnion joint & $<0.91Nm$\\
    &In-bore joints & $<0.35Nm$\\
    &Linear joints & $<60N$\\
    %\multirow{1}{*}{System-off Gravity Force} & End-effector & $<1.0N$\\ \hline
    % \multirow{3}{*}{Joint Acceleration} & Needle & $<1.0$ \space N\\
    % &In-bore joints & $<0.35$ \space N/m\\
    % &Linear joints & $<60N$\\ \hline
    % \multirow{3}{*}{Joint Velocity} & Needle & $<1.0$ \space N\\
    % &In-bore joints & $<0.35$ \space N/m\\
    % &Linear joints & $<60N$\\ \hline
\bottomrule
\end{tabular}
\end{center}
\vspace{-3mm}
\end{table}
\section{System Design}
The CRANE platform design focuses on dexterity, accuracy, and safety through its fully active design with precise transmissions, redundant sensors, and a fail-passive design (illustrated in Fig. \ref{fig:experimental_setup_benchtop_system_evaluation} and with system specifications in Table. \ref{table:system_comparison}). %CRANE meets the clinical specs described above (Table \ref{}), leveraging its unique redundant design. 
In the following sections, we will describe the analysis and modeling of the mechanics of the design, as well as supporting electrical and software system architecture.

\subsection{Mechanical Design for Dexterity and Precision}
\begin{figure}[!t]
\centering
\includegraphics[width=\linewidth, trim={0cm 0cm 0cm 0cm},clip]{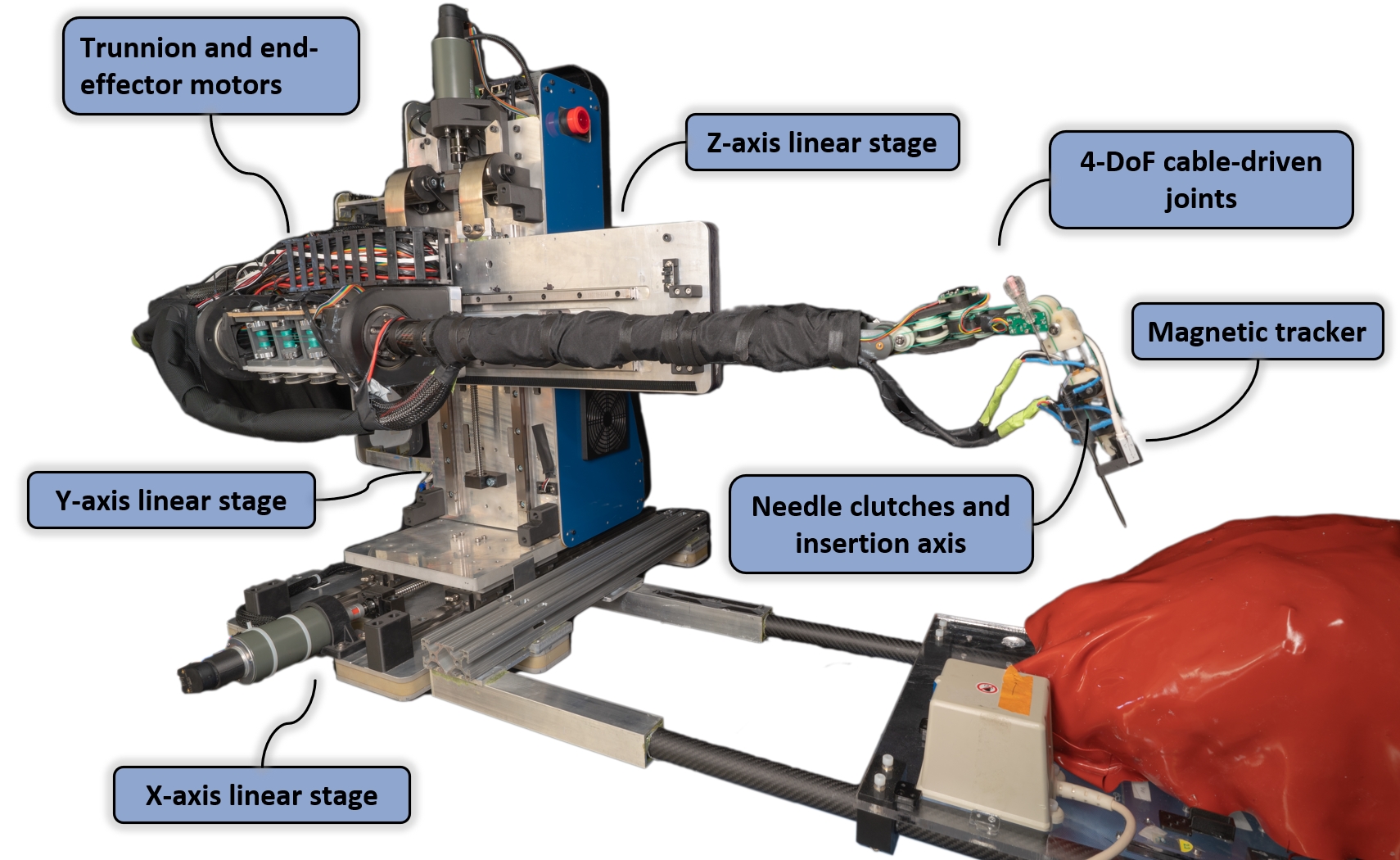}
\caption{\textcolor{black}{Experimental setup for bench-top system evaluation, highlighting the robot platform: the gross positioning stage enables the platform's large workspace and houses the actuators for the in-bore joints, and the redundant cable-driven in-bore joints enable orientation control and reaching around obstacles. The clutching needle insertion axis allows deep needle insertion with a short robot stroke. Together, this provides CRANE with a large workspace and the capability to perform dynamic motions while remaining backdriveable and having a minimal in-bore cross-section.}}
\vspace{-2mm}
\label{fig:experimental_setup_benchtop_system_evaluation}
\end{figure}

\textcolor{black}{Needle entry point control for insertion requires manipulation with three Cartesian position constraints and two Cartesian orientation constraints (excluding roll around the needle's primary axis)\cite{dimaioNeedleSteeringModelBased2003, glozmanImageGuidedRoboticFlexible2007a, cowanRoboticNeedleSteering2011, VanDeBerg2015}. The needle is advanced then along a typically linear trajectory with minor base position and orientation adjustments\cite{guptaCTGuidedInterventionsCurrent2014} along this constrained pose. 
%Physicians use a variety of techniques for advancing the needle, including gripping from the base with their fingers and pushing by hand\cite{guptaCTGuidedInterventionsCurrent2014}, gripping the shaft using a forceps to avoid radiation exposure for their hand\cite{heckAccuracyComplicationsComputed2006}, and using a mallet to provide fine position adjustment while overcoming dense tissue and stiction without overshoot\cite{mauzoEffectsUseMallet2018}. 
A primary challenge is selecting the correct position and orientation for straightline needle insertion \cite{walshSmallerDeeperLesions2011}. Furthermore, mechanical guidance increases accuracy over a real-time visualization of the needle's trajectory \cite{wallachComparisonFreehandnavigatedAiming2014}}. Therefore, the device must have at least 5 degrees of freedom (DoF). Additionally, the robot's joints and links must avoid obstacles (e.g., scanner gantry, patient, self-collision) while inserting and manipulating the needle. CRANE has 8-actuated joints for a 5-DoF problem space with three redundant joints, including a dedicated insertion joint. This enables CRANE to reach around the patient's body to insert a needle while avoiding collisions and singularities.%. The final joint is dedicated for needle insertion, leaving the remaining 7-actuated joints for dexterous needle manipulation, providing 2-redundant-DoF. 
\textcolor{black}{The system maximizes in-bore dexterity and active workspace while retaining a low profile by splitting the mechanics into two subsystems shown in Fig. \ref{fig:experimental_setup_benchtop_system_evaluation}: a large workspace out-of-bore base and a small intra-bore end-effector with remote actuation.} The gross positioning stage is described below, and the end-effector will be further described in the following section. 

%\textcolor{blue}{The mechanical design maximizes in-bore dexterity and active workspace while retaining a low profile with a smooth and low backlash transmission. The system accomplishes this by splitting the mechanics into two subsystems shown in Fig. \ref{fig:experimental_setup_benchtop_system_evaluation}: a large workspace out-of-bore base and a small intra-bore end-effector. The gross positioning stage is active, resides outside the scanner bore, and provides large-scale linear motion enabling automated device setup. The end-effector provides orientation control and final needle insertion within the tightly constrained space between the scanner bore and patient. 
%Highly efficient transmissions with low gear-ratio motors enable precise motion and backdriveability. The in-bore joints are cable driven for a low profile within the scanner driven by remote low gear-ratio low-backlash coreless motors outside the bore. The gross positioning stage is described below and the end-effector will be further described in the following section.}

The gross positioning stage comprises serially linked linear stages providing X-Y-Z cartesian motion with a large workspace ($400mm$ in each axis), high precision, low friction, and low inertia with constant-force spring for Z-axis gravity compensation. Further details on the gross positioning stage are provided within \cite{schreiber_crane_2022}. %\textcolor{blue}{The linear stages utilize ballscrews (Misumi BSST1204 with 4mm pitch, Misumi C-BSS1210 with 10mm pitch), coreless motors (Maxon 2260.886-51.216-200) with high-resolution encoders (US Digital E5-5000), and recirculating ball linear rails (Misumi SSXW33) with a composite foam-aluminum sandwich structure providing high stiffness and damping. The gross positioning stage's Z-axis (joint 2) is gravity counterbalanced via constant force springs (McMaster 9293K69). The first three joints can achieve linear velocities of 0.166 m/s, 0.166 m/s, and 0.332 m/s, respectively. Stiffness and backlash are measured at $>100\text{N/mm}$ and $<0.1\text{mm}$ using a dial indicator (Shars, 0.0005") and Z-style load cell (500N range).}

%Additionally, this decreases imaging artifacts and inertia versus joint-mounted actuators while providing the required forces and torques. 

\subsubsection{End-effector}
The distal end-effector enables dexterous needle manipulation within the tightly constrained space between the scanner bore and the patient. The end-effector has 5-DoF: 4 revolute joints for orientation control and a final needle insertion mechanism (described below) with a prismatic joint. $2N$ cable transmissions couple the proximally located motors to the distal 4-DoF in-bore joints through a thin tube. 
%The 4-DoF in-bore joints comprise three revolute joints and the prismatic needle insertion joints. 
Through a series of cable-driven joints, the motor volume and weight for the last DoF are isolated from the joints themselves, enabling a more compact design and thus greater workspace coverage, while low-gear-ratio motors can then be used to provide low reflected inertia to enhance backdriveability. %\textcolor{blue}{The motors for the in-bore joints of the end-effector are housed within a rotating trunnion. The trunnion provides the end-effector's 5th-DoF and is actuated via a belt drive coupling with a hollow shaft for electrical wire pass-through.}
%Idler pulleys (Fig. \ref{fig:kinematic_diagram}) are placed coaxially with actuated joints to reroute cables to further distal actuated joints to maintain pulley-cable contact and minimize cable loop length change throughout the joint range. 
%Gravity torques are minimal ($\sim0.01Nm$) due to the link's low masses. %\textcolor{blue}{Fiber reinforced plastics and synthetic cables are utilized for the in-bore components to minimize deflection and artifacts in the scanner images.}

\begin{figure}[bt]
\centering
\includegraphics[width=\linewidth, trim={0cm 3cm 0cm 2cm},clip]{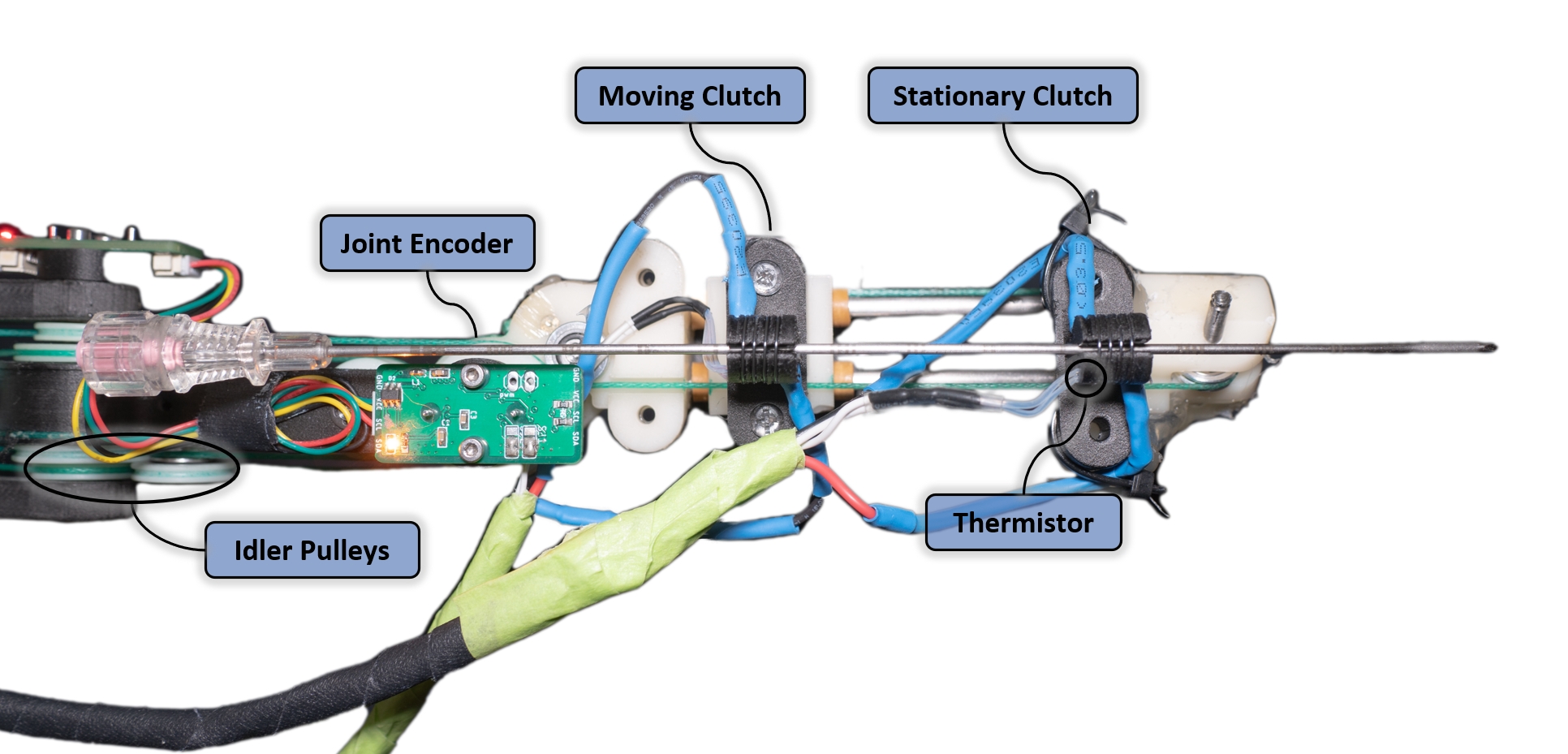}
\caption{\textcolor{black}{The robot end-effector is cable-driven. While cable drives provide low hysteresis and friction, they have limited stiffness, resulting in tracking errors. Joint mounted encoders enable direct sensing of the joints' position allowing controller compensation for cable stretch. The SMA actuated clutches temperature is sensed via thermistor, enabling closed-loop temperature control. The two moving and station clutches enable long travel active insertion or passive needle insertion with a simple and fail-passive mechanism which can be easily replaced for sterility and different size tools.}}
\label{fig:photograph_robot_endeffector_while_cable}
\vspace{-2mm}
\end{figure}

\subsubsection{Needle Insertion Mechanism}
Humans perform needle insertion stepwise, iteratively performing short insertions and grasping the needle. On CRANE, the needle insertion mechanism mechanically imitates this insertion strategy and enables deep insertion within a constrained space. The needle insertion mechanism includes the cable-driven prismatic insertion axis, two-needle clutches, and a guide bushing for precise long-travel needle insertion with a limited travel prismatic axis. The two needle clutches (Fig. \ref{fig:photograph_robot_endeffector_while_cable}) are placed coaxially. One functions as a brake to prevent needle motion. The guide bushing additionally serves as a mount for the Ascension trakSTAR magnetic tracker and the CT alignment marker used for end-effector pose sensing (described in Section \ref{section:alignment_method}).

\textcolor{black}{The needle clutch uses a Shape Memory Alloy (SMA) wire (Flexinol LT, $0.015"$) helically wrapped around a flexure and actuated via Joule heating. When $<40^\circ \text{C}$, the clutches is deactivated and acts as a guide. When heated $>80^\circ \text{C}$}, the wire contracts due to a crystalline structure change in the Nitinol from Martensite to Austenite. This length change applies a compressing the flexure into the needle. Experiments for measurements of slipping force, activation and cooling times, and lifespan were empirically tested (discussed in Section \ref{section:gripper_experiments}). 

\subsubsection{Statics Analysis}\label{section:statics}
\begin{figure}[bt]
\centering
\includegraphics[width=0.85\linewidth, trim={0 0.8cm 0 0.8cm}, clip]{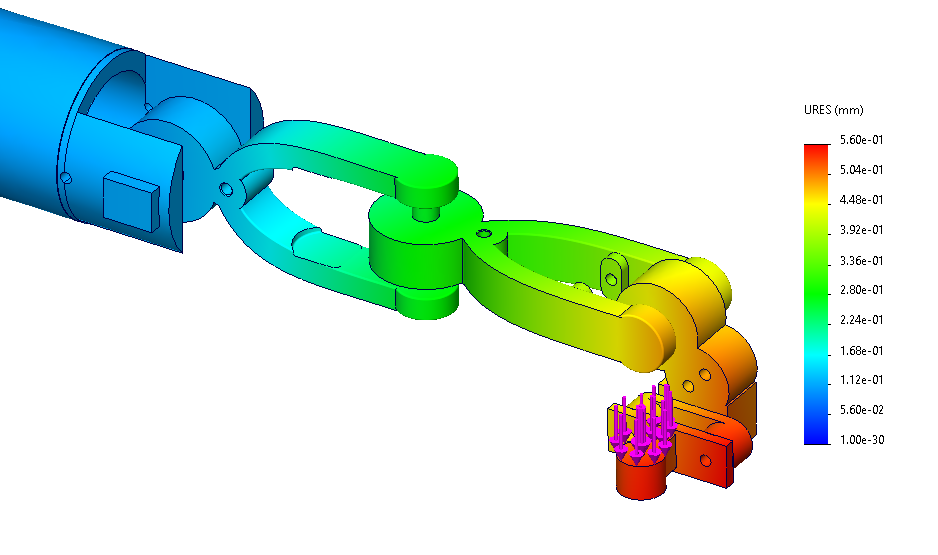}
\caption{\textcolor{black}{The long cantilevered tube for in-bore joints causes deflection illustrated by FEA modeling. This error, not observed by the joint encoders, is tracked and compensated for using a 6-DoF magnetic tracker and end-effector feedback control.}}
\label{fig:FEA}
\vspace{-3mm}
\end{figure}

Because of the long and thin arm section required to reach the patient within the scanner bore, stiffness analysis is performed. The EE's links are constructed from carbon fiber-reinforced plastic (CFRP). FEA analysis, illustrated in Fig. \ref{fig:FEA}, shows \textcolor{black}{$K_\text{link} = 1.79N/mm$} of tip deflection due to link deflection in a nominal configuration and a minimum $2\times$ factor of safety for all links.

Additionally, the cable transmission for the in-bore joints can exhibit deflection due to external forces and internal non-idealities, including friction, creep, and hysteresis. Static cable stretch is evaluated to determine its effect on system accuracy and stiffness. Cable stretch \cite{miyasaka_cable_2017} can be modeled as:

\begin{equation}
\Delta L = \frac{F L_{0}}{A E}
\end{equation}

\noindent where $\Delta L$ is the cable length change, $E$ is the cable's Young's Modulus, $L_{0}$ is the cable's nominal length, $F$ applied force to the cable, and $A$ is the cable's cross-sectional area. From this, angular deflection on cable-driven revolute joints can be calculated as:

\begin{equation}
    \Delta \theta  = \frac{\Delta L}{2\pi r}
\end{equation}
\noindent where $\Delta \theta$ is the joint angle change due to cable stretch and $r$ is the terminating capstan pulley's diameter. Transmission deflection due to cable-stretch results in \textcolor{black}{$K_\text{cable} = 0.80N/mm$} end-effector stiffness when evaluated at the configuration shown in Fig. \ref{fig:FEA}. 

Modeling these two components as series springs, the combined system stiffness at the given configuration is:
\begin{equation}
    K_\text{system} = \frac{K_\text{link} K_\text{cable}}{K_\text{link} + K_\text{cable}} = \textcolor{black}{0.55N/mm}
\end{equation}
%The end-effector deflection resulting from linkage and transmission deflection due to needle insertion forces motivates the addition of sensors to directly track the robot end-effector and joint state. 
Magnetic tracking of the end-effector enables the detection of link deflection, and joint-mounted magnetic encoders enable the detection of true joint position despite cable stretch. Feedback controllers (described in Section \ref{section:controller}) are applied to compensate for these errors. \textcolor{black}{The modeled system stiffness was experimentally validated by applying a 5N (QWORK Slotted Weights) load to the robot's end-effector and measuring deflection using a dial indicator (Clockwise Tools, 0.03mm resolution) across 5 trials and resulting in $K_\text{link,meas} = 3.13\pm0.35mm$ with joint controller and $K_\text{system,meas} = 8.06\pm0.19mm$ without, corresponding to $1.59N/mm$ and $0.63N/mm$, comparing closely to our analytical calculations.}

The transmission's static load rating is analyzed to evaluate the design's ability to exert the forces required for needle insertion and provide sufficient stiffness. The end-effector's idler pulleys are the weakest component. Pulley wrap angles and the associated transmission load ratings change depending on the robot configuration. Configuration-dependent pulley load can be calculated as 
\begin{equation}
    F_\text{N} = F_\text{c} \sin\left(\frac{\theta }{2}\right)
\end{equation} 
where $ F_\text{N}$ is the normal force on the idler pulley's bearings, $ F_\text{c}$ is the cable tension, and $\theta$ is the pulley's wrap angle. 
The pulley wrap angle is represented as $\theta = \theta_\text{joint} + \theta_\text{offset}$ where $\theta_\text{joint}$ is the current joint angle and $\theta_\text{offset}$ is the cable wrap angle at the robot's nominal configuration which is varies based on the idler due to the robot design. At a configuration, a joint transmission's load rating is modeled as $ \tau  =  r_\text{jp}F_{\min }$ if a revolute joint or directly as $ F_{\min }$ if a prismatic joint. $r_\text{jp}$ is the joint pulley radius and $F_{\min }$ is defined as:
\vspace{-1mm}
\begin{equation}
    F_{\min } =  \min_{q \epsilon Q,  i\epsilon N}\frac{F_\text{r}}{\sin\left(\frac{q_{i}}{2}\right)}
\end{equation}
\vspace{-1mm}

for each joint angle, $q$, in the joint's range, $Q$, and every joint, $i$, in the cable-driven end-effector's joints, $N=4$. $F_\text{r}$ is the rated pulley load. The pulley load rating was evaluated throughout the joint range using the idler pulley's bearing load ratings (R2-5 bearing, $117N$ static radial load). The transmission is rated for $2.5Nm$, $1.25Nm$, $1.25Nm$, and $50N$ for the revolute and prismatic cable-driven joints from proximal to the distal end. 

\begin{figure}[tb!]
\centering
\includegraphics[width=\linewidth, trim={0cm 0cm 2.0cm 1.0cm},clip]{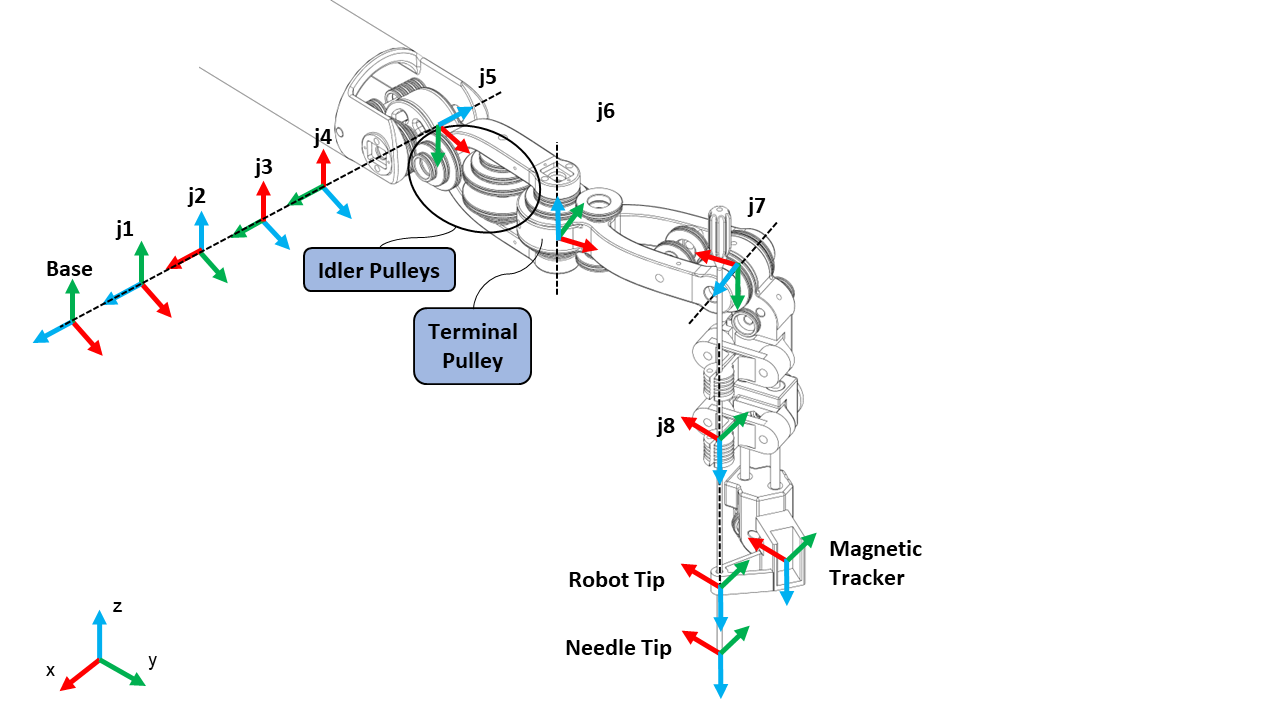}
\caption{Kinematics diagram for the in-bore joints of CRANE. The base joints (outside of the image) are modeled as virtual-joints coinciding at the first in-bore joint and provide gross positioning providing cartesian linear motion. The idler pulleys enable cable pass-throughs and are located on all intermediate joints and links. The pulleys labeled on link 5 provide support for the drive cables for joints 6, 7, and 8.}
\label{fig:kinematic_diagram}
\vspace{-3mm}
\end{figure}

\subsubsection{Joint Space and Actuator Space Relationship}
Actuators are coupled to the joints via the transmissions described in the previous sections. %These transmissions provide help with packaging and speed reduction. 
The base joints and trunnion joint, $\bm{q}_\text{b} = \{ q_1,...,q_4 \}$, are directly actuated via permanent magnet motors, have simple gear-ratios relating to actuator and joint spaces. The cable-driven in-bore joints, $\bm{q}_\text{b} = \{ q_5,...,q_8 \}$, are coupled due to the changing wrap angle on idler pulleys for cables passing over previous active joints. This relationship is modeled as: 
\begin{equation}
    \bm{q} = \bm{M} \bm{\theta}
\end{equation}
for joint positions, $\bm{q} =    
    \begin{bmatrix}
    \bm{q}_\text{b} & \bm{q}_\text{c} 
    \end{bmatrix}^{\top}  \in \mathbb{R}^8$, motor positions , $\bm{\theta} \in \mathbb{R}^8$, and coupling matrix $\bm{M} \in \mathbb{R}^{8\times8}$. $\bm{M}$ is represented as a block diagonal matrix: 
\begin{equation}
\bm{M}=
  \begin{bmatrix}
    \bm{M}_\text{b} & 0 \\
    0 & \bm{M}_\text{c}
  \end{bmatrix}
\end{equation}
\noindent 
for $\bm{M}_\text{b}$ the gear ratios for $\bm{q}_\text{b}$ and $\bm{M}_\text{c}$ the coupling matrix for $\bm{q}_\text{c}$  
\begin{equation}
\bm{M}_\text{b}=
  \begin{bmatrix}
   0.637 & 0 & 0 & 0 \\
    0 & 0.637 & 0 & 0 \\
    0 & 0 & 1.27 & 0 \\
    0 & 0 & 0 & -5.49 \\
  \end{bmatrix} 10^{-3}
\end{equation}
\noindent is found analytically and is diagonal due to the uncoupled design. $\bm{M}_\text{c}$ is found via least-squares regression with motor encoder measurements as inputs and joint-mounted encoder measurements for $\bm{q}_\text{c}$ as:
\begin{equation}
\bm{M}_{\text{c},i,*} = \bm{Q}_{i,*} \bm{\Theta}^{\dagger} \quad \text{where}\quad \bm{\Theta}^{\dagger}=\bm{\Theta}^\top\left(\bm{\Theta}\bm{\Theta}^{\top}\right)^{-1}
\end{equation}
\noindent for each row  $i$ of $\bm{M}_\text{c}$ with $\bm{Q}_{i,*}$ is a time series of $m$ samples of a single joint's angle measured directly via magnetic encoder and $\bm{\Theta} \in \mathbb{R}^{4 \times m}$ is a matrix time series of all motor angles. This results in:
\begin{equation}
\bm{M}_\text{c}=
  \begin{bmatrix}
 -0.21 & 0 & 0 & 0 \\
 0.15 & 0.20 & 0 & 0 \\
 -0.29 & -0.20 & 0.26 & 0 \\
 2.4\text{e-4} & 3.0\text{e-5} & -2.3\text{e-4} & 2.6\text{e-4} \\
  \end{bmatrix}
\end{equation}
\noindent $\bm{M}_\text{c}$ is lower triangular because of positional coupling between preceding and following cable-driven joints. 

\subsection{System and Software Architecture}
\begin{figure*}[htbp]
\includegraphics[width=\linewidth, trim={0cm 2.5cm 0 3cm},clip]{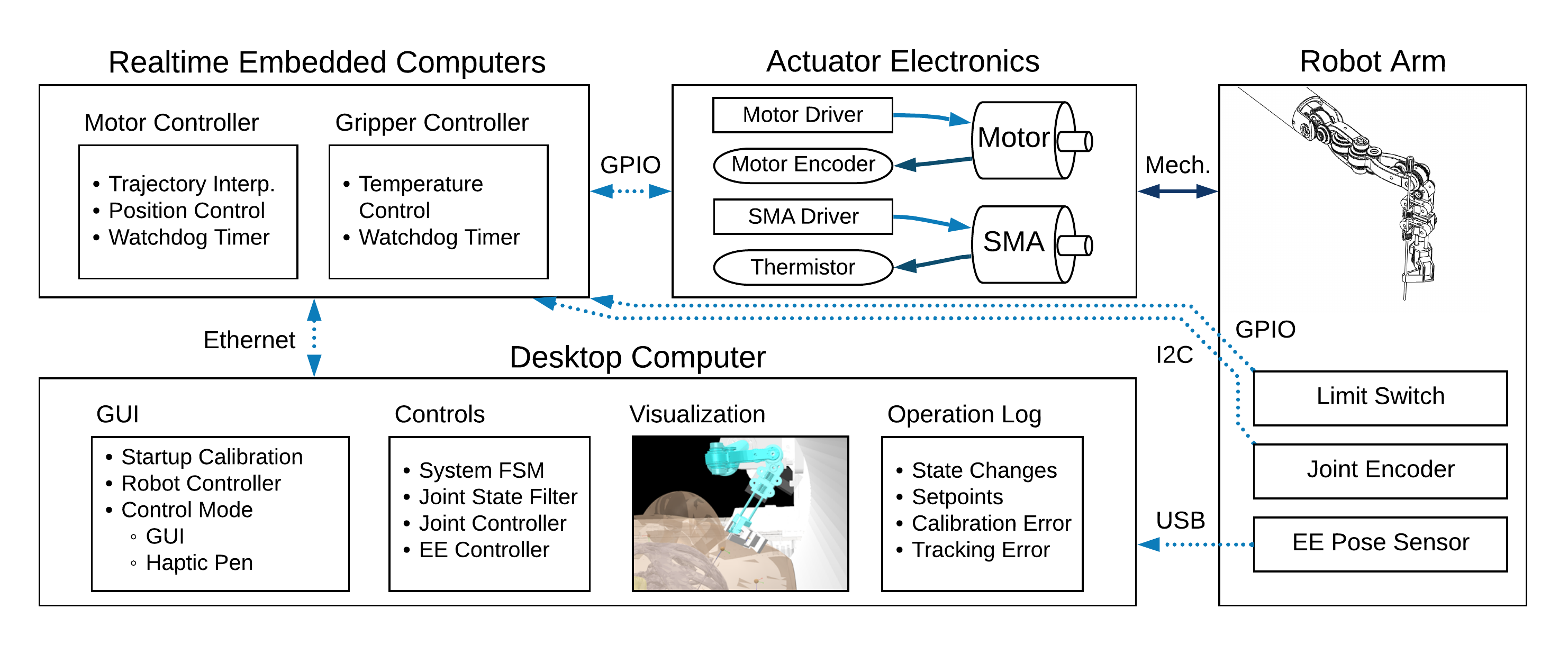}
\vspace{-4mm}
\caption{The Desktop Computer provides the Graphical User Interface and high-level intelligent high-level trajectory planning. Embedded Computers provide hardware sensing and control for safety-critical features and interfacing with the high-level control computer. The system is set up in a distributed architecture providing real-time control for higher rate components with more relaxed constraints for high-level interfacing to ease development and algorithm evaluation while retaining system safety.}
\label{fig:_highlevel_desktop_pc_component}
\vspace{-2mm}
\end{figure*}

The system architecture (Fig. \ref{fig:_highlevel_desktop_pc_component}) focuses on safety and extensibility. Real-time Embedded Computers (FPGAs and microcontrollers) run high frequency and \textcolor{black}{timing-jitter sensitive control}, allowing for independent development and upgrade of the high-level control system. The Desktop Computer hosts the user interface and high-level robot control, including kinematics and path planning.

User Datagram Protocol (UDP) over Ethernet with a dedicated network switch provides an extensible interface between computers. There are two primary electronic subsystems: Motor Controller and Clutch Controller. These components interface with the primary control computer, which performs high-level system coordination and user interface. 
%The centralized motor control motherboard, system power supplies, a network switch, and external interfaces, including emergency stop, power switch, AC input, and RJ-45 ethernet connection, are housed in the robot's base. 
Watchdog timers, redundant position sensors, mechanical gravity counterbalance, and error detection algorithms improve safety and enable a fail passive design. 
%ARM microcontroller boards provide needle gripper temperature control, joint mounted encoder interfaces, and limit switch interfaces. %\textcolor{blue}{are implemented on the real-time embedded computers and automatically disable power to the motors and needle grippers in case of significant tracking errors or low update rates, for instance, due to disconnection or software failure from the Desktop Computer. When combined with the mechanical safeties, this makes the system fail-passive. Redundant position sensors, including magnetic joint encoders and limit switches, are located on all axis and connected to the Desktop Computer via an ARM Cortex M7. An ARM Cortex M3-based control board manages the needle clutches and.} 
An Ascension trakSTAR magnetic tracker provides direct tip position and orientation sensing for the robot's end-effector's tip near the needle insertion point. 
%\textcolor{blue}{ on the human and directly interfaces with the Desktop Computer over USB.}% \cite{noauthor_ascension_nodate}. %The magnetic tracker is mounted distal tip of the robot near the needle insertion point to minimize sensor orientation tracking error.
%The Realtime Embedded Computer's Motor Controller runs $1khz$ synchronized motor position control with synchronized velocity and acceleration limited trajectory interpolation.

Robot Operating System (ROS) forms the basis for the multiple processes on the Desktop Computer via a standard messaging system. The Desktop Computer runs the \textcolor{black}{less timing-jitter} sensitive software, including robot kinematics and planning, image guidance, and user interaction. The device's multiple user interface options provide flexibility depending on the situation and physician's preference. %Control methods range from full automation given their Target Needle Insertion transform to manual joint control. 
The primary User Interface interface is a Qt5-based graphical user interface %(Fig. \ref{fig:system_ui_enables_physicians_plan})
that handles system initialization and setup, direct joint level control, end-effector control, and automated control incorporating path planning (described below in Section \ref{section:automated_planning_method}).  Further details on the electronics and software architecture are provided within \cite{schreiber_crane_2022}.

\section{Coordinate Systems and Transforms}

This system has three reference coordinate systems: the CT scanner, the magnetic tracker, and the robot. Homogeneous transforms provide relationships between coordinate frames, represented as 
\begin{equation}
^{a}\bm{T}^\text{b}_\text{c} =\begin{bmatrix}
\bm{R} & \bm{t} \\ 
0 & 1 \\ 
\end{bmatrix} \in SE(3)
\end{equation}
providing the transform from $c$ to $b$ relative to basis vectors $a$ for rotation matrix $\bm{R} \in SO(3)$ and translation vector $\bm{t} \in \mathbb{R}^3$. 
Typically the basis vectors $a$ are excluded from the transformation notation and assumed to be their system's base coordinate system. 
Transformation matrices represent both transforms between coordinate systems (as an operator) and poses within a coordinate system. 

The CT scanner and magnetic tracker are both sensors used as inputs to our system, with the robot as the primary coordinate system used internally. The transformation matrices below define poses relative to the three fixed reference coordinate systems relevant to our system.

Key poses defined relative to the CT scanner base frame, $\bm{T}^\text{sb}$, are:
\begin{itemize}
	\item $\bm{T}^\text{sb}_\text{tn}$ defines the Target Needle Insertion pose. This is typically on the patient's skin and serves as the Remote Center of Motion (RCM) location for adjustments made after insertion. This is created based on input by the physician using the GUI. \textcolor{black}{$\bm{T}^\text{sb}_\text{tn}$ is initialized to $I_4$. Each time the user presses a GUI button corresponding to a X, Y, and Z cartesian translation directions or roll, pitch, and yaw orientation adjustments, $\bm{T}^\text{sb}_\text{tn} \leftarrow \bm{T}^\text{sb}_{\delta \text{tn}} \bm{T}^\text{sb}_\text{tn}$ for the corresponding unit translation or rotation homogeneous transform $\bm{T}^\text{sb}_{\delta \text{tn}}$. The magnitude of $\bm{T}^\text{sb}_{\delta \text{tn}}$ is specified by a GUI slider.}  %$\bm{t}$ is provided by the physician as a point, and $\bm{T}_\text{R}$ is defined according to the vector direction from $\bm{t}$ towards the tumor within the body. 
	%$\bm{T}_\text{Rx}$ and $\bm{T}_\text{Ry}$ are defined arbitrarily (due to our symmetric needle assumption) such that $\bm{T}_\text{R}$ is an orthogonal matrix. %The cartesian point functions as the fixed fulcrum point for the remote center of motion.
	\item $\bm{T}^\text{sb}_{\text{fd}_{i}}$ defines the pose for the spherical CT-visible fiducials, $\text{fd}_i$, within the CT scanner frame and embedded in the magnetic tracker mount used for referencing the robot and scanner's coordinate system. These are located during the calibration step within the CT images (described in Section \ref{section:calibration}).
\end{itemize}

Key poses defined relative to the magnetic tracker base frame, $\bm{T}^\text{mb}$, are:
\begin{itemize}
    \item $\bm{T}^\text{mb}_\text{mtrkt}$, $\bm{T}^\text{mb}_\text{mtrkb}$ define the pose of the magnetic tracker mounted to robot tip and phantom; measured by the Ascension trackStar magnetic tracker at $100Hz$.
	\item $\bm{T}^\text{mtrkb}_{\text{fd}_{1,2,3,4}}$, is the static transform to each CT-visible fiducials relative to the magnetic tracker used for referencing the magnetic tracker and scanner's coordinate systems.
    %\item $\bm{T}^\text{mb}_\text{mtrkb}$ defines the pose of the magnetic tracker mounted to phantom and provided by the Ascension trackStar magnetic tracker at $100Hz$.
    %\item $\bm{T}^\text{mb}_{\text{fb}_{1,2,3,4}}$, is the static transform to each CT-visible fiducials relative to the phantom magnetic marker and embedded in the magnetic tracker mount used for referencing the magnetic tracker and scanner's coordinate systems.
\end{itemize}

Key poses defined relative to the robot base frame, $\bm{T}^b$, are:
\begin{itemize}
	\item $\bm{T}^\text{b}_\text{EE}$ defines the robot's EE pose calculated via the Forward Kinematics function (defined in Section \ref{section:fk}). This pose is attached to the needle insertion mechanism's needle guide, and the Z-axis is coaxial with the needle grippers and needle guide directed distally to the robot.

% 	\item $\bm{T}^{EE}_{nt}$ is a static transform from the robot's EE, $EE$, to needle insertion pose, $nt$ and called Needle Tip, pose, calculated as a static offset from Link 7. $\bm{R}$ matches joint 8's $R$ matrix.

	\item $\bm{T}^\text{EE}_\text{trk}$, is the static transform from the robot's EE to the magnetic tracker mounted on the robot needle guide.

	%\item $\bm{T}^\text{EE}_{\text{fde}_{1,2,3,4}}$, is the static transform to each CT-visible fiducials relative to the robot tip and embedded in the magnetic tracker mount used for referencing the robot and scanner's coordinate systems.
\end{itemize}

\noindent The transforms between the three coordinate base frames are calculated in a pre-procedural calibration step providing $\bm{T}^\text{b}_\text{mb}$, between the magnetic tracker's base and the robot base, and $\bm{T}^\text{b}_\text{sb}$, between the scanner's base and the robot base (described in Section \ref{section:alignment_method}).

\subsection{Forward Kinematics}\label{section:fk}
The robot's kinematics chain is described using \textit{Modified Denavit-Hartenberg (DH) Parameters} which attaches coordinate systems to each joint of the robot and defines the transform between these coordinate frames as $\bm{T}_i^{i-1}(q_i) = \bm{T}_\text{x}(\alpha_i, a_i) \bm{T}_\text{z}(\theta_i, d_i)$ where
%\begin{equation}%align*}
% \begin{align}
%     \bm{T}_\text{x}(\alpha_i, a_i) =
%     \begin{bmatrix}
%         1 & 0 & 0 & a_i \\
%         0 & \cos(\alpha_i) & -\sin(\alpha_i) & 0 \\
%         0 & \sin(\alpha_i) & \cos(\alpha_i) & 0 \\
%         0 & 0 & 0 & 1
%     \end{bmatrix} \\
% %\end{equation} 
% %\begin{equation}
%     \bm{T}_\text{z}(\theta_i, d_i) = 
%     \begin{bmatrix}
%         \cos(\theta_i) & -\sin(\theta_i) & 0 & 0 \\
%         \sin(\theta_i) & \cos(\theta_i) & 0 & 0 \\
%         0 & 0 & 1 & d_i \\
%         0 & 0 & 0 & 1
%     \end{bmatrix} 
% \end{align}
\begin{align}
    \bm{T}_\text{x}(\alpha_i, a_i) =
    \begin{bmatrix}
        1 & 0 & 0 & a_i \\
        0 & \cos(\alpha_i) & -\sin(\alpha_i) & 0 \\
        0 & \sin(\alpha_i) & \cos(\alpha_i) & 0 \\
        0 & 0 & 0 & 1
    \end{bmatrix} \\
%\end{equation} 
%\begin{equation}
    \bm{T}_\text{z}(\theta_i, d_i) = 
    \begin{bmatrix}
        \cos(\theta_i) & -\sin(\theta_i) & 0 & 0 \\
        \sin(\theta_i) & \cos(\theta_i) & 0 & 0 \\
        0 & 0 & 1 & d_i \\
        0 & 0 & 0 & 1
    \end{bmatrix} 
\end{align}

\noindent and $q_i$ is substituted into $d_i$ for prismatic joints and $q_i$ is substituted into $\theta_i$ for revolute joints. 

The forward kinematics function $\text{FK}(\bm{q})$, defines the robot base to end-effector (EE) transform by chaining together the transforms described by the DH convention:
%. For this robot, the first 7 joints are used for active robot tip positioning while the 8th joint is reserved for needle insertion and excluded from the forward kinematics functions:
\begin{equation}
    \label{eq:forward_kinematics}
     \text{FK} \left(\bm{q}\right) = \bm{T}^\text{b}_\text{EE} = \bm{T}^\text{b}_1% \Big(
     \prod \limits_{i=2}^{8} \bm{T}_i^{i-1} (\bm{q}_i)%\Big) \textbf{T}^{7}_{ee}
\end{equation}

\noindent with DH-parameters provided in Table \ref{table:dh_parameters}.

% \begin{figure*}[tb!]
% \centering
% \includegraphics[width=\linewidth]{./images/image7.png}
% \caption{System UI enables physicians to plan within their natural coordinate frames. Internal coordinate transforms transform image space targets to robot base frame targets.}
% \label{fig:system_ui_enables_physicians_plan}
% \end{figure*}

\begin{figure}[!t]
\centering
\includegraphics[width=0.8\columnwidth]{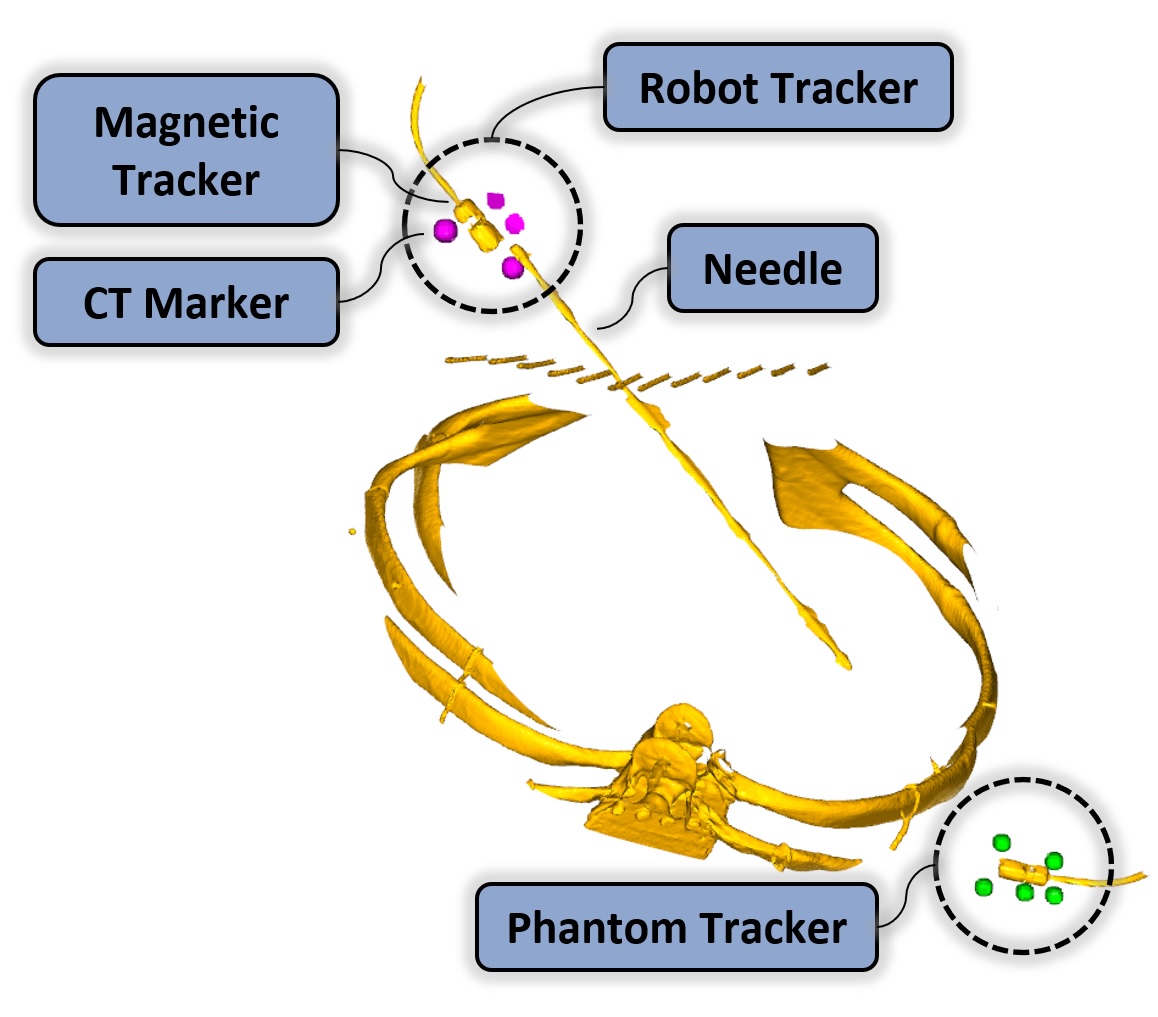}
\caption{The robot and phantom are localized within the CT scanner based on multiple rigid body fiducials, visualized here as part of the alignment process. The fiducials have a high Hounsfield Unit (HU) value of $3000HU$, and the nylon container has a value of $60HU$ and fully encases the marker sphere. This allows the marker to be segmented via thresholding, even if in contact with other high $HU$ materials. The magnetic tracker, needle, and phantom ribs are additionally visible.}
\label{fig:the_fiducials_have_high_hounsfield}
\end{figure}

% \begin{table}[bt!]
%     \caption{DH-parameters for CRANE where p is prismatic, and r is revolute. All angles and distances are in radians and meters, respectively.}
%     \label{table:dh_parameters}
%     \setlength\tabcolsep{1.0em}
%     \centering
%     \scriptsize
%     \begin{tabular*}{\columnwidth}{c @{\extracolsep{\fill}}c|c|cccc}
%           Frame & Type & $a_{i-1}$ & $\alpha_{i-1}$ & $D_i$ & $\theta_i$ \\ \hline
%           1 & p & 0 & $0$ & $q_1$ & $0$ \\
%           2 & p & 0 & $- \frac{\pi}{2}$ & $q_2$ & $- \frac{\pi}{2}$ \\
%           3 & p & 0 & $- \frac{\pi}{2}$ & $q_3$ & $- \frac{\pi}{2}$ \\
%           4 & r & 0 & 0 & 0 & $q_4$ \\
%           5 & r & 0 & $\frac{\pi}{2}$ & 0 & $q_5 + \frac{\pi}{2}$ \\
%           6 & r & 7e-2 & $\frac{\pi}{2}$ & 0 & $q_6$ \\
%           7 & r & 7e-2 & $\frac{\pi}{2}$ & 3e-2 & $q_7 - \frac{\pi}{2}$ \\
%           8 & - & 1e-2 & $-\frac{\pi}{2}$ & 2e-2 & 0 \\

%           9 & p & 0 & 0 & q_8 & 0 \\
%     \end{tabular*}
% \end{table}

\setlength{\tabcolsep}{10pt}
\begin{table}[b!]
\caption{DH-parameters for CRANE where p is prismatic and r is revolute. All angles and distances are in radians and meters, respectively.}
\label{table:dh_parameters}
\begin{center}
\begin{tabular}{llllll}
\toprule
Frame & Type & $a_{i-1}$ & $\alpha_{i-1}$ & $D_i$ & $\theta_i$ \\ \midrule
1 & p & 0 & $0$ & $q_1$ & $0$ \\
2 & p & 0 & $- \frac{\pi}{2}$ & $q_2$ & $- \frac{\pi}{2}$ \\
3 & p & 0 & $- \frac{\pi}{2}$ & $q_3$ & $- \frac{\pi}{2}$ \\
4 & r & 0 & 0 & 0 & $q_4$ \\
5 & r & 0 & $\frac{\pi}{2}$ & 0 & $q_5 + \frac{\pi}{2}$ \\
6 & r & 7e-2 & $\frac{\pi}{2}$ & 0 & $q_6$ \\
7 & r & 7e-2 & $\frac{\pi}{2}$ & 3e-2 & $q_7 - \frac{\pi}{2}$ \\
8 & - & 1e-2 & $-\frac{\pi}{2}$ & 2e-2 $+q_8$ & 0 \\
%9 & p & 0 & 0 & $q_8$ & 0 \\
\bottomrule
\end{tabular}
\end{center}
\end{table}

\subsection{Magnetic Tracker, Scanner, and Robot Calibration}\label{section:alignment_method} \label{section:calibration}
The robot to magnetic tracker calibration, $\bm{T}^\text{b}_\text{mb}$, and robot to scanner calibration $\bm{T}^\text{b}_\text{sb}$, transforms are calculated based on a calibration procedure to minimize a least squares error. 

%http://nghiaho.com/?page_id=671
\begin{equation}
    \begin{aligned}
        \min_{\bm{T}^\text{b}_\text{t}\in SE(3)} \sum_{i=0}^{N}
        \left\vert\left\vert \bm{\hat{P}}_{i} -\bm{T}^\text{b}_\text{t}
        \bm{P}_{i}\right\vert\right\vert_{2}% \text{ where } \\
        % \bm{\hat{ P }} = 
        % \begin{bmatrix}
        % x_{1} & y_{1} & z_{1} & 1 \\ 
        % \vdots  & \vdots  & \vdots  & \vdots  \\ 
        % x_{n} & y_{n} & z_{n} & 1
        % \end{bmatrix}
        % \in \mathbb{R}^{N \times 4}
    \end{aligned}
\end{equation}
where $\bm{T}^\text{b}_\text{t}$ is the calibration transform between robot base frame and the target base frame. $\bm{P} = \{\bm{p}_0, \ldots, \bm{p}_N \} \text{ and } \bm{\hat{P}} = \{\bm{\hat{p}}_0, \ldots, \bm{\hat{p}}_N \}$ are the augmented Cartesian calibration point sets for $N$ sample points and $\bm{p}, \bm{\hat{p}} = \begin{bmatrix}
x & y & z & 1 \\ 
\end{bmatrix}^\top$. 

For $\bm{T}^\text{b}_\text{mb}$, a calibration trajectory is performed using the robot's linear base joints, $\bm{Q}_\text{b} \in \mathbb{R}^{N \times 3}$. Each sample $\bm{p}_{i}
=\bm{t}_\text{FK}$ for $\bm{T}_\text{FK} = \text{FK} ( \bm{q}_i ) \bm{T}^\text{EE}_\text{trk}$ 
%for is a single time-sample of the predicted magnetic tracker position in the robot's base frame based on robot's kinematics and each row of 
and $\bm{\hat{p}}_{i}= \bm{t}_\text{mtrk}$ for the corresponding timestep $\bm{T}^\text{mb}_{\text{mtrkt},i}$. % the corresponding pose from the Ascension TrackSTAR tracker mounted to the robot's EE within the magnetic tracker base frame. %Our calibration trajectory is a prismatic motion using the gross positioning stage due to its high accuracy with $N=1000$ samples.
For $\bm{T}^{b}_{sb}$, a single CT scan is taken and processed to localize the fiducials within the scanner's coordinate system. Each sample $\bm{p}_{i}
= \bm{t}_\text{fid}$ for $\bm{T}_\text{fid} = \bm{T}^\text{b}_\text{mb} \bm{T}^\text{mb}_\text{mtrkb} ( \bm{q} ) \bm{T}^\text{mtrkb}_{\text{fd},i} \text{ for } i \in \{1\text{..}N\}$
and $\bm{\hat{p}}_{i}$ are the measured positions of the fiducial's centroids found within the CT image. For CRANE, the rigid body markers consists of multiple CT visible fiducial spheres  (Beekley CT-SPOT 120) (Fig. \ref{fig:the_fiducials_have_high_hounsfield}).
%is the predicted fiducial's position based on the robot's kinematics and each row of $\bm{P}_{i}  \text{ for } i \in \{1\text{..}N\}$
%A single CT scan is taken and processed to localize the fiducials within the scanner's coordinate system. 
%The fiducials (Beekley CT-SPOT 120) are embedded in a polymer housing (Markforged Onyx) which additionally serves as the needle guide and mounts the magnetic tracker to the EE (. 
% The spheres are localized within the scanner frame and detected based on a sphere fit quality.
%and association are made between points in $\bm{P}$ and $\bm{\hat{P}}$ to minimize the calibration error. 
\section{Automated Robot Trajectory Planning}\label{section:automated_planning_method}

\setlength{\tabcolsep}{8pt}
\begin{table}[t!]
    \caption{Mathematical symbols for Automated Robot Trajectory Planning}
    \label{table:symbols}
    \begin{center}
    \begin{tabular}{lm{57.5mm}}
    \toprule
    Symbols & Definition \\ \midrule
    % $\Delta L$ & cable length change\\
    % $F$ & Force used for pulley load calculation\\
    % $A$ & Cable cross-sectional area\\
    % $E$ & Young's modulus\\
    % $r$ & pulley radius\\
    % $\alpha, a, \theta, d$ & DH parameters \\
    % $\Delta \theta$ & joint angle change due to cable length change\\
    % $K$ & spring constant \\
    % $\bm{P}$ & matrix of points for base transform calibration\\
    $\bm{q}$, $\bm{\theta}$ & joint and actuator configuration\\
    $\bm{Q}$ & trajectory of joint configurations expressed as a matrix\\
    $\mathcal{X}$ & set of poses\\
    $\mathcal{B}$ & set of obstacles\\
    $D$ & dimension of a configuration, i.e., the DoF of the robot\\
    % $\bm{\Theta}$ & trajectory of actuator configurations\\
    $\bm{T}$, $\bm{R}$, $\bm{t}$ & transformation matrix \\&comprising rotation matrix and translation vector with rotation sub-matrix and translation sub-vector\\
    $\text{\sc rotMat}(\bm{n}, \psi)$ & rotation matrix via Rodrigues' formula around $\bm{n}$ axis an angle of $\psi$\\%$\omega$ axis of $\psi$ angle\\ % from 9.4.1.1 Exponential map of https://ingmec.ual.es/~jlblanco/papers/jlblanco2010geometry3D_techrep.pdf
    % $\bm{T}^\text{b}_\text{tn}$ & Target Needle Insertion transform in robot base frame\\
    %$\bm{x}$ & Workspace pose\\
    $\mathcal{C}$ & Configuration space (C-Space)\\
    % $\mathcal{C}_\text{free}, \; \mathcal{C}_\text{obs}$ & obstacle-free subspace and obstacle subspace \\
    % $\mathcal{C}_\text{feas}, \; \mathcal{C}_\text{goal}, \; \mathcal{C}_\text{adj}$ & feasible subspace, goal subspace, and adjustable subspace\\% , $\mathcal{X}$ and Workspace\\
    $d_{\textit{B}}(\bm{q})$ & distance to obstacles $\mathbbm{B}$ in Workspace\\
    $d_{\textit{o}}(\bm{T}_\text{tar}, \bm{T}_\text{cur})$ & task orientation difference between poses\\
    $d_{\textit{p}}(\bm{T}_\text{tar}, \bm{T}_\text{cur})$ & position difference between poses\\
    $\Delta_\text{adj}$ & zenith angle for conical RCM adjustability evaluation \\
    %$\bm{x}$, $\bm{p}$, $\bm{n}$ & target needle insertion vector \\&comprising target point and normal vector\\
    
    $\text{FK}(\bm{q})$ & Forward Kinematics pose \\
    
    $\bm{J}(\bm{q})$, $\bm{J}^{\dag}(\bm{q})$, $\lambda$ & Jacobian matrix, damped inverse Jacobian matrix, and damping term\\
    
    %$\nu,  & twist \\
    $\bm{f}$ & wrench\\ %https://modernrobotics.northwestern.edu/nu-gm-book-resource/3-4-wrenches/
    
    $\epsilon$, $\epsilon_p$, $\epsilon_o$ & pose error tolerance consisting of position and orientation\\
    $\bm{e}, \bm{e}_\text{p}, \bm{e}_\text{o}$ & pose error vector comprising position and orientation\\
    
    $c(\bm{q})$ & configuration cost function\\
  
    $\bm{M}$, $\bm{K}$ & actuator to joint space matrix and actuator controller gains\\
    $\tau$ & actuator torque\\
\bottomrule
\end{tabular}
\end{center}
\end{table}

This section describes a method enabling the robot to safely and dexterously manipulate a needle within a scanner while avoiding collisions with the environment, enabling the physician to be hands-off during device setup. This is accomplished via a hierarchical planning and control scheme for two phases of the procedure: Automated Device Setup to $\bm{T}_\text{tn}^\text{b}$ during \textbf{Procedure Planning Phase} and fine adjustment of the robot's configuration around $\bm{T}_\text{tn}^\text{b}$ during \textbf{Procedure Execution Phase}. For device setup, an optimal Dexterous Device Setup Configuration, $\bm{q}^\ast$, for $\bm{T}_\text{tn}^\text{b}$ is determined. Then, a collision-free configuration space trajectory, $\bm{Q}: \left\{\bm{q}_{0}, \bm{q}_{1},\ldots, \bm{q}_{n-1}, \bm{q}^{\ast}\right\}$, is determined using a sampling-based motion planner ($\text{ \sc {BiRRT}}$ implemented in OMPL \cite{sucan_open_2012}). Once $\bm{Q}$ is executed, a Local Controller is used for servoing the end-effector in the case of small adjustments of $\bm{T}_\text{tn}^\text{b}$.

\subsection{Dexterous Robot Setup Configuration: Problem Definition}
% The workspace $\mathcal{X}$ consists of needle insertion vectors, $\bm{x}$, and is the concatenation of a position $\bm{p}$ and normal vector $\bm{n}$:
% \begin{equation}
%     \bm{x} = 
%     \begin{bmatrix}
%     \bm{p} \\
%     \bm{n}
%     \end{bmatrix} \text{ for }
%     \bm{p} = 
%     \begin{bmatrix}
%         x\\
%         y\\
%         z\\
%     \end{bmatrix} \text{ and }
%     \bm{n} = 
%     \begin{bmatrix}
%         a\\
%         b\\
%         c\\
%     \end{bmatrix}
% \end{equation}
% The function $FK: C \rightarrow X$ translates between the configuration space and the workspace.
%  map between X and T by BLAH FUNCTION.
%The robot's Configuration Space (C-space) defines the possible robot configurations. 
A Dexterous Robot Setup Configuration is one for which the robot can feasibly insert a needle into the patient and manipulate the needle around the nominal $\bm{T}_\text{tn}^\text{b}$. This is defined based on several metrics, including joint limits, workspace singularities, and collisions with the environment. The C-space $\mathcal{C} \subset \mathbb{R}^{D}$ where $D$ is the robot's number of joints defines the possible robot configurations and contains the following important sub-spaces for our optimization problem:-
\begin{enumerate}
    \item Feasible C-space, $\mathcal{C}_\text{feas}$, can apply sufficient forces and moments to insert and adjust the angulation of the needle. %where the robot is sufficiently far from joint limits and c
    \item Collision-free C-space, $\mathcal{C}_\text{goal}$, where $\text{FK}(\mathcal{C})$ is sufficiently close to $\bm{T}_\text{tn}^\text{b}$.
    \item Collision-free C-space, $\mathcal{C}_\text{free}$, where the robot is sufficiently far from a collision with obstacles, including environment and self-collisions.
    \item Adjustable C-space, $\mathcal{C}_\text{adj}$, where the robot can perform an RCM adjustment around the Target Needle Insertion pose for a defined conical region.
\end{enumerate}
%The C-space $\mathcal{C} \subset \mathbb{R}^{D}$ where $D$ is the robot's number of joints defines the possible robot configurations and is bounded by upper joint limits $\bm{q}_\text{high}$ and lower joint limits $\bm{q}_\text{low}$. 

% The function $FK: C \rightarrow X$ translates between the configuration space and the workspace.
% The workspace is represented by:
% \begin{equation}
%     \mathcal{T} = \{ \bm{T} \vert \bm{T} =  \text{TF}\left(\bm{q} \right), \forall \bm{q} \in C \} \subset SE(3) 
% \end{equation}
% \noindent comprising the obstacle space, $X_\text{free} = \text{FK}( C_\text{free} )$ and $X_\text{obs} = X \setminus X_\text{free}$.
%\{ \bm{x} \vert \bm{x} =  \text{FK}\left(\bm{q} \right), \forall \bm{q} \in C_\text{obs} \} \subset SE(3) $, and the obstacle-free space $X_\text{free} = \{ \bm{x} \vert \bm{x} =  \text{FK}\left(\bm{q} \right), \forall \bm{q} \in C_\text{free} \} \subset SE(3) $.

For our needle insertion task, we define $\mathcal{C}_\text{feas}$ comprising non-singular configurations where the system can manipulate a needle with sufficient force. The Jacobian matrix, $\bm{J}(\bm{q}) \in \mathbb{R}^{6 \times d}$ and denoted as $\bm{J}$, relates EE twists, $\bm{\nu} \in \mathbb{R}^6$ and joint velocities, $\dot{\bm{q}} \in \mathbb{R}^{D}$ as well as relating EE wrenches, $\bm{f} \in \mathbb{R}^6$, and joint torques, $\bm{\tau} \in \mathbb{R}^{D}$:
\begin{align}
    \bm{\nu} = \bm{J} \dot{\bm{q}} \text{ and }
    \bm{\tau} = \bm{J}^\top  \bm{f}
\end{align}
The space and body Jacobians, $\bm{J}^\text{s}$ and $\bm{J}^\text{b}$, represented in the robot's base coordinate system and $EE$ coordinate systems are used throughout this section. Using these relationships, $\mathcal{C}_\text{feas}$ is:
\begin{equation}\label{equation:C_feas}
\begin{aligned}
\mathcal{C}_\text{feas} = \{ 
\bm{q} \in \mathcal{C} \vert  
%\bm{q}_\text{low} + \epsilon_\text{q} <\bm{q}<\bm{q}_\text{high} -  \epsilon_\text{q}   \\
\bm{J}^{\text{b},\top} \bm{F}_\text{req}<\bm{\tau}_\text{max} \}% \\
% \bm{\nu}_\text{req}< \bm{J}^\text{b} \dot{\bm{q}}_\text{max} \}
\end{aligned}
\end{equation}
%\noindent where $ \epsilon_\text{q} $ is how close nominal configurations may be to joint limits, where $ \bm{J}$ is the robot's space Jacobian function, 
\noindent where $\bm{f}_\text{req} \in \mathbb{R}^6 $ is the required force and moment to insert and manipulate the needle represented as a wrench in the $EE$ frame and $\bm{\tau}_\text{max} \in \mathbb{R}^{D}$ is the robot's maximum joint torques.%, $\bm{\nu}_\text{req} \in \mathbb{R}^6$ is a twist in the $EE$ frame and the maximum target needle velocity, and $\dot{\bm{q}}_\text{max}$ is the robot's maximum joint velocity. 

Furthermore, we define $\mathcal{C}_\text{goal}$ as configurations with the robot's EE near the Target Needle Insertion pose $\bm{T}^\text{b}_\text{tn}$. The function $d_{\textit{p}}(\bm{T}_\text{tar}, \bm{T}_\text{cur} )$ computes the vector difference between the position vectors of two poses and $d_{\textit{o}}(\bm{T}_\text{tar}, \bm{T}_\text{cur} )$ computes the axis-angle difference between the Z-axis of two poses: 
\begin{equation}
    d_{\textit{p}}(\bm{T}_\text{tar}, \bm{T}_\text{cur} ) = \bm{t}_\text{tar} - \bm{t}_\text{cur}
\end{equation}
\begin{equation}
\begin{aligned}
    d_{\textit{o}}(\bm{T}_\text{tar}, \bm{T}_\text{cur} ) = \cos^{-1} \left( \frac{\tilde{\bm{z}}^\top \bm{z}}{\|\tilde{\bm{z}}\|_2 \|\bm{z}\|_2} \right) \left( \tilde{\bm{z}} \times \bm{z} \right) \\
\end{aligned}
\end{equation}
where $\tilde{\bm{z}} = \bm{R}_{\text{targ},[z]}$ and $\bm{z} = \bm{R}_{\text{cur},[z]}$ are the Z-axis vectors of the rotation matrix from their corresponding poses.
Using these difference functions, $\mathcal{C}_\text{goal}$ is:
% \begin{equation}
% C_\text{goal} =\left\{ \bm{q} \vert \;
% \|  \text{FK} \left( \bm{q} \right)-\bm{x}_{goal}\|_{2}\leq \epsilon_p \right\}
% \end{equation}
\begin{equation}
\begin{aligned}
\mathcal{C}_\text{goal} =\{ \bm{q} \in \mathcal{C}_\text{feas} \vert \;
\|  d_\textit{o} (\text{FK} \left( \bm{q} \right), \bm{T}^\text{b}_\text{tn}) \|_{2} \leq \epsilon_o \\ 
\text{ and } \|  d_{\textit{p}} (\text{FK} \left( \bm{q} \right), \bm{T}^\text{b}_\text{tn}) \|_{2}\leq \epsilon_p \}
\end{aligned}
\end{equation}
where $\epsilon_\text{p}$ is a predefined position error tolerance and $\epsilon_\text{o}$ is a predefined orientation error tolerance. %Typically, $\bm{x}_\text{goal} = \bm{T}^\text{b}_\text{tn}$.

The obstacle-free C-space is the set of robot configurations in which the robot's links are sufficiently far from obstacles, $\mathcal{B}$, (e.g., scanner, patient, self-collisions):
\begin{equation}
    \mathcal{C}_\text{free} =
    \{ \bm{q} \in \mathcal{C} \vert d_\textit{B} \left( \bm{q} \right) > \epsilon_d \}
\end{equation}
\noindent where $d_\textit{B} \left( \bm{q} \right)$ is the minimum distance from the robot to obstacles and $\epsilon_d$ is a specified minimum distance to collision for environment padding.

Finally, we define $\mathcal{C}_\text{adj}$ comprising the region of the configuration space where the robot can perform an RCM adjustment of the needle in a conical region around the current configuration:
\begin{equation}\label{eq:C_adj}
\begin{aligned}
\mathcal{C}_\text{adj} = \{ \bm{q} \in \mathcal{C}_\text{goal} \cap \mathcal{C}_\text{free} \vert \exists \bar{\bm{q}} \in \mathcal{C}_\text{feas} \text{ s.t. } \\
\text{ \sc connectable}(\bar{\bm{q}}, \bm{q}), \\ 
\| d_\textit{o} (\bm{T}_\text{rcm}, \text{FK}(\bar{\bm{q}})) \|_2 < \epsilon_\text{o}, \\
\| d_\textit{p} (\bm{T}_\text{rcm}, \text{FK}(\bar{\bm{q}})) \|_2 < \epsilon_\text{p}, \\ 
\forall \bm{T}_\text{rcm} \in \mathcal{X}_\text{rcm}  \}
% C_\text{adj} = \{ \bm{q} \in C_\text{free} \vert \text{FK}\left( \bm{q} \right)+\Delta \bm{x} \in X_\text{free}, \\
% \forall  \Delta \bm{x} \in \Delta X_{\text{RCM}}  \}
%\; \\\text{s.t.} \; || \Delta \bm{x} || < \Delta \bm{x}_\text{ori}\\ \text{s.t.} \; \Delta \bm{x}_{t} = 0 \}
\end{aligned}
\end{equation}
\noindent Robot configurations $\bm{q}$ and $\bar{\bm{q}}$ are $\text{ \sc connectable}$ if a simple Local Planner can provide a collision-free trajectory, $\bm{Q} = \{\bm{q}, \text{..}, \bar{\bm{q}} \}$, between them. Here, the Local Planner is a gradient descent IK method with a nullspace objective. $\bm{T}_\text{rcm}$ is a pose within $\mathcal{X}_\text{rcm}$ where $\mathcal{X}_\text{rcm} \in SE(3)$ are the space of homogeneous poses within a conical orientation adjustment of $\bm{T}^\text{b}_\text{tn}$. $\mathcal{X}_\text{rcm}$ is constructed using $\text{\sc rotMat}(\bm{n}, \psi)$ which defines the $SO(3)$ rotation matrix from an arbitration rotation angle, $\psi$, around an arbitrary axis, $\bm{n}$, via Rodrigues' Formula as:
\begin{equation}\label{eq:RCM_constraint}
\begin{aligned}
     \mathcal{X}_{\text{rcm}} = \{ \bm{T}_\text{adj}\bm{T}^\text{b}_\text{tn} \text{ where } \\
     \bm{R}_\text{adj} = \text{\sc rotMat}(x, \delta) \text{\sc rotMat}(z, \gamma) \text{ and } \\
     \bm{t}_\text{adj} = \bm{0} \; \forall \delta \in [0, \Delta_\text{adj}] \text{ and } \forall\gamma \in [0, 2\pi] \} 
    %$\Delta R_{zenith} \gets \bm{R}(x, \Delta x_{ori} / M)$
\end{aligned}
\end{equation}
% \begin{equation}
% \begin{aligned}
%     \Delta X_{\text{RCM}} = \{ \bm{x}  | \; || \text{\sc posDiff}( \bm{x} ,\bm{I}_\text{4}) ||_2 = 0 \\
%     \text{ and } || \text{\sc oriDiff}( \bm{x} ,\bm{I}_\text{4}) ||_2 < \epsilon_{adj} \}
%     %$\Delta R_{zenith} \gets \bm{R}(x, \Delta x_{ori} / M)$
% \end{aligned}
% \end{equation}
\noindent where $\Delta_\text{adj}$ is the zenith angle of the conical region, $\bm{T}_\text{adj}$ is the transform used for RCM angle adjustment defined with with rotation submatrix $\bm{R}_\text{adj}$ and translation subvector $\bm{t}_\text{adj}$. $\text{\sc rotMat}(x, \alpha)$ and $\text{\sc rotMat}(z, \gamma)$ are the rotation matrices around the $x$ and $z$ axis for RCM orientation adjustment, and the translation $\bm{t}_\text{adj}$ is zero due to the RCM motion constraint. Only angulation adjustments are evaluated as the needle is inside the body, and translation would result in significant forces being applied to the tissue.

Using the defined C-space, $\bm{q}^\ast$ corresponds to minimizing a corresponding cost function:
\begin{equation} \label{eq:q_optimization} 
    \bm{q}^{\ast} = \argmin_{\bm{q} \in \mathcal{C}_\text{adj}} c \left( \bm{q} \right)
\end{equation}
where $c(\bm{q})$ can be defined arbitrarily.
Our $c(\bm{q})$ focusing on maximizing dexterity and distance to collision while minimizing joint motion is defined as:
\begin{equation}\label{eq:cost_dexterity}
c \left( \bm{q} \right) =
\frac{\alpha }{w( \bm{q}) } + 
\frac{1-\beta}{d_{\textit{B}\{\text{bor}\}}\left( \bm{q} \right)} + 
\frac{\beta}{d_{\textit{B}\{\text{pat}\}}\left( \bm{q} \right)} + \frac{\gamma} {d_{\bm{q_0}}\left(\bm{q}\right)}
\end{equation}
\noindent where  $ \alpha, \beta \gamma$ are the optimization priorities, 
$w \left( \bm{q} \right) = \sqrt{ \left| \bm{J}^\text{b}_{ \left[ \alpha,\beta \right]} \bm{J}^{\text{b},\top}_{ \left[ \alpha,\beta \right] } \right| }$
is a modified version of Yoshikawa manipulability index \cite{yoshikawa_manipulability_1985} for the configuration calculated using the rows of the body Jacobian corresponding to roll and pitch orientation axes. Other common dexterity indices\cite{klein_dexterity_1987,asada_geometrical_1983} can also be applied. $d_{\textit{B}\{\text{bor}\}}$ is the minimum distance-to-collision between the robot and the scanner bore, $d_{\textit{B}\{\text{pat}\}}$ is the minimum distance-to-collision between the robot (excluding needle insertion mechanism) and the patient, and $d_{\bm{q_0}}\left(\bm{q}\right)$ is the distance from the evaluated robot configuration to initial robot configuration. 

\subsection{Dexterous Robot Configuration Generation}
%The optimal joint configuration, 
A general-purpose global optimization algorithm \cite{noauthor_simplicial_nodate} with $\bm{q} \in \mathcal{C}_\text{feas}$ directly turned into an inequality constraint is used to determine $\bm{q}^\ast$. Eq. \eqref{eq:q_optimization} and  $\mathcal{C}_\text{adj}$ are directly evaluated in the main optimization function with $\bm{f}_{ [ \text{z} ]} = 10N$, $\bm{f}_{ \left[ \alpha,\beta \right] } = 0.05Nm$, and $\Delta_\text{adj} = 15^\circ$.
%\begin{bmatrix}
%0 & 0 & 10 & 0.05 & 0.05 & 0.0 
%\end{bmatrix}^\top$
%with $F_\text{req} = \begin{bmatrix}
    % 10 & 10 & 10 & 0.05 & 0.05 &0.05 
    % \end{bmatrix}^\top $. 
The obstacles set for our optimization is $\mathcal{B} = \{\text{robot, scanner bore, patient body}\}$.

For a redundant robot, the joints can be partitioned as $\bm{q}=     
    \begin{bmatrix}
    \tilde{\bm{q}} & \mathring{\bm{q}} 
    \end{bmatrix}^\top $ 
for the non-redundant joints, $\tilde{\bm{q}}$, and redundant joints $\mathring{\bm{q}}$ with respective non-redundant C-space $\tilde{\mathcal{C}} \subset \mathcal{C}$ and redundant C-space $\mathring{\mathcal{C}} \subset \mathcal{C}$. Correspondingly, $\bm{J}$ can be reorganized into redundant and non-redundant block matrices:
\begin{equation}
\bm{J} =    
    \begin{bmatrix}
    \tilde{\bm{J}} & \mathring{\bm{J}} 
    \end{bmatrix} 
\end{equation}
where $\tilde{\bm{J}} \in \mathbb{R}^{6 \times 6}$ corresponding to the non-redundant joints and $\mathring{\bm{J}} \in \mathbb{R}^{6 \times D-6}$ corresponding to the redundant joints. 

%The global optimizer samples $\mathring{\bm{q}} \in \mathring{\mathcal{C}}$ due to its lower dimensionality and then a gradient-descent-based IK formulation determines $\tilde{\bm{q}} \in \tilde{\mathcal{C}}$.
The function $\text{\sc IKConfigurationLoss}(\bm{T}^\text{b}_\text{tn}, \bm{q}_0)$ jointly evaluates the optimization objectives $c(\bm{q})$ and if $\bm{q} \in \mathcal{C}_\text{adj}$. $\text{\sc IKConfigurationLoss}(\bm{T}^\text{b}_\text{tn}, \bm{q})$  returns $c(\bm{q})$ and $\bm{q}$ if $\bm{q} \in C_\text{adj}$ or otherwise returns a large cost, $c_\text{infeasible}$, and, initial joint configuration, $\bm{q}_0$. $\text{\sc IKConfigurationLoss}$ evaluates a nominal robot configuration's dexterity. The nominal configuration is determined by fixing $\mathring{\bm{q}}$ to the configuration determined by the global optimizer to resolve redundancy and solving for $\tilde{\bm{q}}$ using $\tilde{\bm{J}}$ with a gradient-descent-based IK formulation. 
For CRANE, $\tilde{\bm{q}} = \{ q_1, q_2, q_3, q_4, q_7\}$, $\mathring{\bm{q}} = \{ q_5, q_6 \}$, and $q_8$ is excluded.

IK is implemented following the Levenberg-Marquardt algorithm for stability near singularities $\bm{J}^{\dag}=\left( \bm{J}^{\top}\bm{J}+ \lambda \bm{I} \right)^{-1} \bm{J}^{\top}$ with damping term $\lambda$. The pose error between two $SE(3)$ poses is calculated as the cartesian position error and axis-angle orientation error between the pose's $R_\text{z}$ vectors. During the \emph{Solve IK} portion of $\text{\sc IKConfigurationLoss}$, a nominal IK solution is found for $\tilde{\bm{q}}$ with $\tilde{\bm{J}}$, providing a full robot configuration, $\bm{q}$, when combined with $\mathring{\bm{q}}$ from the general-purpose global optimization algorithm. 

\begin{figure}[!t]
\centering
    \begin{subfigure}{0.85\columnwidth}
         \centering
        \includegraphics[width=\textwidth]{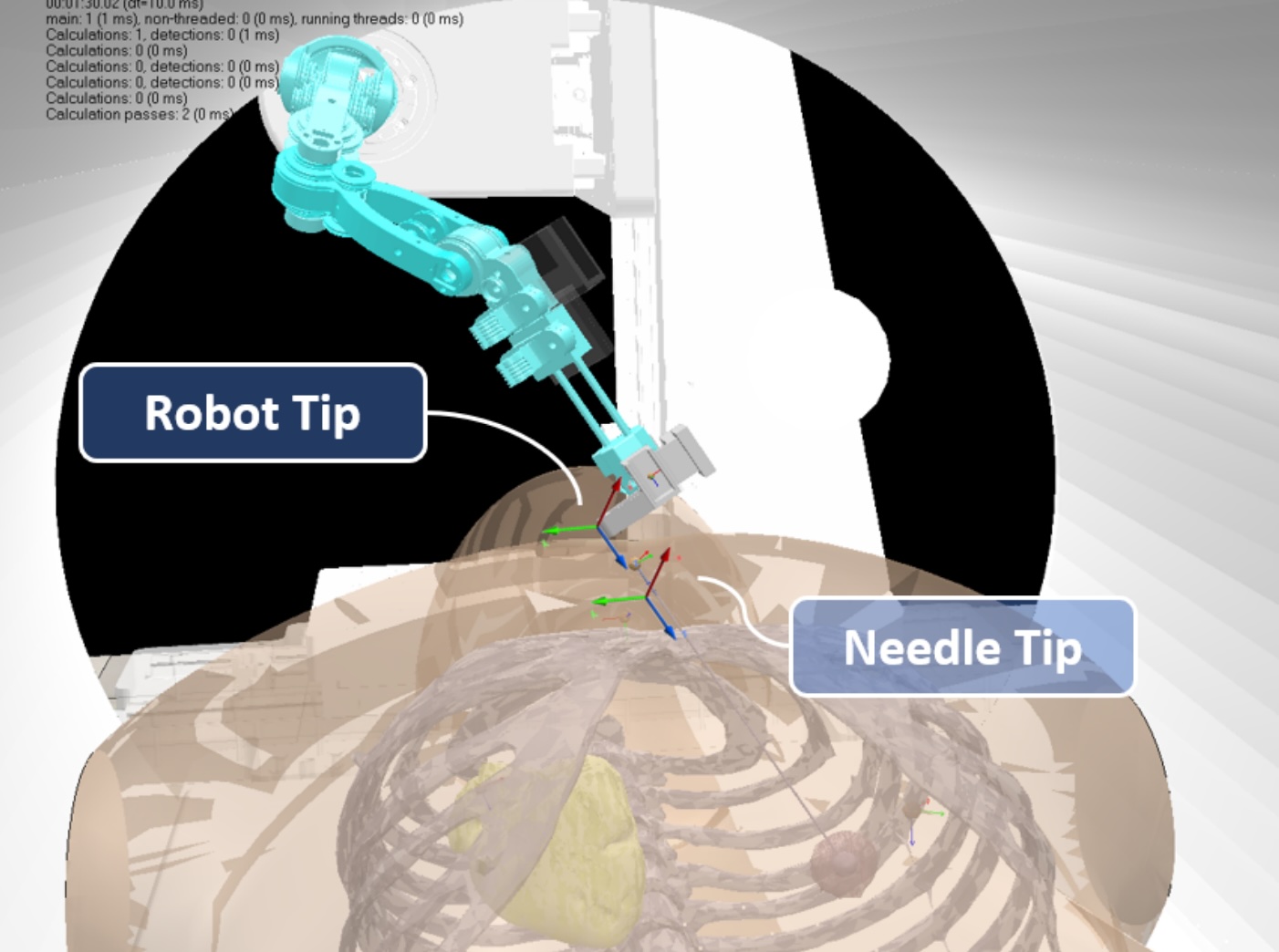}
         \caption{Prioritized patient avoidance, $\beta < 0.5$}
         %\%%\label{fig:three sin x}
     \end{subfigure}
    \hfill
    \\
     \begin{subfigure}{0.85\columnwidth}
         \centering
        \includegraphics[width=\textwidth]{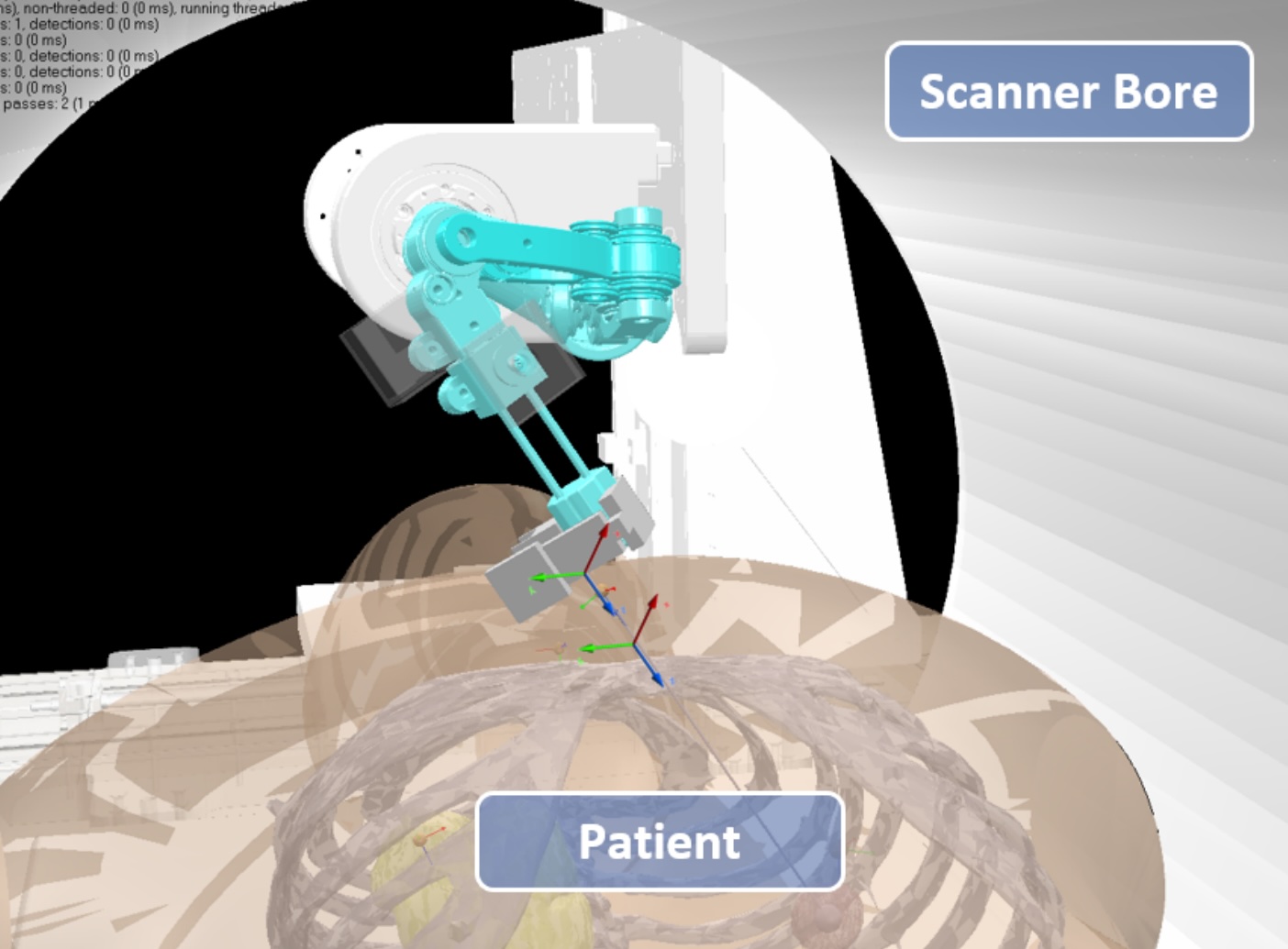}
         \caption{Equal priority patient and scanner avoidance, $\beta = 0.5$}
         %%\label{fig:five over x}
     \end{subfigure}
     \caption{Solution of multiple joint configurations for a $\bm{T}^\text{b}_\text{tn}$ based on optimization weights and utilizing CRANE's redundant in-bore joints. CRANE provides multiple solutions for a Target Needle Insertion pose, considering physician preferences on obstacle avoidance while ensuring a well-conditioned dexterous solution.}
\label{fig:solution_multiple_joint_configurations_target}
\vspace{-3mm}
\end{figure}

During \emph{Evaluate Adjustability} of $\text{\sc IKConfigurationLoss}$, the nominal joint configuration is evaluated for RCM adjustability as defined in Eq. \eqref{eq:C_adj}. $\text{\sc calcLocalTargets} (\cdot)$ creates a set of adjustable angle target poses $X_\text{adj}$ around $\bm{T}^\text{b}_\text{tn}$ satisfying the RCM constraint by rotating the nominal insertion pose. During this evaluation, the full Jacobian, $\bm{J}$, is used with a nullspace objective to remain near the previously determined nominal robot configuration. For each step of the gradient-descent, $\bm{q}_\text{local}$ is evaluated for if it remains within $\mathcal{C}_\text{feas}$. This evaluation implicitly checks the path between $\bm{T}_\text{target}$ and $\bm{T}_\text{local}$ is collision-free due to the limited step size of the gradient step. $c_\text{infeasible}$ specified as a large value for in-collision configurations and is selected sufficiently large that it is greater than any cost from $c(\bm{q})$. Depending on the cost function weighting, different robot configurations are optimal (Fig. \ref{fig:solution_multiple_joint_configurations_target}). 
% IK configuration loss returns $c(q)$ and $q$ after constraint resolution.
%$K_{\text{pos}}$, $ K_{\text{ori}}, K_{secondary}$ are the proportionality step-size gain constants. 
%Within each step $\Delta \bm{q}_{\text{primary}}$ optimizes the primary end-effector pose constraint to minimize the end-effector pose error and $\Delta \bm{q}_{\text{secondary}}$ provides redundancy resolution to remain close to $\bm{q}_{nom}$ within the null space of the primary pose constraint. This resolves the system's redundancy while allowing full use of the joints for local dexterous motions. 

\begin{algorithm}[tbp!]

\DontPrintSemicolon
\SetKwInput{Param}{Parameters}
\SetKw{break}{break}
 \KwIn{$\bm{T}_{tar}$: Target Needle Insertion transform, $\bm{q}_0$: initial oint configuration}
 \Param{
 $c(\cdot)$: cost function for a configuration; $\text{FK}(\cdot)$: forward kinematics function; $\bm{J}(\cdot)$: space Jacobian ; $\tilde{\bm{J}}(\cdot)$: non-redundant space Jacobian; $\epsilon_p$: position error tolerance for IK solution; $\epsilon_o$: orientation error tolerance for IK solution; $\bm{K}_\text{e}$ end-effector task gain matrix; $\bm{K}_\text{c}$ null-space task gain matrix}
 \KwOut{IK configuration loss $c( \bm{q} )$, IK configuration $\bm{q}$}
\SetKwRepeat{Do}{do}{while}
\tcp{Solve IK}
$\bm{q} \gets \bm{q}_0$ \par
\Do{$\left\| e_p \right\|_2 > \epsilon_p \text{ and } \left\| e_o \right\|_2 > \epsilon_o$}
{

    $\bm{e} \gets 
    \begin{bmatrix}
        \bm{d}_\text{p} (\bm{T}_{tar}, \text{FK}(\bm{q}))  &
        \bm{d}_\text{o} (\bm{T}_{tar}, \text{FK}(\bm{q}))
    \end{bmatrix}^\top$ \par

    $\tilde{\bm{q}} \gets \tilde{\bm{q}} + \tilde{\bm{J}}^{\dag} \bm{K}_\text{e}\bm{e}$ \par
}
     
\tcp{Evaluate adjustability}
\lIf{$\bm{q} \not \in C_{free}$}{\textbf{return} $c_{infeasible}, \; \bm{q}_{0}$}
$X_{adj} \gets \text{\sc calcLocalTargets} (\bm{T}_{tar})$ \par
\ForEach{$\bm{T}_{local} \in X_{adj}$}
{
    $\bm{q}_{local} \gets \bm{q}$ \par
    \Do{$\left\| \bm{e}_\text{p} \right\|_2 > \epsilon_\text{p} \text{ and } \left\| \bm{e}_\text{o} \right\|_2 > \epsilon_\text{o}$}
     {
         $\bm{e} \gets 
            \begin{bmatrix}
                \bm{d}_\text{p} (\bm{T}_{local}, \text{FK}(\bm{q})) & 
                \bm{d}_\text{o} (\bm{T}_{local}, \text{FK}(\bm{q}))
            \end{bmatrix}^\top $ \par
        $\bm{q}_{local} \gets \bm{q}_{local} + \bm{\bm{J}}^{\dag}\bm{K}_\text{e} \bm{e} + (I-\bm{\bm{J}}^{\dag}\bm{\bm{J}})\bm{K}_\text{c}( \bm{q}-\bm{q}_{local} )$ \par
        \lIf{$\bm{q}_{local} \not \in C_{feas}$}{\textbf{return} $c_{infeasible}, \; \bm{q}_{0}$}
     }
}
\textbf{return} {$c(\bm{q}) , \; \bm{q}$}

\caption{IKConfigurationLoss}
\label{alg:configuration_loss}
\end{algorithm}

% \begin{algorithm}[tb]
% \DontPrintSemicolon
% \SetKwInput{Param}{Parameters}
% \SetKw{break}{break}
%  \KwIn{$\tilde{\bm{x}}$: target pose $\in SE(3)$; \\$\bm{x}$: current pose $\in SE(3)$}
%  \KwOut{pose error vector $\bm{e}_\text{pos} \in \mathbb{R}^3$}
% $\tilde{\bm{y}} \gets \tilde{\bm{x}}_{1:3,4}$, $\bm{y} \gets \bm{x}_{1:3,4}$ \par
% $\bm{e}_\text{pos} \gets \tilde{\bm{y}} - \bm{y}$ \par
% \textbf{return} {$\bm{e}_\text{pos}$}

% \caption{posDiff}
% \label{alg:position_error}
% \end{algorithm}

% \begin{algorithm}[tb]
% \DontPrintSemicolon
% \SetKwInput{Param}{Parameters}
% \SetKw{break}{break}
%  \KwIn{$\tilde{\bm{x}}$: target pose $\in SE(3)$; $\bm{x}$: current pose $\in SE(3)$}
%  \KwOut{orientation error vector $\bm{e}_\text{ori} \in \mathbb{R}^3$}
% $\tilde{\bm{z}} \gets \tilde{\bm{x}}_{1:3,3}$, $\bm{z} \gets \bm{x}_{1:3,3}$ \par
% $\bm{e}_{ori} \gets \cos^{-1} \left( \frac{\tilde{\bm{z}}^\top \bm{z}}{\|\tilde{\bm{z}}\|_2 \|\bm{z}\|_2} \right) \left( \tilde{\bm{z}} \times \bm{z} \right) $ \par
% \textbf{return} {$\bm{e}_\text{ori}$}

% \caption{oriDiff}
% \label{alg:orientation_error}
% \end{algorithm}

 \begin{algorithm}[tb]
\caption{calcLocalTargets}
\label{alg:local_configs}

\DontPrintSemicolon
\SetKwInput{Param}{Parameters}
\SetKw{break}{break}
 \KwIn{$\bm{T}_\text{nom}$: nominal transform $\in SE(3)$}
\Param{ $\Delta_\text{zenith}$: max zenith angle off nominal insertion axis for adjustability evaluation; $N$: number of target vectors to evaluate in radial direction; $M$: number of target vectors to evaluate in the zenith direction
}
 \KwOut{$\bm{X}_\text{adj} \in SE(3) \times N M$: target poses for adjustability evaluation}
% $y \gets \bm{x}_{1:3,4}$,  $z \gets \bm{x}_{1:3,3}$ \par
$\Delta \bm{R}_{zenith} \gets \text{\sc rotMat}(x,  \Delta_\text{zenith} / M)$ \par % \algorithmiccomment{rotation around $x$-axis} \par
$\Delta \bm{R}_{radial} \gets \text{\sc rotMat}(z, 2 \pi / N)$ \par %\algorithmiccomment{rotation around $z$-axis} \par
 $\bm{X}_\text{adj} \gets \emptyset$ \par
 $\bm{T}_\text{local} \gets T_\text{nom}$
\For {$i = 1$ to $N$}
{
    $\bm{R}_\text{local} \gets \Delta \bm{R}_{zenith} \bm{R}_\text{local}$ \par
    $\bm{X}_\text{adj} \gets \bm{X}_\text{adj} \cup \{\bm{T}_\text{local} \}$ \par
    \For {$j = 1$ to $M$}
    {
        $\bm{R}_\text{local} \gets \Delta \bm{R}_{radial} \bm{R}_\text{local}$ \par
        $\bm{X}_\text{adj} \gets \bm{X}_\text{adj} \cup \{\bm{T}_\text{local} \}$ \par
    }
}
\textbf{return} {$\bm{X}_\text{adj}$}
\end{algorithm}

% \begin{figure}[!t]
% \centering
% \subfloat[Prioritized patient avoidance, $\beta < 0.5$]{\includegraphics[width=0.45\linewidth]{images/equal_bore_and_patient_avoidance.png}%\label{fig:three sin x}}
% \hfill
% \subfloat[Equal priority patient and scanner avoidance, $\beta = 0.5$]{\includegraphics[width=0.45\linewidth]{images/patient_avoidance.png}%\label{fig:three sin x}%%\label{fig:five over x}}
%      \caption{Solution of multiple joint configurations for a Target Needle Insertion transform based on optimization weights and utilizing CRANE's redundant in-bore joints. CRANE provides multiple solutions for a Target Needle Insertion transform, considering physician preferences on obstacle avoidance while ensuring a well-conditioned dexterous solution.}
% \label{fig:solution_multiple_joint_configurations_target}
% \end{figure}
% \input{algorithms}

\vspace{-3mm}

\subsection{Local Controller}\label{section:controller}
The local controller minimizes the distance between $\bm{T}^\text{b}_\text{tn}$ and estimated EE pose, $\hat{\bm{T}}^\text{b}_\text{EE}$, calculated from the measured tip-mounted magnetic tracker's pose for feedback to compensate for system mechanical deflection (described in Section \ref{section:statics}). This is calculated as:
%transformed to the needle trip pose as the current robot tip position 
\begin{equation}
    \hat{\bm{T}}^\text{b}_\text{EE} = \bm{T}^\text{b}_\text{mb} \bm{T}^\text{mb}_\text{mtrkt} \bm{T}^\text{trk}_\text{EE}
\end{equation}
%calculated as$\text{\sc calcPoseError} (\bm{T}^b_{tn}, \tilde{\bm{T}}^{b}_{nt})$ 
Additionally, the local controller minimizes the distance between the estimated joint configuration, $\hat{\bm{q}}$, and $\bm{q}^\ast$ via a nullspace controller. 
%This method provides for minor needle adjustments during the procedure and is used in the robot's high-frequency control loop, where it is called once per timestep.
Specifically, the joint configuration setpoint, $\bm{q}_{set}$, is updated following:  
\begin{equation}\label{eq:pose_error}
    \bm{e} \gets 
    \begin{bmatrix}
        \bm{d}_\text{p} (\bm{T}^\text{b}_\text{tn}, \hat{\bm{T}}^\text{b}_\text{EE})  &
        \bm{d}_\text{o} (\bm{T}^\text{b}_\text{tn}, \hat{\bm{T}}^\text{b}_\text{EE})
    \end{bmatrix}^\top 
\end{equation}
\begin{equation}
    \bm{q}_\text{set} \gets
    \hat{\bm{q}} + \bm{J}^{\dag} \bm{K}_\text{e} \bm{e} + (\bm{I}-\bm{J}^{\dag}\bm{J})\bm{K}_\text{c}(\bm{q}^\ast-\hat{\bm{q}}) 
\end{equation}
\noindent where $\bm{J}^{\dag}=\left( \bm{J}^\top\bm{J}+ \lambda \bm{I} \right)^{-1} \bm{J}^\top$ with damping term $\lambda$ evaluated at the estimated joint configuration, $\bm{K}_\text{c}$ and $\bm{K}_\text{c}$ are end-effector and null-space task gain matrices, $\hat{\bm{q}}$, the target pose described by $\bm{T}^\text{b}_\text{EE}$ defined above and with $\bm{T}^\text{b}_\text{tn}$ provided from the User Interface. The estimated joint configuration, $\hat{\bm{q}}$, is described in the following section.

\subsection{Robot Joint Control}
%resulting in 
The robots joint configuration estimate, $\hat{\bm{q}} =    
    \begin{bmatrix}
    \hat{\bm{q}}_\text{b} & \hat{\bm{q}}_\text{c} 
    \end{bmatrix}^{\top} $, is calculated as:
\begin{equation}
    \hat{\bm{q}} \leftarrow  
    \begin{bmatrix}
     \bm{M}_\text{b} \bm{\theta}_\text{b} \\
    \alpha \bm{M}_\text{c} \dot{\bm{\theta}}_\text{c} \Delta T + (1- a) \tilde{\bm{q}}_\text{c} \\
    \end{bmatrix}
\end{equation}
for motor position, $\bm{\theta}$, motor velocity $\dot{\bm{\theta}}$, sampling time, $\Delta T$, and weighting parameter, $\alpha$, setting the complementary filter's changeover frequency.
The base joint positions $\bm{q}_\text{b}$ are calculated based on the gear ratios and can be directly used for control due to the transmission's high stiffness. The in-bore cable-driven joint configuration, $\hat{\bm{q}}_\text{c}$, is estimated using a complementary filter between the joint mounted magnetic encoders (AMS AS5048B),
$\tilde{\bm{q}}_\text{c}$, and motor's velocity, $\dot{\bm{\theta}}_\text{c}$, (Maxon ENX encoder, Maxon ESCON 50/5) to compensate for errors in the coupling relations $\bm{M_\text{c}}$ due to the cable transmission's stretch and hysteresis, and the joint encoder's noisy readings.

Motor torque setpoints, $\bm{\tau}_\text{set}$, are calculated based on the error between the target joint configuration, $\bm{q}_\text{set}$, and the current joint configuration estimate, $\hat{\bm{q}}$, as: 
\begin{equation}\label{eq:joint_controller}
\bm{\tau}_{set} =\left(\bm{K}_\text{p}+\bm{K}_\text{d}\frac{d}{dt_{}}\right) \bm{M}^{-1} \big( \bm{q}_\text{set}-\hat{\bm{q}} \big)
\end{equation}
where $\bm{K}_\text{p}$ and $\bm{K}_\text{d}$ are the actuator-space proportional and derivative gains.

\section{Experiments and Results}

\subsection{Simulated Dexterity Analysis}
Using the Automated Device Setup method described previously, CRANE's kinematic and static dexterity is evaluated across several simulated environments. This method enables the evaluation of kinematic designs for in-bore surgical manipulation, considering the limited space and variety of patient body habitus and insertion points that may be encountered clinically. Two environment styles are tested: retrospective clinical cases for needle biopsy in a scanner and comprehensive simulated cases. CoppelliaSim\cite{rohmer2013coppeliasim} with PyRep\cite{james_pyrep_2019} bindings provides distance-to-collision, $d_\textit{B}(\cdot)$, calculations.

\begin{figure}[t]
\centering
    \begin{subfigure}[tb!]{0.85\columnwidth}
         \centering
        \includegraphics[width=\columnwidth,  trim={0cm 0cm 0 0cm}, clip]{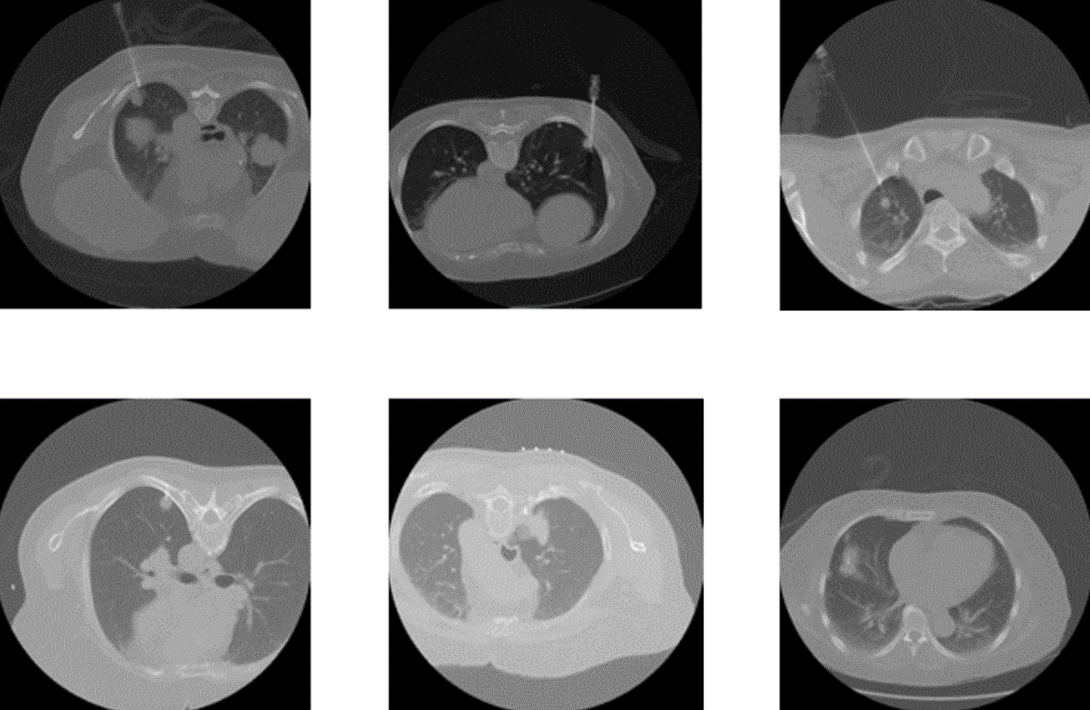}
         \caption{CT scans of clinical cases}
         %\label{fig:three sin x}
     \end{subfigure} \\
      \vspace{1mm}
     \begin{subfigure}[tbp]{0.85\columnwidth}
         \centering
        \includegraphics[width=\columnwidth,  trim={0cm 0cm 0 0cm}, clip]{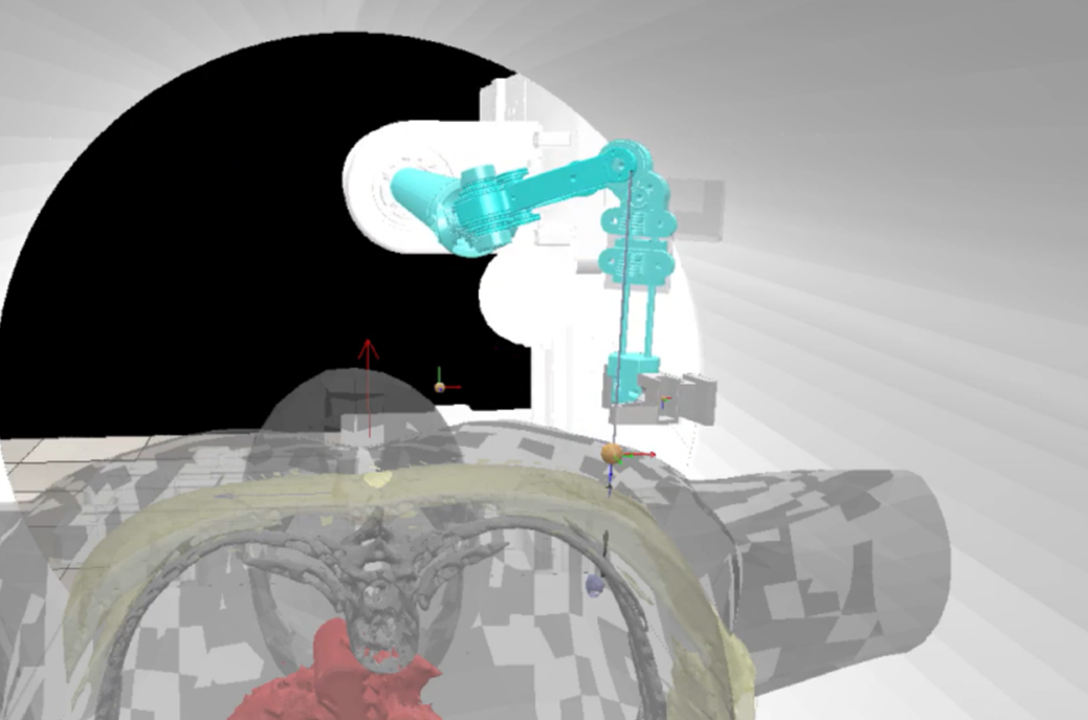}
         \caption{Simulated Case Environment}
         %%\label{fig:five over x}
     \end{subfigure}
\caption{Six clinical cases DICOM scans are used for retrospective dexterity analysis by creating a virtual reconstruction of the procedures within a virtual environment. (a) shows single axial CT slices of each clinical case with needle inserted. Several needle insertion trajectories are out-of-plane providing limited visability in the slice. (b) shows a single representative setup of 3D CT scanner room with the patient placed in bore and robot reaching the target needle insertion pose performed by the physician and determined from the CT scan. The patient body is filled in based on height and weight information combined with direct mesh matching. CRANE has sufficient workspace and dexterity to perform these clinical cases matching the needle insertion performed by a physician.}
\label{fig:single_representative_setup_3d_ct}
\vspace{-3mm}
\end{figure}

\begin{figure*}[!t]
\centering
    \centering
     \begin{subfigure}[b]{0.9\textwidth}
         \centering
         \includegraphics[width=\textwidth]{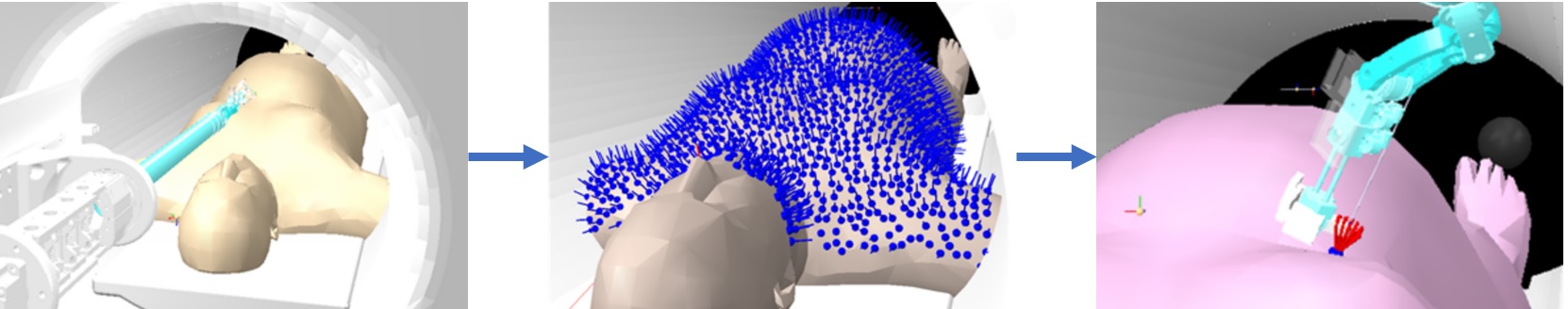}
         \caption{Workflow for comprehensive evaluation}
         %\label{fig:y equals x}
     \end{subfigure} \\
     \begin{subfigure}[b]{0.15\textwidth}
         \centering
         \adjincludegraphics[width=\textwidth,trim={{.1\width} {.15\height} {.1\height} {.25\height}},clip]{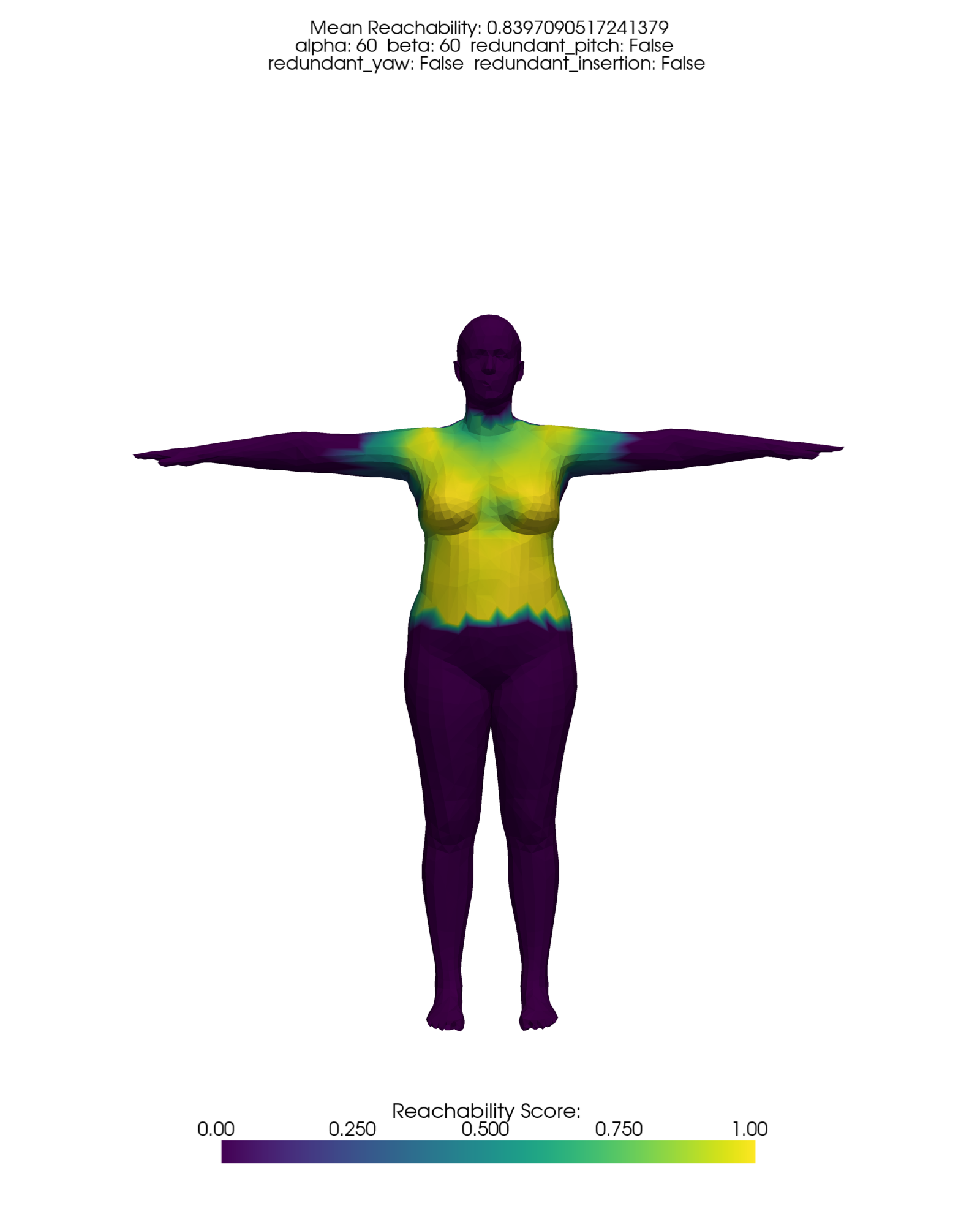}
         \caption{\textcolor{black}{Female, $+0\sigma$, \newline$84.8\%$ reachable}}
         %\label{fig:y equals x}
     \end{subfigure}
     \hfill
     \begin{subfigure}[b]{0.15\textwidth}
         \centering
         \adjincludegraphics[width=\textwidth,trim={{.1\width} {.15\height} {.1\height} {.25\height}},clip]{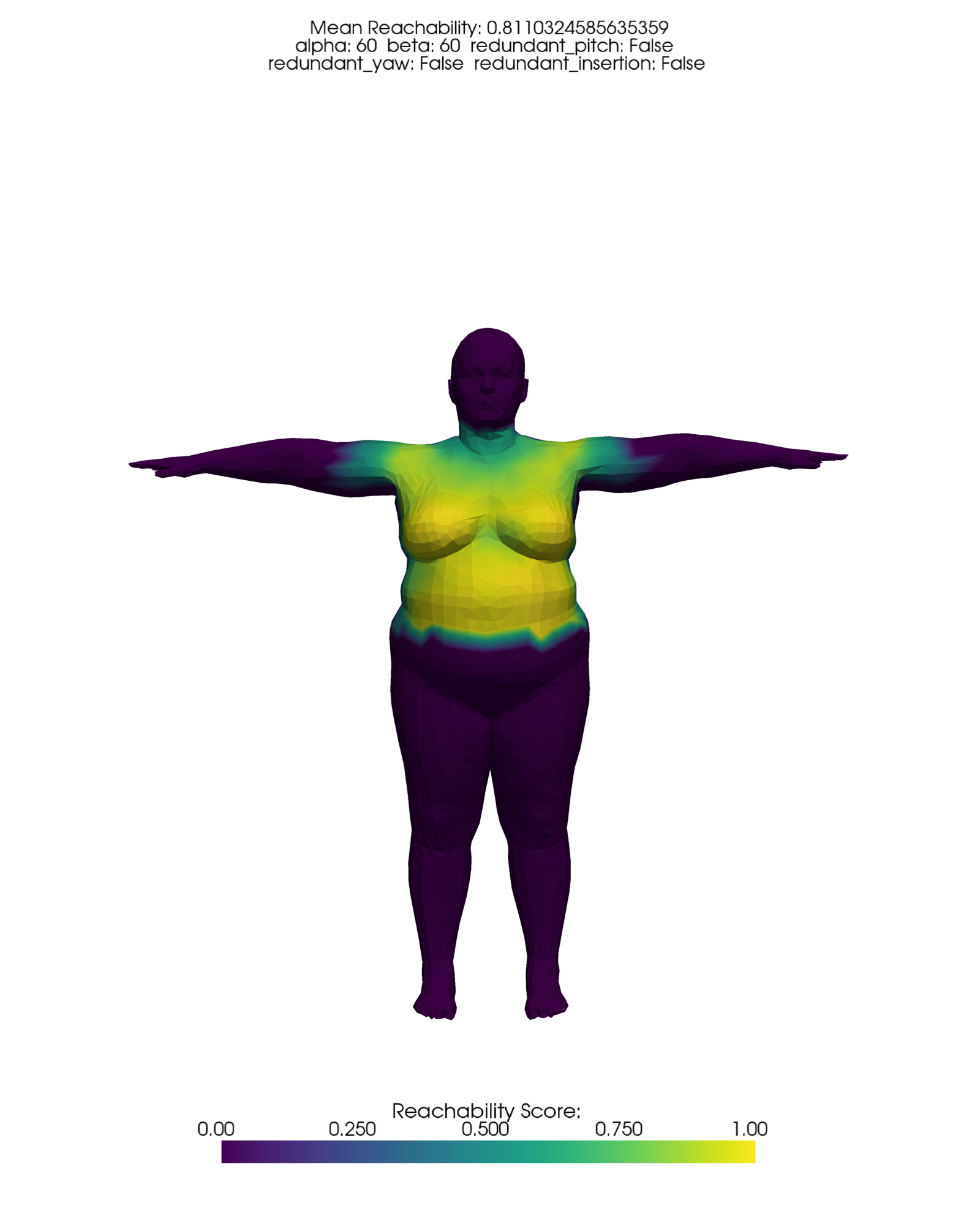}
         \caption{\textcolor{black}{Female, $+1.5\sigma$, \newline$80.0\%$ reachable}}
         %\label{fig:three sin x}
     \end{subfigure}
     \hfill
     \begin{subfigure}[b]{0.15\textwidth}
         \centering
         \adjincludegraphics[width=\textwidth,trim={{.1\width} {.15\height} {.1\height} {.25\height}},clip]{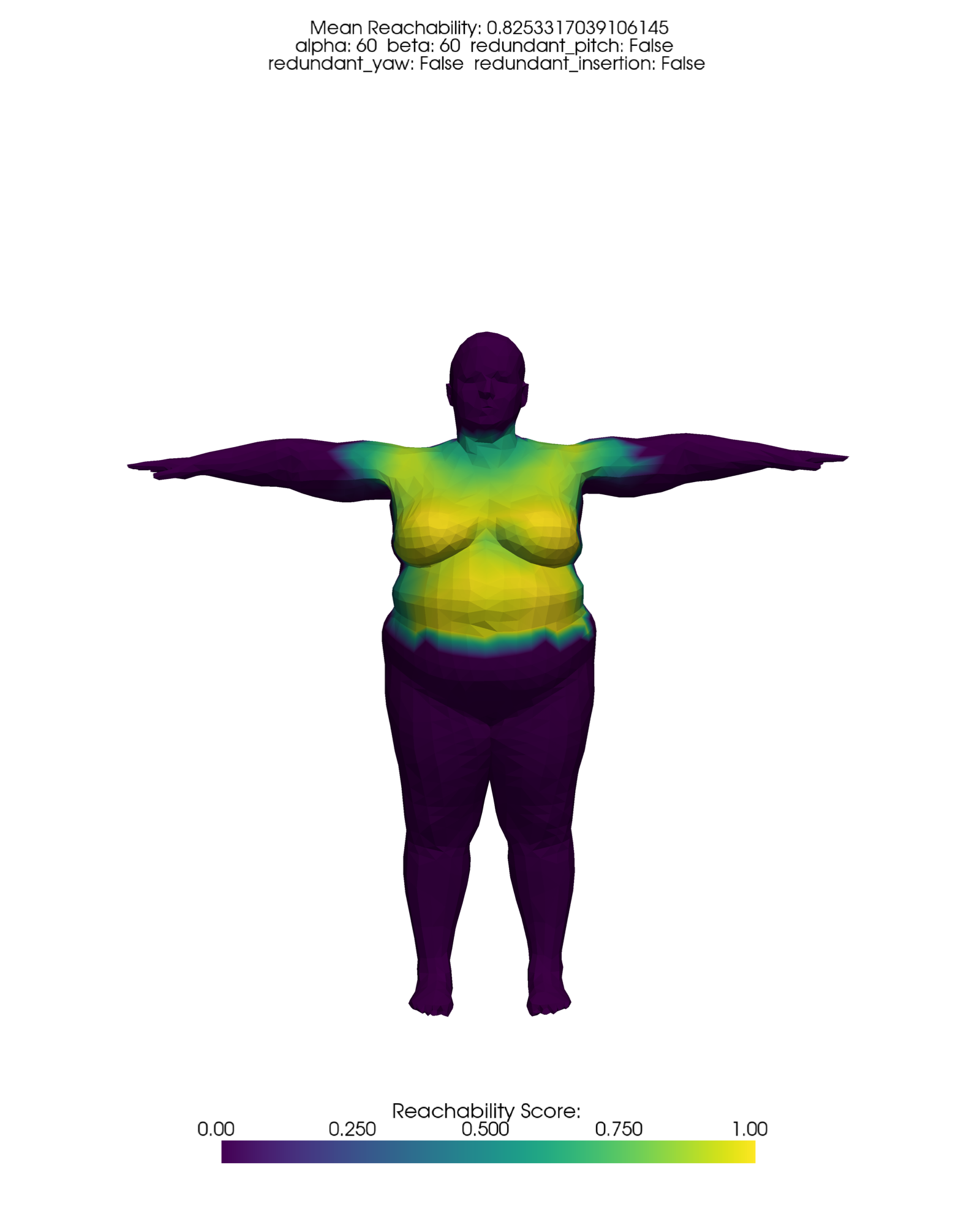}
         \caption{\textcolor{black}{Female, $+3\sigma$, \newline$80.6\%$ reachable}}
         %%\label{fig:five over x}
     \end{subfigure}
     \begin{subfigure}[b]{0.15\textwidth}
         \centering
         \adjincludegraphics[width=\textwidth,trim={{.1\width} {.15\height} {.1\height} {.25\height}},clip]{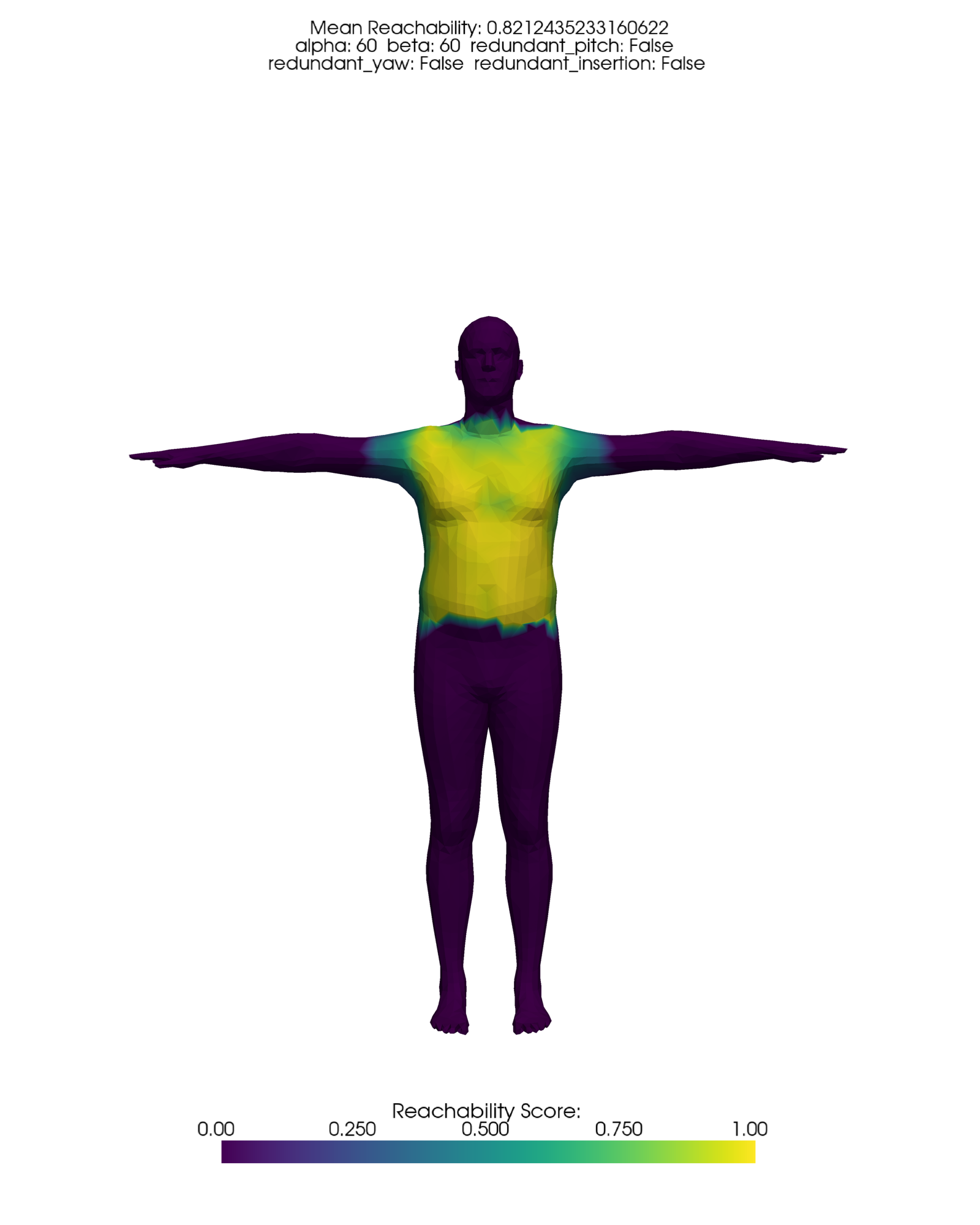}
         \caption{\textcolor{black}{Male, $+0\sigma$, \newline$84.1\%$ reachable}}
         %\label{fig:y equals x}
     \end{subfigure}
     \hfill
     \begin{subfigure}[b]{0.15\textwidth}
         \centering
         \adjincludegraphics[width=\textwidth,trim={{.1\width} {.15\height} {.1\height} {.25\height}},clip]{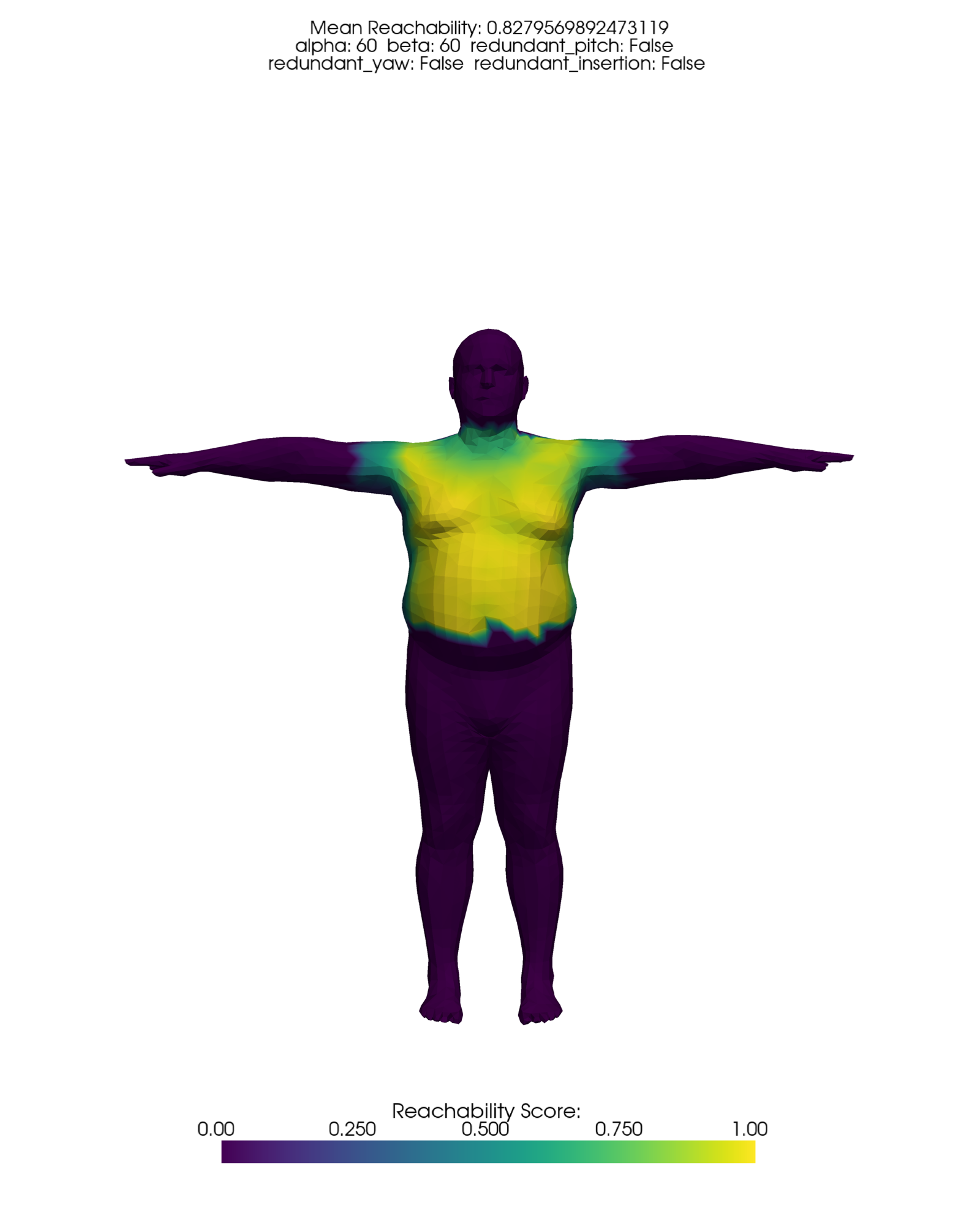}
         \caption{\textcolor{black}{Male, $+1.5\sigma$, \newline$82.8\%$ reachable}}
         %\label{fig:three sin x}
     \end{subfigure}
     \hfill
     \begin{subfigure}[b]{0.15\textwidth}
         \centering
         \adjincludegraphics[width=\textwidth,trim={{.1\width} {.15\height} {.1\height} {.25\height}},clip]{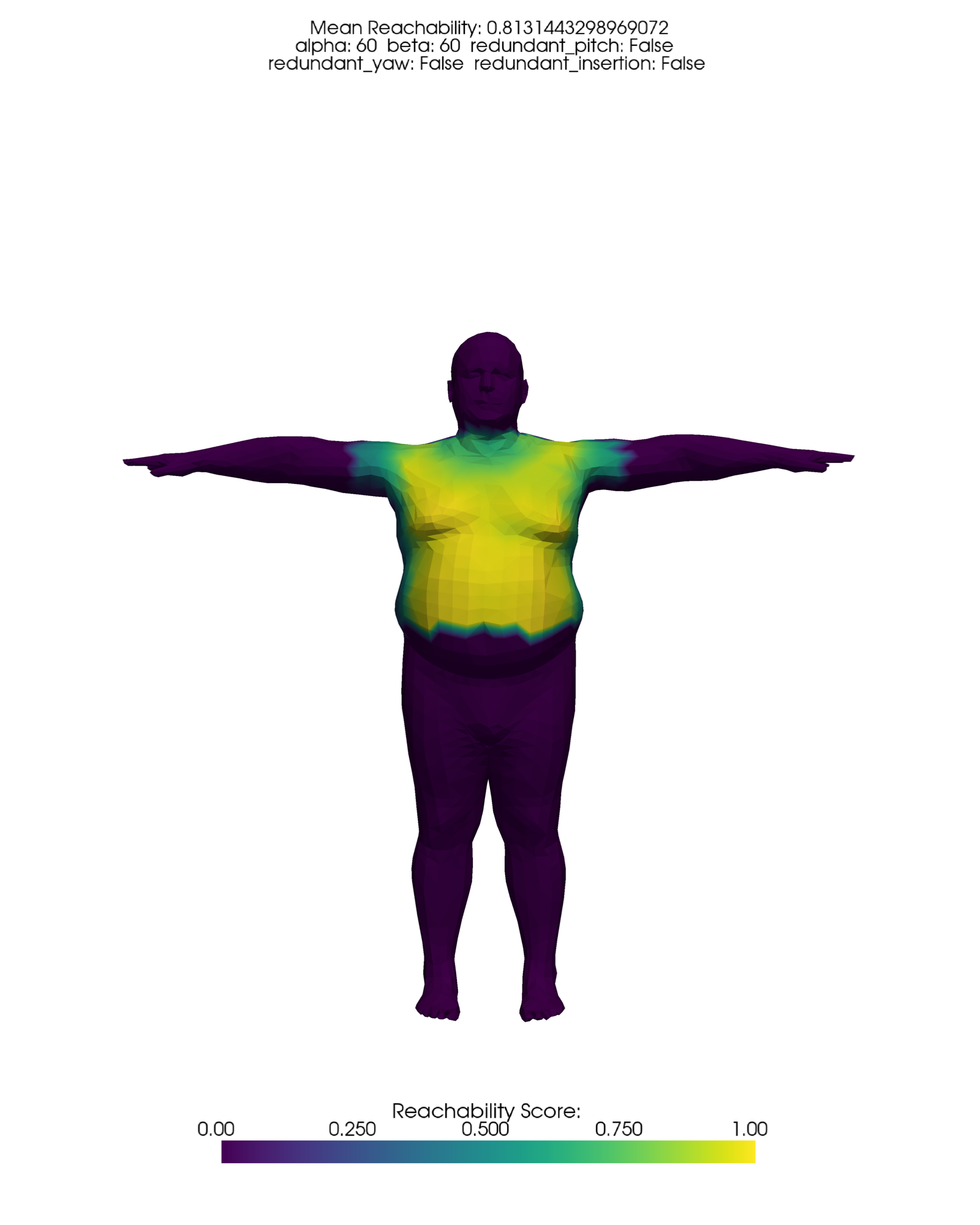}
         \caption{\textcolor{black}{Male, $+3\sigma$, \newline$80.8\%$ reachable}}
         %%\label{fig:five over x}
     \end{subfigure}
     \\
     \centering
     \begin{subfigure}[b]{\textwidth}
         \adjincludegraphics[width=1.0\textwidth,trim={{.00\width} {0} {.0\height} {0.0}},clip]{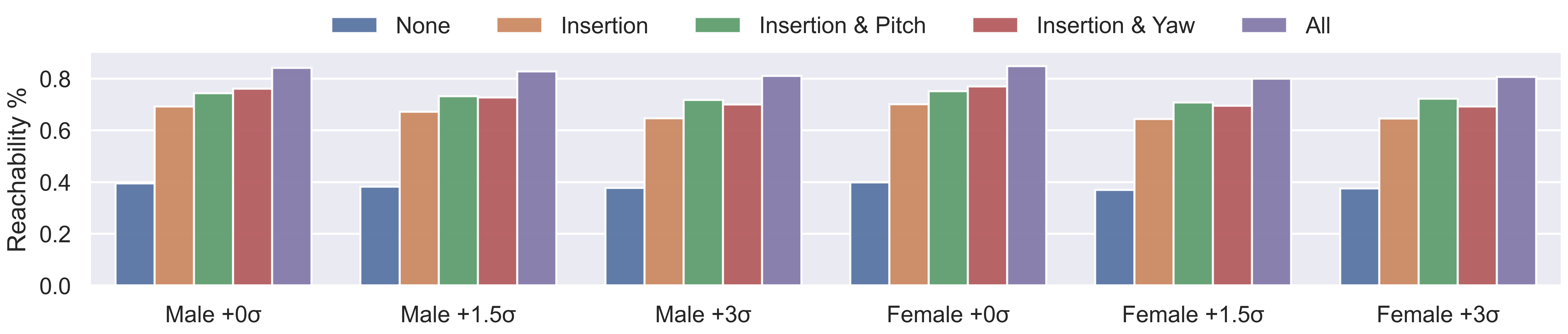}
         \caption{\textcolor{black}{Reachability evaluation for corresponding vertical figures}}
         %\label{fig:y equals x}
     \end{subfigure}
     
    %  \adjincludegraphics[height=4.7cm,trim={{.6\width} {.52\height} 0 {.05\height}},clip]{images/2022-01-12-Scene.png} %left lower right upper, https://tex.stackexchange.com/questions/57418/crop-an-inserted-image
\caption{(a) shows the evaluation workflow. An environment (including the robot, an imaging bore, and a target patient) is setup. Visible vertices areas are selected as candidate needle insertion positions with the nominal trajectory specified as the vertex normal and additional nearby orientations around the nominal normal vector. \textcolor{black}{(b-g) show results of evaluation of robot dexterity across multiple generated human mythologies targeting $60^\circ$ adjustability. The reachability metric describes the average ability of the robot to reach the candidate needle insertion trajectories. High values (shown as bright values on the patient bodies) denote high dexterity, while low values (darker areas) denote low dexterity. CRANE provides high dexterity across a wide variety of patient sizes and is able to reach across the abdominal and thoracic regions even in substantial body habitus patients. (h) shows the corresponding reachability to the above figures from (b-g) with results varying across patient bodies and robot redundant joints. Statistical significance was evaluated and is described in detail within the corresponding section.}}
\label{fig:workflow_comphrensive_evaluation_environment_is}
\vspace{-2mm}
\end{figure*}

\subsubsection{Retrospective Clinical Cases}\label{section:retrospective_clinical_cases}

The volumetric CT scans (DICOM format) of six transthoracic CT lung biopsy cases (Fig. \ref{fig:single_representative_setup_3d_ct}) from UCSD Health were segmented and integrated into a CoppelliaSim environment. The environment consists of collision meshes, $\mathcal{B} = \{\text{robot, scanner bore, patient body}\}$, and $\bm{T}^\text{sb}_\text{tn}$. Furthermore, a synthetic patient body is generated using STAR \cite{osman_star_2020} based on the patient's body habitus to fill in the portion of the patient's body not visible in the clinical scan. 

This experiment evaluates the system's ability to reach $\bm{T}^\text{sb}_\text{tn}$ for the Automated Device Setup method, while satisfying the prerequisite optimization constraints. $\bm{T}^\text{sb}_\text{tn}$ is attached to the segmented patient body which is placed within the scanner such the the DICOM's isocenter aligns with the scanner bore's isocenter. An individual $\bm{T}^\text{sb}_\text{tn} \in \mathcal{X}_\text{adj}$ is $\text{\sc reachable}$ if a dexterous configuration was found by the Automated Device Setup method and determined by $c(\bm{q}^\ast)<c_\text{infeasible}$. %$\bm{T}^\text{b}_\text{sb}$ is set by the location the robot is placed within the simulated environment. 
All six retrospective cases were reachable. This illustrates CRANE's ability to automatically setup for dexterous needle insertion within a clinical environment.%accomplished with the prescribed dexterity requirements.

\subsubsection{Comprehensive Simulated Clinical Cases}
Motivated by the retrospective clinical cases, a comprehensive test was created to enable the general purpose evaluation of an in-bore needle insertion robot's dexterity. Different sizes of human bodies were generated using STAR\cite{osman_star_2020} to $+3/-0 \sigma$ BMI for U.S. males and females. Larger patients result in less in-bore space and a more challenging environment. These human bodies are placed within the scanner bore following the procedure described above in Section \ref{section:retrospective_clinical_cases}. This test results in visual plots of a robot's ability to dexterously insert needles across a variety of patient bodies at various angles. 

For each environment's human body mesh $\in \mathcal{B}$, a set of $\bm{T}^\text{sb}_\text{tn}$, $\mathcal{X}$, is created from the mesh vertices and surface normals, $\{\mathbbm{V}, \; \mathbbm{N} \}$, of the simulated patient; defined as:
\begin{equation}
\begin{aligned}
    \mathcal{X} = \{ \bm{T} | \bm{t} = \mathbbm{v}, \; \bm{R} = \text{\sc rotMat}(\delta,\xi) \\
    \forall \{ \mathbbm{v}, \; \mathbbm{n} \} \in \{\mathbbm{V}, \; \mathbbm{N} \}
\end{aligned}
\end{equation}
where $\delta = \mathbbm{n} \times z $ and $\xi = \cos^{-1} \left( \frac{\mathbbm{n}^\top z}{\|\mathbbm{n}\|_2 \|z\|_2} \right)$ and 
$z$ is the Z-axis unit vector. % = \begin{bmatrix}        0 & 0 & 1    \end{bmatrix}^\top 
% \begin{equation}
% \begin{aligned}
%     \mathcal{X}_\text{norm} = \{ \mathcal{X} | \mathcal{X}_\text{pos} = vert, \; \mathcal{X}_\text{ori} = SO3_z(normal) \\
%     \forall \{vert, \; normal \} \in \{verts, \; normals \}
% \end{aligned}
% \end{equation}
%the Z-vector pointing into the patient%, and the X and Y orientation vectors are defined arbitrarily orthonormal to the normal vector, %$\bm{n}$ is flipped and set as the target Z-vector for a homogeneous transform. 
%and position of this transform are set to the vertex point. 
Physicians typically insert needles in an orthogonal fashion to the patient's skin to prevent slipping and needle bending from tissue boundary layers. Therefore, the surface normal vector is used as a the nominal insertion vector.
However, off-normal insertions are also performed clinically and are additionally evaluated around the nominal surface normal (shown in Fig. \ref{fig:workflow_comphrensive_evaluation_environment_is}), providing a set of insertion poses for an individual vertex connected pose $\bm{T} \in \mathcal{X}$ as:
\begin{equation}
\begin{aligned}
    \mathcal{X}_\text{adj} = \text{\sc calcLocalTargets} (\bm{T}, \; 60^{\circ}, \; 8, \; 8) 
\end{aligned}
\end{equation}
\noindent
Given this set of Target Needle Insertion points for a single vertex, $\mathbbm{v}$, the robot's ability to perform the insertion dexterously is evaluated using the Automated Device Setup method (described in Section \ref{section:automated_planning_method}) and returning $\bm{q}$. An individual $\bm{T}^\text{sb}_\text{tn} \in \mathcal{X}_\text{adj}$ is $\text{\sc reachable}$ if a dexterous configuration was found by the Automated Device Setup method and determined by $c(\bm{q}^\ast)<c_\text{infeasible}$. 

% Each vertex's, $\mathbbm{v} \in \mathbbm{V}$, color, $\bm{r}$, is set based on the percentage of reachable orientations; defined as:
% \begin{equation}
% \bm{r} = 255 \; \frac{\lVert \text{\sc reachable} (\mathcal{X}_\text{adj}) \rVert_0}{\lVert \mathcal{X}_\text{adj} \rVert_0}
% \end{equation}% with $\text{\sc reachable}$ being .

We conducted paired Student's T-tests to compare different combinations of robot types (5-DoF, 6-DoF with Insertion, 7-DoF with Insertion and Yaw, 7-DoF with Insertion and Pitch, and 8-DoF) across concatenating all of the various patient body habitus reachability performances into signal sets per robot joints used in the evaluation. A significance threshold of $p < 0.05$ was applied to assess whether these combinations belonged to the same distribution. The results indicated that the combinations 5-DoF, 6-DoF with Insertion, 7-DoF with Insertion and Yaw, 7-DoF with Insertion and Pitch, and 8-DoF had $p<0.05$ and were statistically significant. However, for the combinations 7-DoF with Insertion and Yaw and 7-DoF with Insertion and Pitch, the $p=0.578$, and we could not reject the null hypothesis of them belonging to the same distribution.

As shown in Fig. \ref{fig:workflow_comphrensive_evaluation_environment_is}, CRANE can dexterously insert needles across patients with a wide variety of body sizes and morphology. Large patients decrease the accessible dexterous region and are primarily limited by the length of the robot's final EE insertion axis, shorter than many needles used during clinical procedures and therefore not a significant limitation. This test shows that CRANE's low profile design and the redundant kinematic chain enable the dexterous insertion of a needle for clinical cases.

\subsection{Robot Experiments}
\subsubsection{Trajectory Tracking}

% The system's accuracy was evaluated by performing a virtual Remote Center of Motion trajectory where the robot revolved around a virtual needle tip location, similar to the adjustments performed by a physician in the clinic. Here, the robot's end-effector follows a cone trajectory simulating the workspace a physician would use during an actual procedure. Open loop testing resulted in $2.6^{\circ} orientation error and $6.1mm$ position error. Closed loop testing resulted in $0.71{\circ}$ orientation error and $0.27mm$ positioning error at the noise floor of the magnetic tracker used for evaluation and control.

\begin{figure}[tb!]
     \centering
     \begin{subfigure}[tb!]{0.95\columnwidth}
        \centering
        \includegraphics[width=0.9\columnwidth]{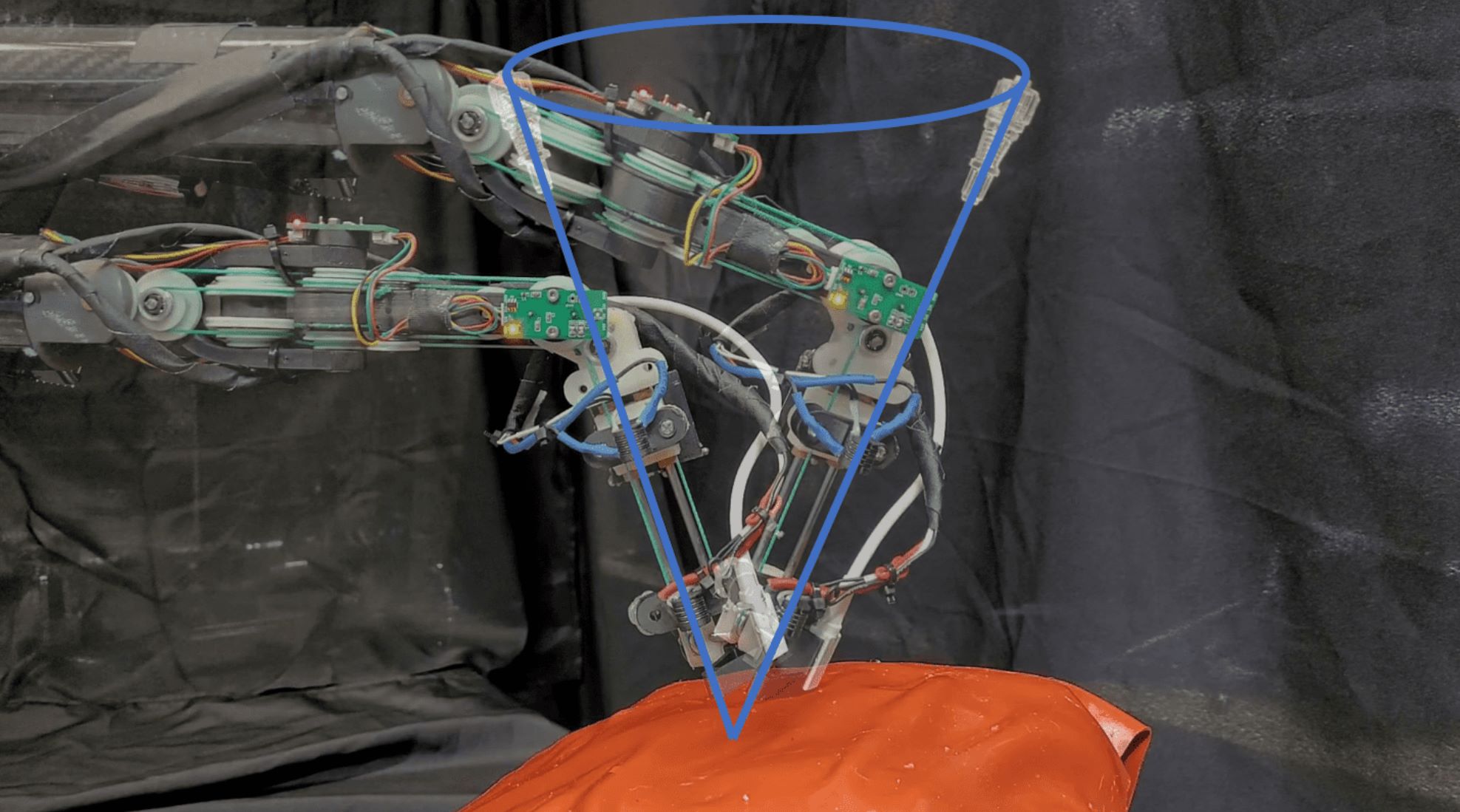}
        \caption{Illustration of RCM trajectory}
        %\label{fig:three sin x}
     \end{subfigure} \\
     \begin{subfigure}[tbp]{\columnwidth}
        \centering
        \includegraphics[width=0.9\columnwidth,  trim={0cm 0cm 0 0cm}, clip]{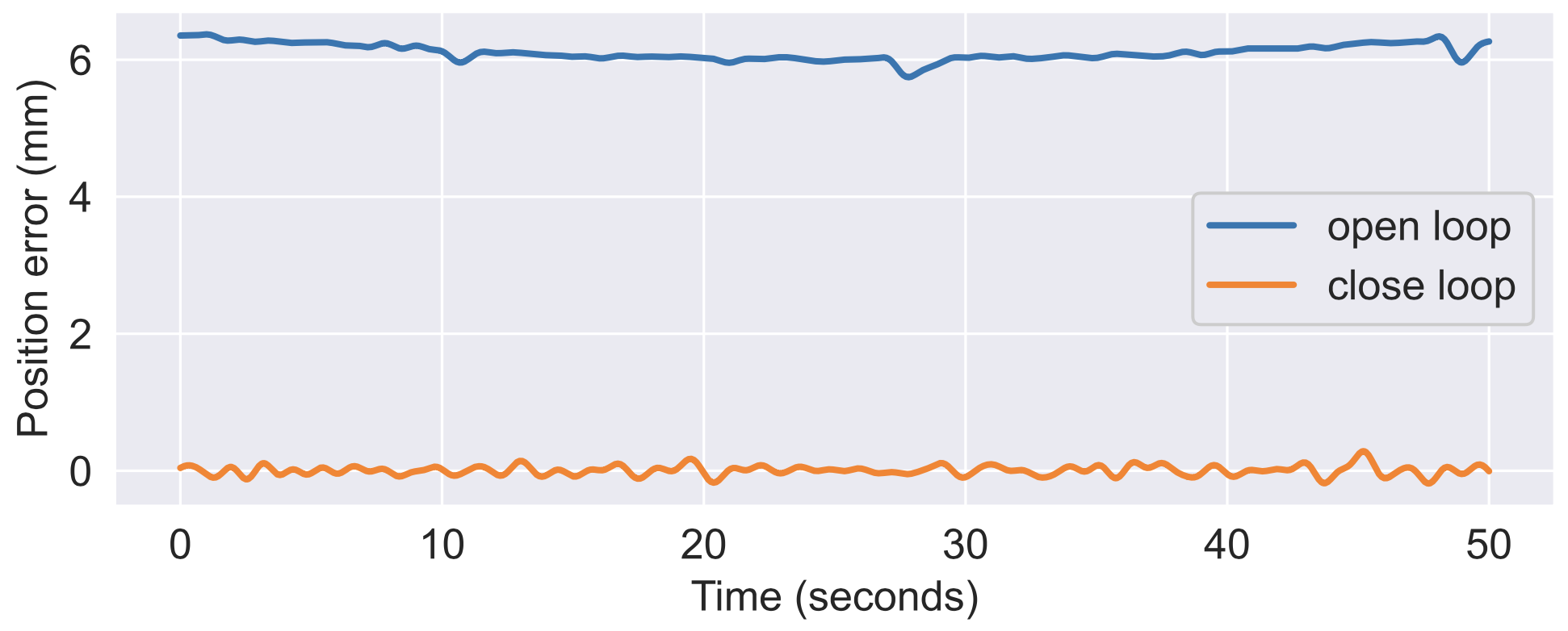}
        \caption{Position Error Plot for RCM trajectory}
        %%\label{fig:five over x}
     \end{subfigure}
     %\hfill
     \begin{subfigure}[tbp]{\columnwidth}
        \centering
        \includegraphics[width=0.9\columnwidth,  trim={0cm 0cm 0 0cm}, clip]{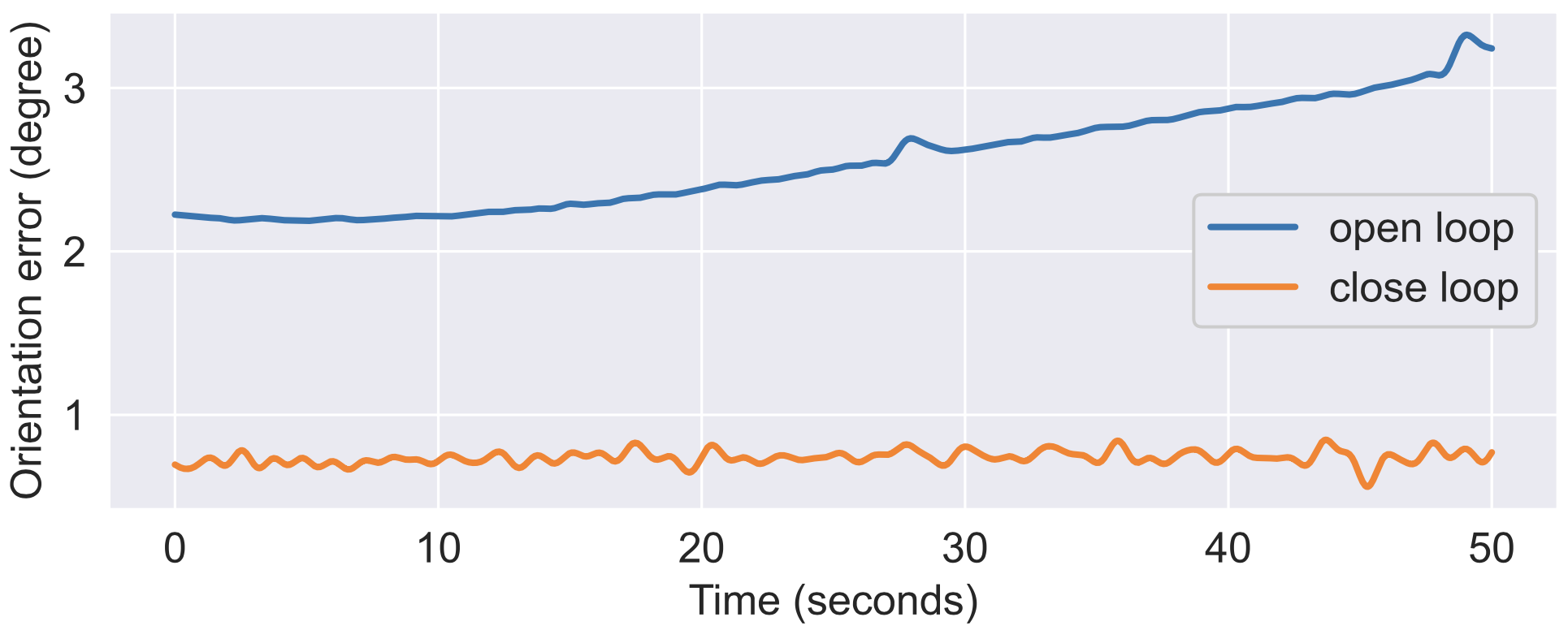}
        \caption{Orientation Error Plot for RCM Trajectory}
        %\label{fig:five over x}
     \end{subfigure}
     \caption{(a) RCM trajectory is chosen due to common use within image-guided surgery. (b) The open-loop evaluation uses motor position control without feedback from joint encoders or end-effector controllers. (c) Close-loop control runs position control using joint encoders for feedback and direct end-effector position measurement from a magnetic 6D tracker, matching the tracker's accuracy. Open-loop accuracy is surprisingly good given CRANE's long and low profile kinematic chain. Closed loop control demonstrates \textcolor{black}{excellent servoing performance and far higher accuracy}, partially enabled by the low backlash, friction, and hysteresis transmission.}
    \label{fig:robot_accuracy}
\end{figure}
The system's \textcolor{black}{open loop accuracy and closed loop servoing performance} was evaluated by performing a virtual Remote Center of Motion trajectory where the robot revolved around a virtual needle tip location using the experimental setup shown in Fig. \ref{fig:experimental_setup_benchtop_system_evaluation}. Here, the robot's end-effector follows a cone trajectory simulating the workspace a physician would use during an actual procedure. The Ascension TrackStar magnetic tracker was used for \textcolor{black}{accuracy measurement in the open-loop test and feedback reference for servoing performance in the closed loop test}. The mean resulting closed-loop servoing performace, shown as a time series in Fig. \ref{fig:robot_accuracy}, across the trajectory was $0.27mm$ and $0.71^\circ$. \textcolor{black}{The Ascension Trackstar is rated for RMS accuracy numbers of $1.4mm$ and $0.5^\circ$.}

%Two evaluations of the system's trajectory tracking accuracy were performed. 
In the open-loop test, joint and end-effector controllers were disabled. Joint angles were calculated off the ideal coupling matrix, $\bm{M}$, without compensation for cable stretch and hysteresis in the in-bore transmission. EE measurements were replaced for the $J^{-1}$ controller with predicted EE positions based on the forward kinematics of the calculated joint angles from the motor. With controllers disabled, position and orientation errors are increased due to joint tracking errors, system deflection, and manufacturing errors. \textcolor{black}{Closed loop trajectory tracking} using direct end-effector tracking enables the system to accurately reach targets despite these challenges.

\subsubsection{Needle Gripper: Slipping Force and Speed}\label{section:gripper_experiments}

Performance regarding clutching force, cycle time, insertion force, and thermal transfer to the needle are evaluated. The needle clutch is 3D printed in a nylon-carbon composite material. The clutches are designed to grip a 15-gauge needle. Slipping forces were measured using spring scales ($0-50N$ and $0-5N$ ranges). Activated slipping forces are $18N$ and $20N.$ Deactivated slipping forces are at $1N$ and $2.25N$. \textcolor{black}{This compares favorable to \cite{Walsh2008}, where $8.6N$ was the maximum achieved.}

\begin{figure}[bt!]
\centering
\begin{subfigure}[tbp]{0.45\columnwidth}
         \centering
        \includegraphics[height = 3cm]{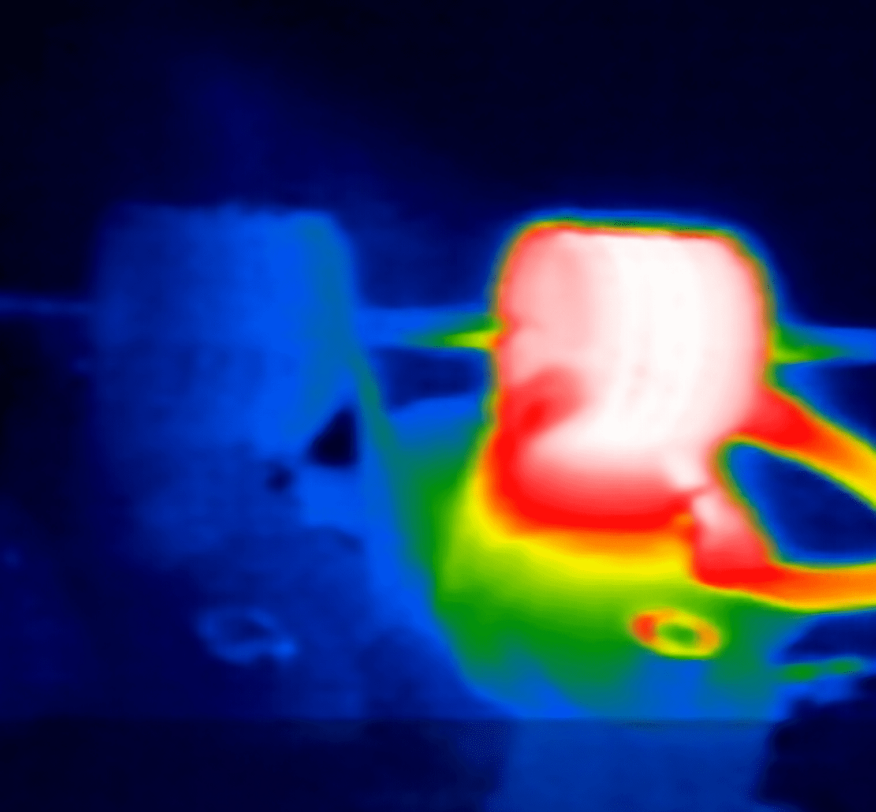}
         \caption{Thermal image}
         %\label{fig:three sin x}
     \end{subfigure}
     \begin{subfigure}[tbp]{0.45\columnwidth}
         \centering
        \includegraphics[height = 3cm, trim={0.1cm, 0.1 0.1 0.1},clip]{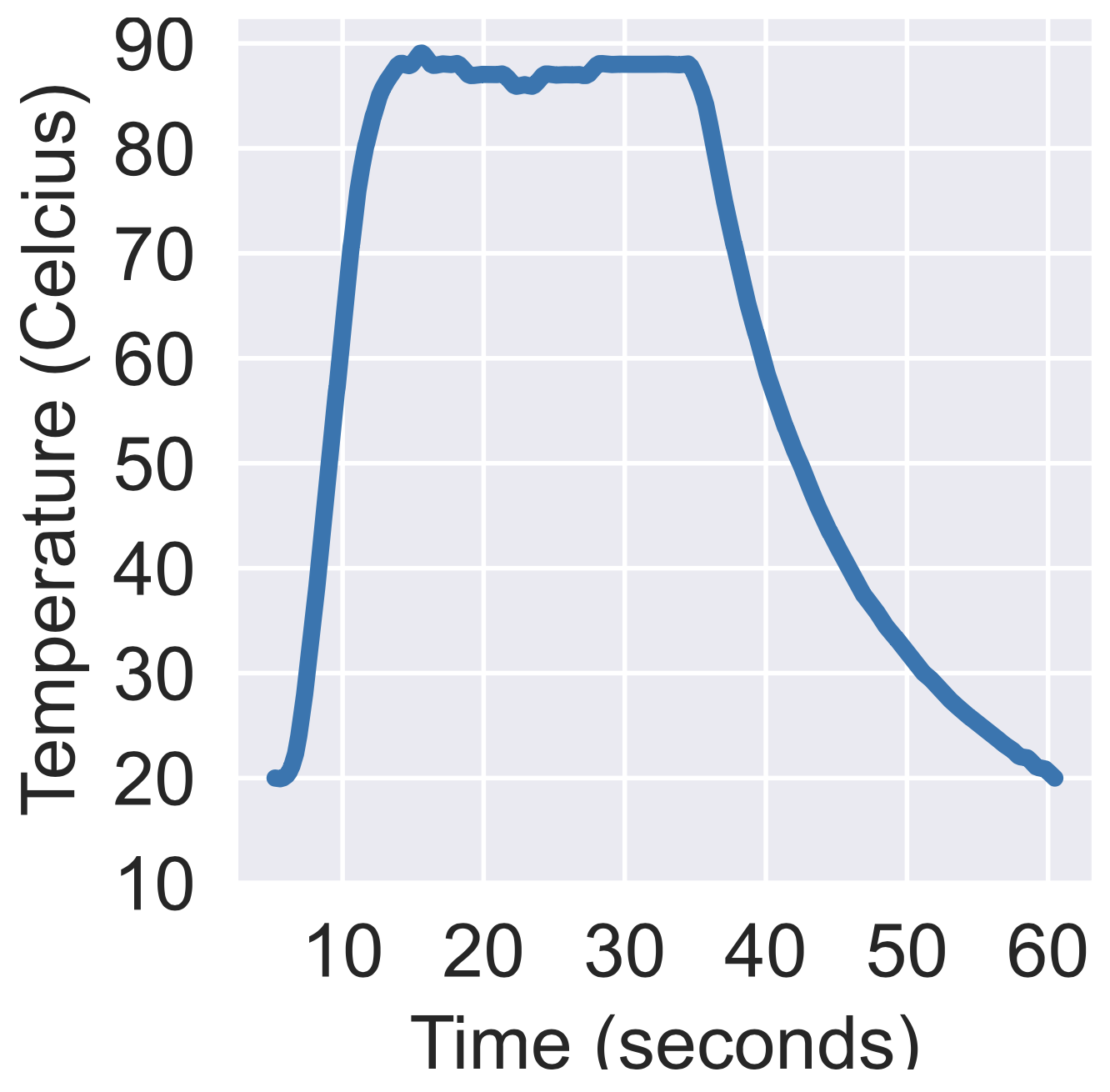}
         \caption{Step response}
         %\label{fig:five over x}
     \end{subfigure}
\caption{(a) shows a thermal image of a clutching needle driver on the robot end-effector with one clutch activated. (b) shows step response collected from the clutch with Joule heating and air-blast assisted cooling. \textcolor{black}{The clutch has a $2.5s$ risetime to $80^\circ C$ and a $10.1s$ falltime to $40^\circ C$.}
}
\label{fig:clutch_exp}
\vspace{-3mm}
\end{figure}

\begin{figure*}[tb]
\centering
     \begin{subfigure}[tbp]{0.25\textwidth}
         \centering
        \includegraphics[height=4.7cm]{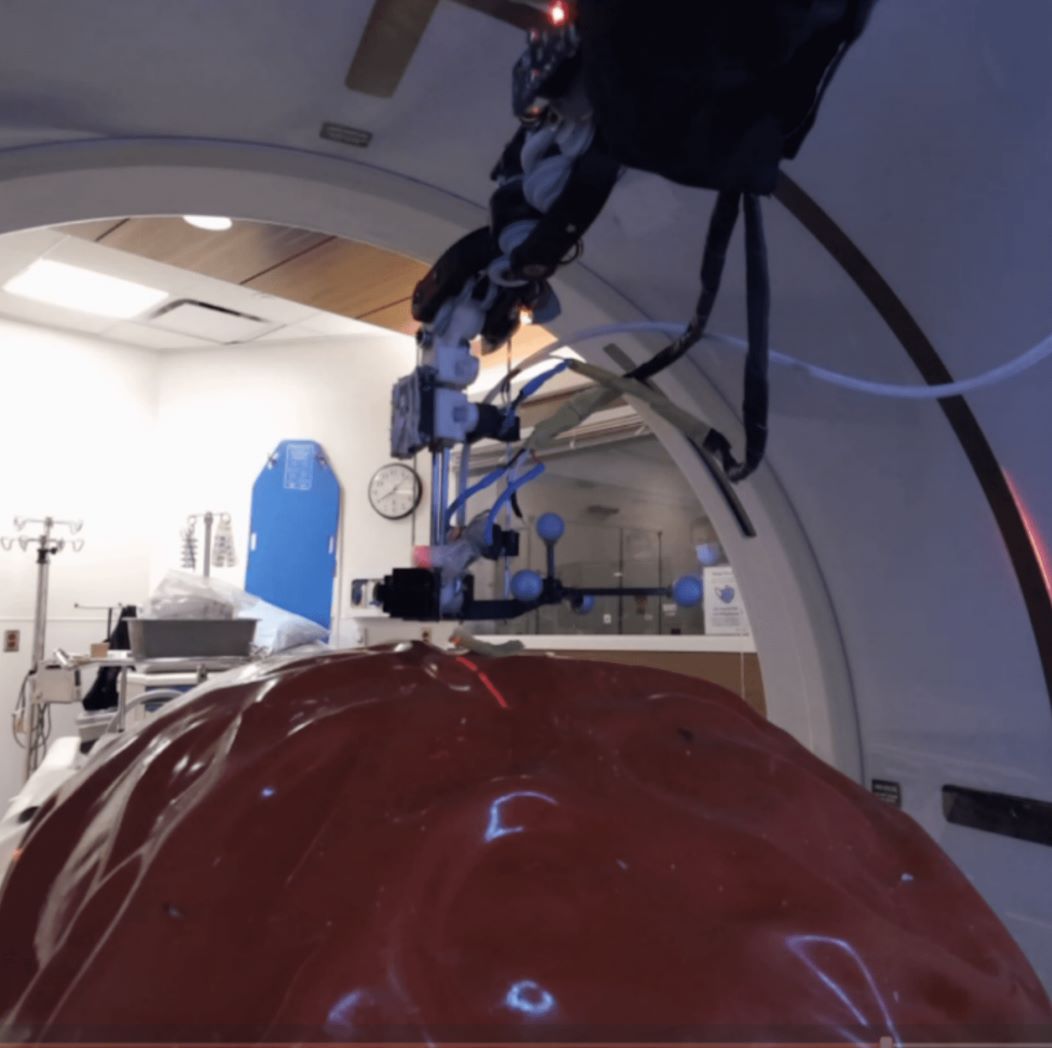}
         \caption{In-bore view}
         %\label{fig:three sin x}
     \end{subfigure}
     \hfill
     \begin{subfigure}[tbp]{0.25\textwidth}
         \centering
        \includegraphics[height=4.7cm]{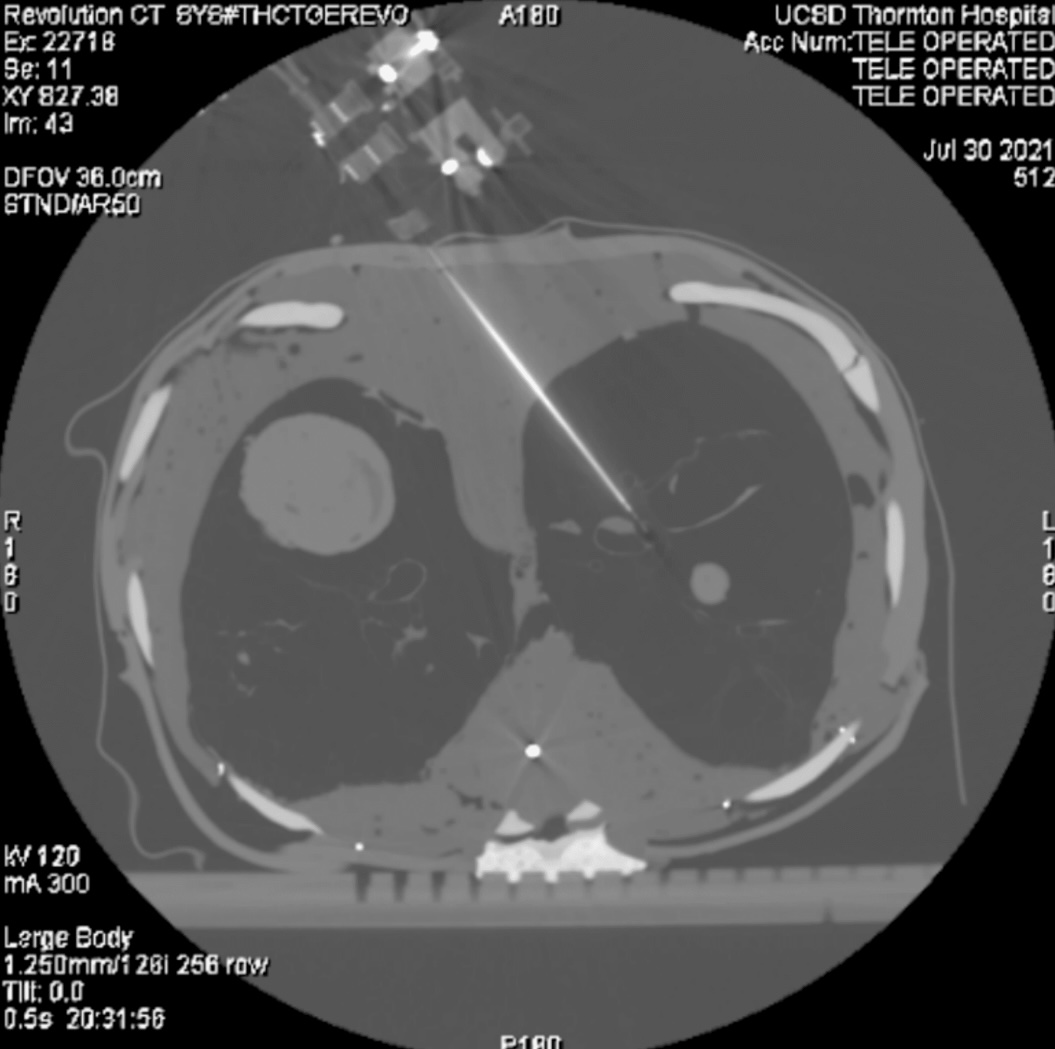}
         \caption{Intermediate CT scan}
         %\label{fig:three sin x}
     \end{subfigure}
    \hfill
     \begin{subfigure}[tbp]{0.4\textwidth}
         \centering
         \adjincludegraphics[height=4.7cm]{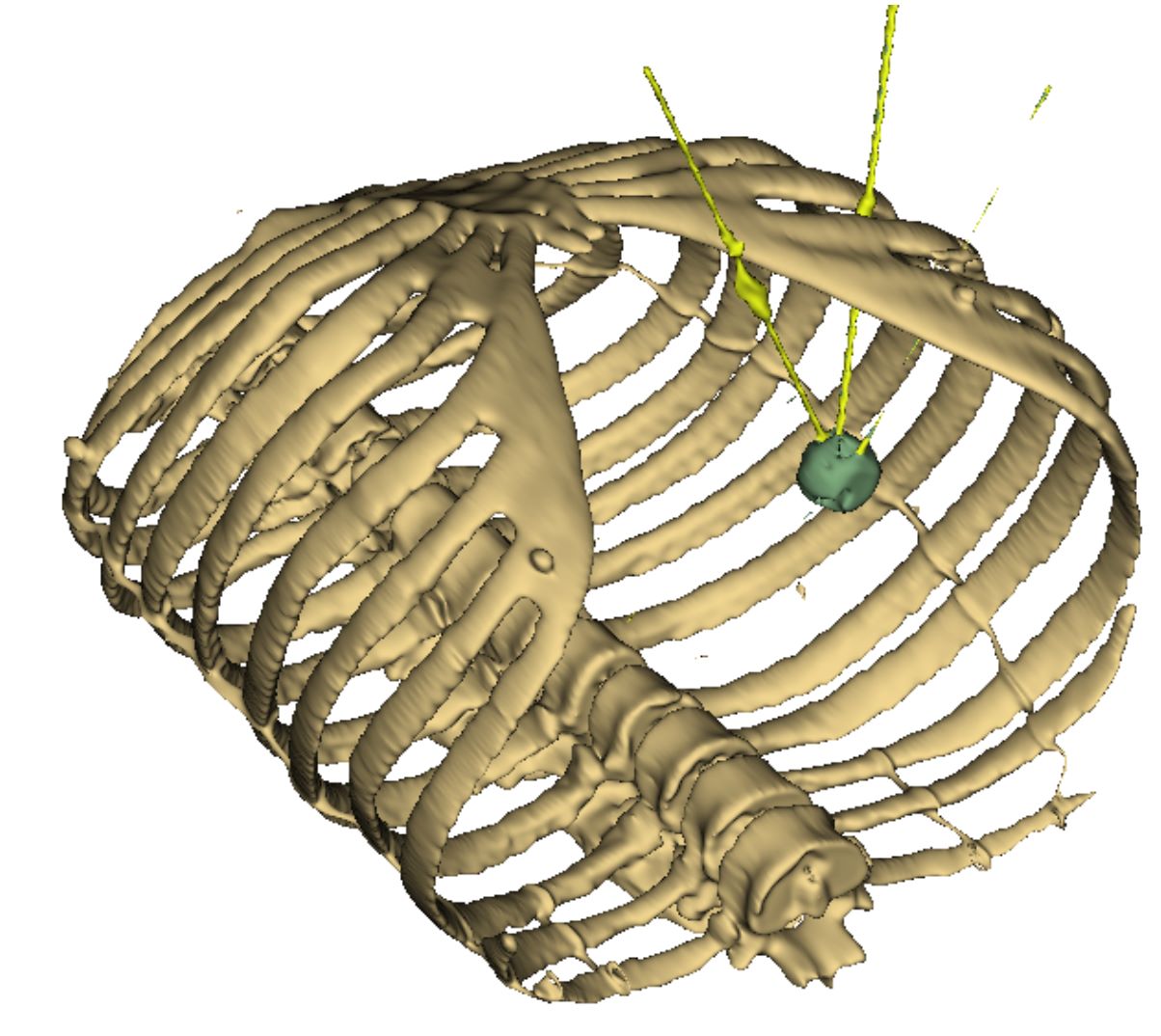} %left lower right upper, https://tex.stackexchange.com/questions/57418/crop-an-inserted-image
         \caption{Multiple Needle CT scan}
         %\label{fig:five over x}
     \end{subfigure}

\caption{(a) CRANE within a CT scanner, remotely teleoperated by an experienced radiologist and technician, visible behind the glass window. (b) CT image of the robot being teleoperated to reach a lesion in the right central lung of the phantom with the needle orientation set to result in the puncture of the tumor. (c) CT scans visualization 3D of three inserted needles into a tumor via the single-shot approach.}
\label{fig:crane_ct_scanner_remotely_teleoperated}
\vspace{-3mm}
\end{figure*}

Additionally, a step-response test was performed. The results, reported in Fig. \ref{fig:clutch_exp},  show short on and off times, implying the possibility for rapid clutching for deep needle insertion and clutch deactivation for safety.

\subsubsection{Clinical Validation: Phantom Study in CT scanner}
Multiple in-situ needle insertions were performed using CRANE and CT guidance on a custom static lung phantom to evaluate procedural feasibility. The static lung phantom comprises a plastic resin rib cage, a preserved and dried pig lung, and multiple tissue layers. Teleoperated, closed-loop image servoed, and single-shot automated procedures were performed. The needle is aligned to a target trajectory and then advanced. The null-space controller is disabled for these tests.

During the teleoperated experiment, an interventional radiologist controlled the device to advance the needle inside a CT scanner (GE Revolution, 80cm bore)  \textcolor{black}{using the clutching needle-driver technique (as demonstrated with other teleoperated systems with different needle grippers in \cite{frishman2021, ghelfi_evaluation_2018, bricault_light_2008, Frishman2020})} into a target tumor within the phantom using multiple CT scans for guidance, directly providing a sequence of target transforms for the robot to execute, which were translated to joint angles by the robot's controller.

\textcolor{black}{During the visual servoing experiment, $\bm{T}^\text{sb}_\text{tn}$ is selected based on an initial pre-operative scans and trajectory adjustments are automatically calculated based on CRF for $\bm{T}^\text{sb}_\text{EE}$ as defined in Eq. \ref{eq:pose_error}. Three control steps are applied, demonstrating the efficacy of closed-loop control and the increase in error during insertion due to link deflection without closed-loop control. Results are shown in Table \ref{table:image_guidance_results}.} 

During the single-shot automated procedure, a preoperative scan is performed from which $\bm{T}^\text{sb}_\text{tn}$ and environment obstacles, $\mathbb{B}$ are created, and a collision-free motion plan, $\bm{Q}$, from the initial robot configuration $\bm{q}_\text{start}$ to $\bm{q}^*$ is calculated and executed within a virtual imaging bore (70cm diameter). The needle is inserted to the tumor, and a post-operative scan is performed. The phantom body segmentation is determined using the Marching Cubes algorithm\cite{lorensen_marching_1987} with a threshold of $-200 HU$ and a model of the CT scanner bore. \textcolor{black}{The resulting mesh is decimated and then a convex decomposition is calculated using V-HACD.} Three tests were performed using different $\bm{T}^\text{sb}_\text{tn}$ towards a single tumor within the phantom. Table \ref{table:accuracy_experiment_single_shot} shows the results from this experiment which achieved high accuracy. The position error and angle error are calculated as the L2 norm of the position and orientation components of the position and orientation error between between the $\bm{T}^\text{sb}_\text{tn}$ and $\bm{T}^\text{sb}_\text{EE}$ as defined in Eq. \ref{eq:pose_error}. $\bm{T}^\text{sb}_\text{EE}$ is manually determined from the post-operative scan using 3DSlicer\cite{fedorov20123d}. \textcolor{black}{These results compare favorably with studies performed using other CT-guided systems, which achieved $1mm - 5mm$ error \cite{martinez_ct-guided_2014-1, Stoianovici2003, melzer_innomotion_2008, seitel_development_2009, Bricault2008}}.

\setlength{\tabcolsep}{7pt}
\begin{table}[t!]
\caption{Accuracy results from automated single-shot needle insertion illustrating high accuracy for a full depth needle insertion with single planning scan}
\label{table:accuracy_experiment_single_shot}
\vspace{-3mm}

\begin{center}
\begin{tabular}{llll}
\toprule
    \textbf{} & Trajectory 1 & Trajectory 2 & Trajectory 3 \\ \midrule
    Angle RMSE (deg) & 2.0 & 1.8 & 1.9   \\
    Position RMSE (mm) & 3.4 &  2.4 & 3.8  \\
\bottomrule
\end{tabular}
\vspace{-3mm}

\end{center}
\end{table}

\setlength{\tabcolsep}{7pt}
\begin{table}[t!]
\caption{EE pose RMSE across time for closed-loop image-space control.}
%\caption{Tracking deviation between the robot's FK and calculated tip transform from CT tracking illustrating low RMSE error at each time step and maximum error. This demonstrates the robot to scanner calibration is correct and matches the robot kinematic model.}

\label{table:image_guidance_results}
\begin{center}
\color{black}\begin{tabular}{lllll}
\toprule
    Scan & Description & Position (mm) & Orientation (deg)  \\ \midrule
    0 & Setup & $100.1$ & $96.8$\\%$0.07$\\
    1 & Control 1 & $16.3$ & $6.8$\\%$0.01$\\
    2 & Control 2 & $0.3$ & $1.1$\\%$0.09$\\  
    3 & Control 3 & $0.4$ & $0.3$\\%$0.05$\\
    4 & Partial Insertion 1 & $0.7$ & 0.6\\%$0.05$\\
    5 & Partial Insertion 2 & $1.5$ & 0.8\\%$0.16$\\
    6 & Full Insertion & $2.0$ & $1.0$\\ %\midrule%$0.07$\\\midrule
    %Max & $0.40$ & 2.42\\%$0.16$ \\
    %Min & $0.05$ & 2.26\\%$0.01$ \\
\bottomrule
\end{tabular}
\end{center}
\vspace{-3mm}
\end{table}

% \subsubsection{CT Scanner Registration Evaluation}

% \subsubsection{CT Image Interference}
% Large amounts of metal in the frame can cause reconstruction artifacts which decrease the physician's ability to interpret the image and target the lesion. Minimal artifacts are observed in our design due to the minimization of metallic components in the bore, consisting of similar quantities to the metal as already present from the needle. Contrast to noise is evaluated with and without the robot in the frame and with and without the needle in the frame on top of the realistic tissue phantom.

\section{Discussion and Conclusions}
Computed Tomography (CT) guided biopsies and ablations frequently require multiple needle insertions and repeat procedures, increasing patient risks, costs, and hospital stay lengths. Robots can eliminate the multiple punctures and procedures required while enabling physicians to treat small early-stage cancer via a minimally invasive approach. However, prior robotic platforms had complex and lengthy setups, limited applicability and large size, imaging artifacts, insufficient accuracy, and limited needle compatibility, which limited their clinical application. 

This work presents CRANE: a robust robotics platform for in-bore needle procedures. CRANE's design and methodology provide a simple and validated approach for fully automatic device setup for in-bore needle insertion, filling a critical need for the widespread use of robotics in this space. 

CRANE's low profile mechanical design maximizes in-bore workspace while the redundant joints support avoiding collisions and retaining dexterity during insertion. The solid-state needle gripper enables deep insertion into the patient and the automatic robot setup method supports a simple workflow to utilize the redundancy and workspace, removing the cognitive burden from the physician to set up the device.  Closed-loop feedback control enables high accuracy despite the deflection inherent in a long and low-profile kinematic chain. 
Furthermore, the presented metrics and method for planning and redundant robot control within an imaging bore can be applied more broadly to other system designs, contributing several critical considerations for image-guided in-bore robot design. Simulation, benchtop, and in-situ experiments demonstrate CRANE's excellent dexterity across a comprehensive clinical workspace evaluation, high accuracy, and feasibility for clinical in-situ use. 
Through these contributions, CRANE demonstrates initial validation towards solving several clinical challenges physicians face using robotics for interventional image-guided procedures. 

\textcolor{black}{The presented approach shows initial system validation with promise for future clinical testing, which must be performed, including studies in animal models with respiratory motion and across multiple users of varying experience levels. Despite the high potential, several critical limitations of the existing robot design and testing must be addressed before clinical use, including complex electronics and difficulty transporting the system, and the need for a traditional multi-slice UI for DICOM. Limitations of the current in-situ evaluation include a small study size with a single user and testing within a static environment. Future work will focus on translating this system to clinical testing (e.g., statistical user studies and animal testing).} \textcolor{black}{Finally, the IK solution used within this paper may experience local minima; future work could extend this approach to a provably global IK solution.}

CRANE demonstrates a step forward for automated device setup and control for in-bore needle insertion. The ideas presented will hopefully support the development of more accessible and accurate clinical robotic systems for interventional image-guided surgery in the future.

%As a result, image-guided procedures can be applied more frequently and safely to applications such as the diagnosis and treatment of thoracic cancer (e.g. biopsy, ablation, radio-pharmaceutical, and brachytherapy of liver, kidney, lung, lymph-node metastasis), pain management and structural management of degenerative spine conditions (e.g., lumbo-sacral nerve-block, vertobroplasty, microdiscectomy, spinal fusion), and movement disorders (e.g., deep brain stimulus for Parkinsons, essential tremor, dystonia). 

% multiple punctures and procedures currently required while enabling physicians to treat small early-stage cancer via a minimally invasive approach. However, existing robotic platforms have a difficult and long setup, limited applicability, and require hands-on control. 

%The CRANE platform design focuses on dexterity, accuracy, and safety while serving as an extensible base for clinical workflow development and evaluation. 

\section{Acknowledgments}
We thank Drs. Aryafar, Berman, Meisinger, Minocha, Taddonio, Tadros, Theilmann, and Tutton for their invaluable discussion on clinical needs with a robot and facilitation of both procedural observation and in-scanner experiments. We thank Hanpeng Jiang, Lucas Jonasch, and Julie Yu, Guosong Li, and Renjie Zhu for their work on the system's mechatronics. Finally, we thank Jacob Johnson, Florian Richter, Nikhil Das, Zih-Yun Chiu, and Yuheng Zhi for discussing for general paper review. UCSD's AIM grant partly supported system and experimental development. D. Schreiber was supported via the NSF Graduate Research Fellowship, UCSD's Accelerating Innovations to Market grant, and Air Surgical, Inc. M. Yip and A. Norbash were partly supported by NIH Award 1R01CA278703-01. 

\bibliographystyle{IEEEtran_mod}
%\bibliography{IEEEabrv,references/base_references_note, references/UpdatedTROCitations}

\bibliography{IEEEabrv,ref/main}

%\bibliography{IEEEabrv,references_zotero.bib}

\begin{IEEEbiography}
[{\includegraphics[width=1in, height=1.25in,clip,keepaspectratio]{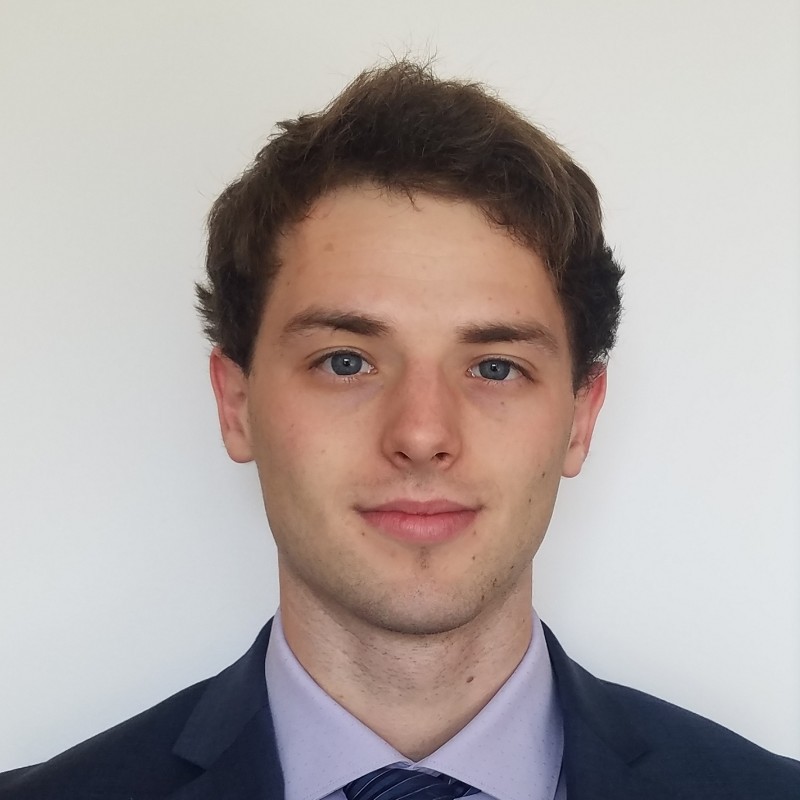}}]{Dimitrious Schreiber} is a founder of Air Surgical, Inc., a start-up company that focuses on developing and commercializing image-guided surgical robots. His research focuses on redundant robot design and control, medical imaging, and visual servoing. He is a recipient of the NSF Graduate Research Fellowship and ARCS Foundation Scholarship. He received a B.Sc., M.S., and Ph.D from UC San Diego.
\end{IEEEbiography}

\begin{IEEEbiography}
[{\includegraphics[width=1in, height=1.25in,clip,keepaspectratio]{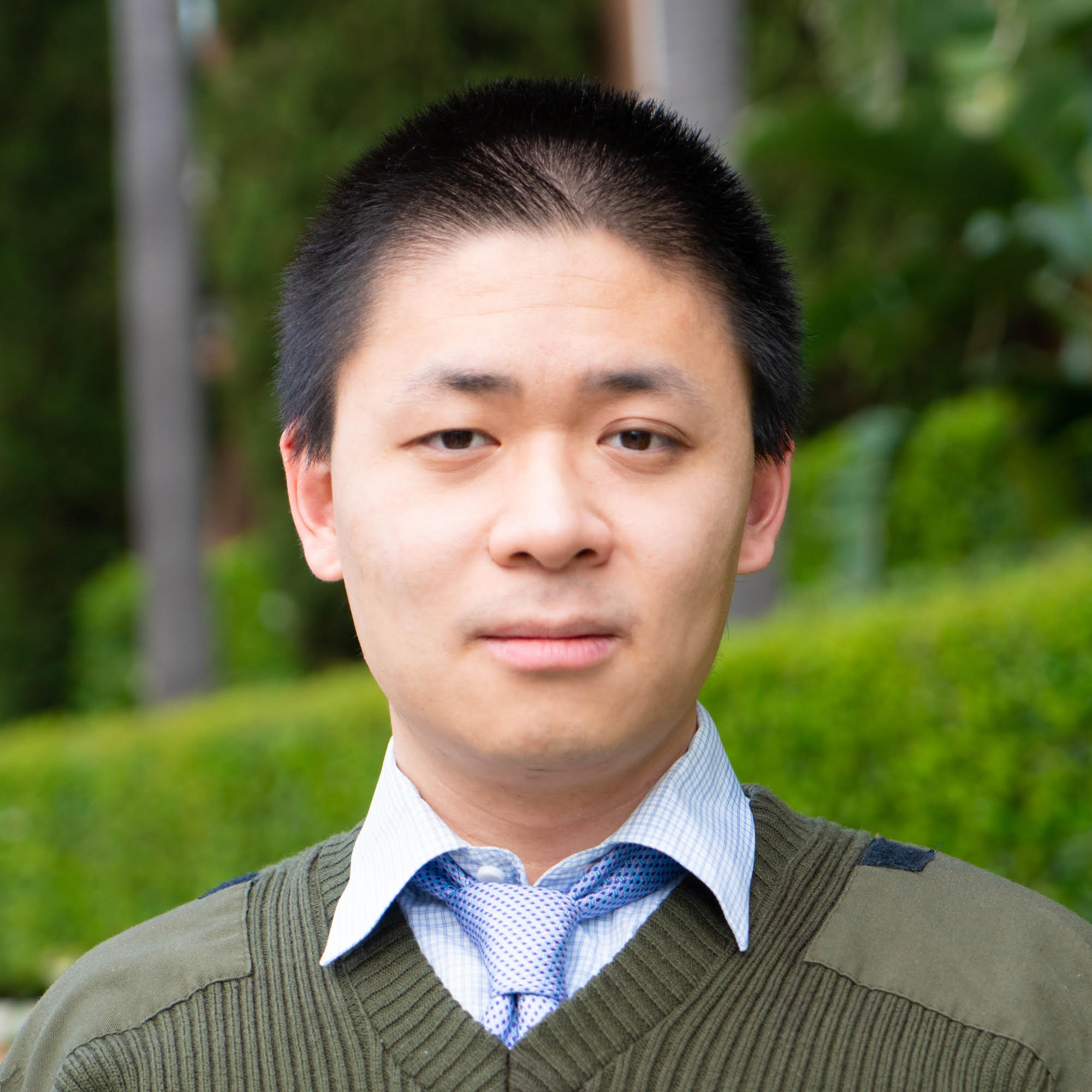}}]{Zhaowei Yu} studied communication systems as an undergrad and intelligence
systems, robotics, and control as a master's student at UCSD. He
joined the ARClab in the summer
of 2019 and has worked on multiple projects involving surgical and
underwater robots with Dimitri under the supervision of Professor
Michael C. Yip. After graduation in 2022, Zhaowei joined Zoox as a
software engineer and has been working on safety and
validation/simulation frameworks for perception systems and planner on
autonomous vehicles.
\end{IEEEbiography}

\begin{IEEEbiography}
[{\includegraphics[width=1in, height=1.25in,clip,keepaspectratio]{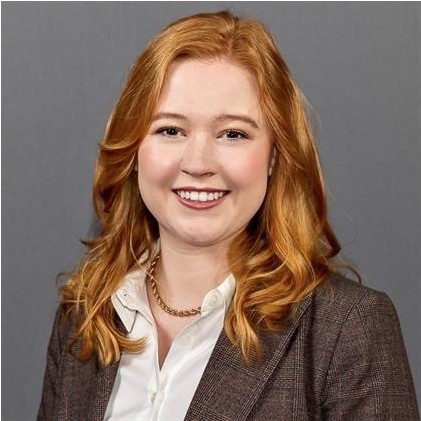}}]{Taylor Henderson} is a Program Manager at Northrop Grumman, a company dedicated to pioneering next-generation technology solutions for our customers around the globe. She received a B.Sc. and M.S. from UC San Diego. Her research includes redundant robot control, artificial muscle actuators, and scalable reinforcement learning techniques for surgical robotics.
\end{IEEEbiography}

\begin{IEEEbiography}
[{\includegraphics[width=1in, height=1.25in,clip,keepaspectratio]{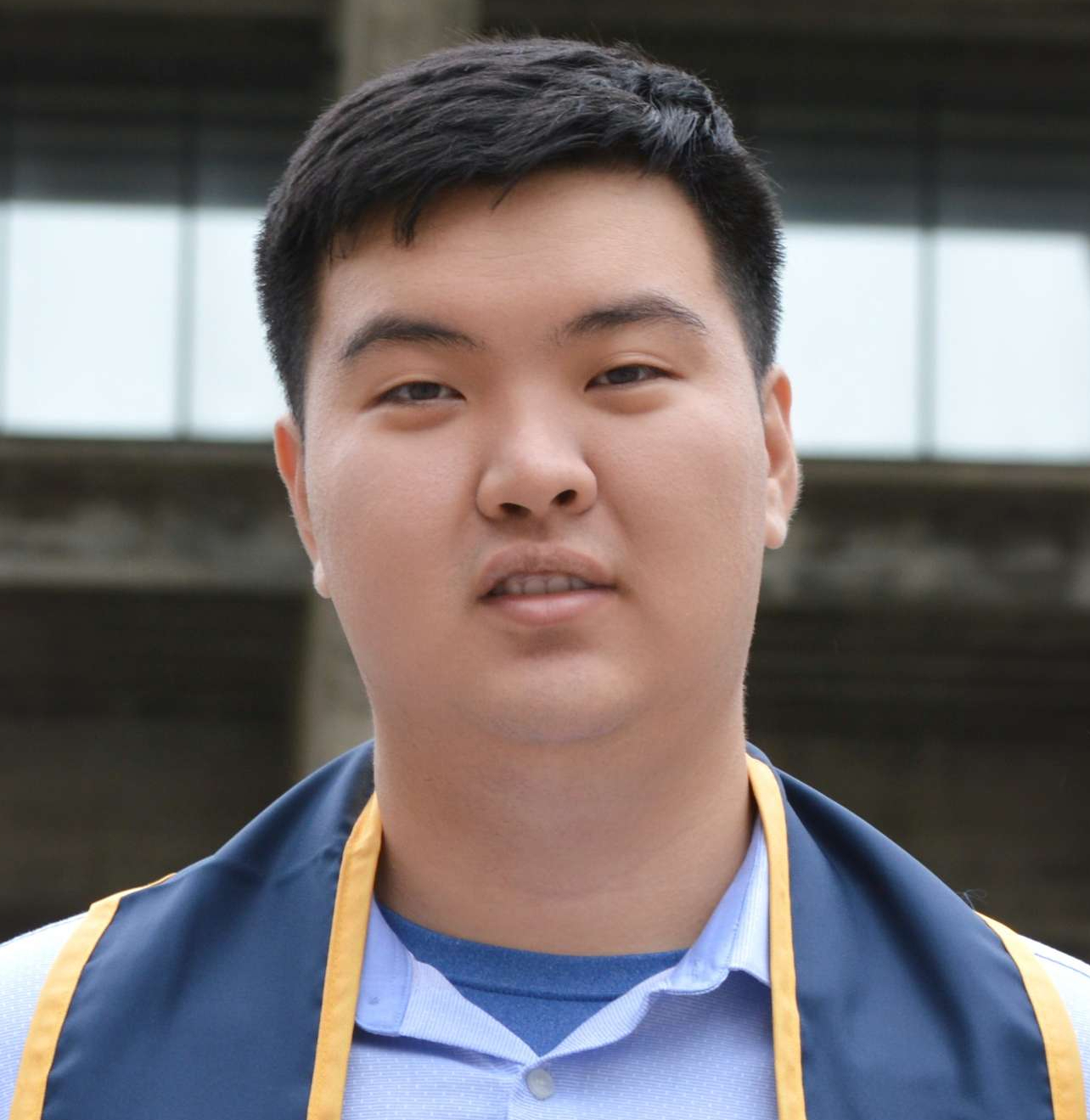}}]{Derek Chen} is a second-year master's student in the Electrical and Computer Engineering (ECE) department at UC San Diego, specializing in machine learning and data science. Currently interning at pSemi Corporation, Derek previously contributed to research in the field of medical devices within the Advanced Robotics and Controls Laboratory (ARCLab) at UCSD.
\end{IEEEbiography}

\begin{IEEEbiography}
[{\includegraphics[width=1in, height=1.25in,clip,keepaspectratio]{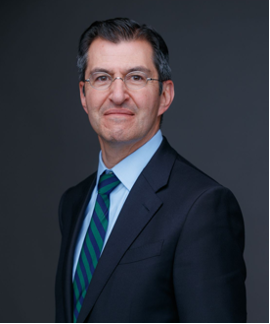}}]{Alexander Norbash} is a Professor of Radiology at the University of California, San Diego, (UCSD) in the School of Medicine,
appointed in 2015, and served as Chair of the Radiology Department at UCSD from 2015 through 2023. He has accepted
a position as Dean of the School of Medicine at the University of Missouri-Kansas City, starting March 11 th 2024. He
practiced as an interventional and diagnostic neuroradiologist from 1994 through 2015 and is currently actively
practicing as a diagnostic neuroradiologist. His translational research interests include engineering collaborations, creating novel tools and materials for endovascular neurologic therapies, and interventional robotics.
\end{IEEEbiography}

\begin{IEEEbiography}
[{\includegraphics[width=1in, height=1.25in,clip,keepaspectratio]{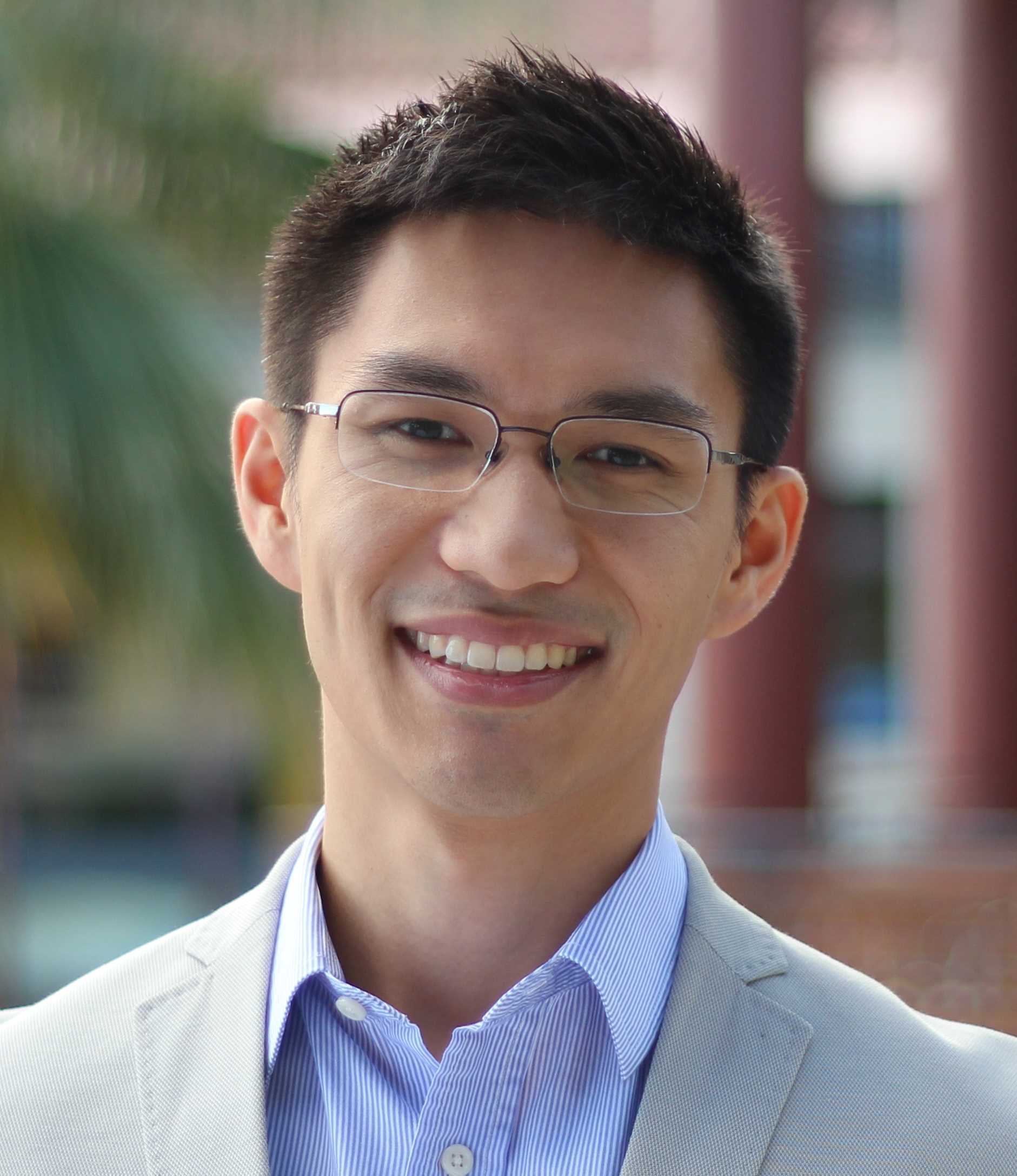}}]{Michael Yip} is an Associate Professor at UC San Diego  and directs the Advanced Robotics and Controls Laboratory. They focus on surgical robotics, 3D visual tracking and servoing, data-efficient machine learning for control and motion planning, and medical imaging. He has been nominated for and won several best paper awards at ICRA and IROS, the 2017 best paper award for RA-L, the NSF CAREER and the NIH Trailblazer award. Dr. Yip was previously with Disney Research, and Amazon Robotics. He received a B.Sc. from the University of Waterloo, an M.S. University of British Columbia, and a Ph.D. from Stanford University.
\end{IEEEbiography}

\end{document}